\documentclass[twoside,11pt]{article}

\usepackage{blindtext}

\usepackage{sty}

\usepackage{comment,times,setspace}
\usepackage[linesnumbered,ruled]{algorithm2e}
\usepackage{bm}
\usepackage{float}
\usepackage{enumitem}
\usepackage{natbib}
\usepackage{booktabs}
\usepackage{tabularx}
\usepackage{makecell}   % for easy manual line‐breaks in cells

\usepackage{multirow}
\usepackage{url,hyperref} % not crucial - just used below for the URL 
\usepackage{xr} % used to maintain crossref in a separate supplementary file
\usepackage{tikz}
\usepackage{overarrows}
\usetikzlibrary{arrows.meta}
\usepackage{orcidlink}

\usepackage{stmaryrd}
\SetSymbolFont{stmry}{bold}{U}{stmry}{m}{n}

\usepackage{subcaption}

\newtheorem{prop}{Proposition}

\newtheorem{assumption}{Assumption}

\def\spacingset#1{\renewcommand{\baselinestretch}%
{#1}\small\normalsize} \spacingset{1}

\newcommand{\smc}[1]{\scalebox{0.9}{$\mathcal{#1}$}}
\newcommand{\argmin}{\operatorname*{\arg\min}}

\newcommand{\E}{\operatorname{E}} % expectation
\newcommand{\Var}{\operatorname{Var}}

\newcommand{\bb}{\mathbb}

\usepackage{stmaryrd}

\newcommand{\R}{\bb R}

\newcommand{\Y}{\bm Y}
\newcommand{\y}{\bm y}
\newcommand{\x}{\bm x}
\newcommand{\Z}{\bm Z}
\newcommand{\X}{\bm X}
\newcommand{\z}{\bm z}

\newcommand{\U}{\bm U}
\newcommand{\s}{\bm s}

\newcommand{\V}{\bm V}

\usepackage{lastpage}

\ShortHeadings{Enhancing Accuracy in Generative Models }{Xinyu Tian and Xiaotong Shen}
\firstpageno{1}

\begin{document}

\title{Enhancing Accuracy in Generative Models via Knowledge Transfer}

\author{\name Xinyu Tian \email tianx@umn.edu \\
       \addr School of Statistics\\
       University of Minnesota\\
       Minneapolis, MN 55455, USA
       \AND
       \name Xiaotong Shen \textsuperscript{\orcidlink{0000-0003-1300-1451}} \thanks{Corresponding author}
       \email xshen@umn.edu \\
      \addr School of Statistics\\
       University of Minnesota\\
       Minneapolis, MN 55455, USA}

\editor{My editor}

\maketitle

\begin{abstract}%   <- trailing '%' for backward compatibility of .sty file
This paper investigates the accuracy of generative models and the impact of knowledge transfer on their generation precision. Specifically, we examine a generative model for a target task, fine-tuned using a pre-trained model from a source task. Building on the "Shared Embedding" concept, which bridges the source and target tasks, we introduce a novel framework for transfer learning
under distribution metrics such as the Kullback-Leibler divergence. This framework underscores the importance of leveraging inherent similarities between diverse tasks despite their distinct data distributions. Our theory suggests that the shared structures can augment the generation accuracy for a target task, reliant on the capability of a source model to identify shared structures and effective knowledge transfer from source to target learning. To demonstrate the practical utility of this framework, we explore the theoretical implications for two specific generative models: diffusion and normalizing flows. The results show enhanced performance in both models over their non-transfer counterparts, indicating advancements for diffusion models and providing fresh insights into normalizing flows in transfer and non-transfer settings. These results highlight the significant contribution of knowledge transfer in boosting the generation capabilities of these models.
\end{abstract}


\section{Introduction}

Generative modeling, augmented with transfer learning, has seen considerable advancements in improving learning accuracy with scarce data. This process distills knowledge from extensive, pre-trained models previously trained on large datasets from relevant studies, enabling domain adaptation for specific tasks. At its core is the dynamic between the source (pre-trained) and target (fine-tuning) learning tasks, which tend to converge towards shared, concise representations. Yet, this principle has received less attention in diffusion models \citep{sohl2015deep, dhariwal2021diffusion} and normalizing flows \citep{dinh2014nice, dinh2016density}.
This paper presents a theoretical framework to assess the accuracy of outputs from generative models, offering theoretical support for training generative models via transfer learning. For instance, it supports pre-training and fine-tuning of text-to-image models \citep{rombach2022high, zhou2023shifted} for domain adaptation and a synthesis approach \citep{shen2023boosting} that employs high-fidelity synthetic data to boost the effectiveness of data analytics of downstream tasks through knowledge transfer.

Accurately evaluating the fidelity of data produced by generative models is increasingly critical for downstream analyses and for maintaining users’ trust in synthetic data \citep{liu2024novel}. Although empirical studies show that transfer learning improves diffusion‑based generators for both images and tabular data \citep{wang2023efficient,kotelnikov2023tabddpm,shen2023boosting}, its theoretical effect on generative accuracy remains underexplored. Poorly matched source tasks can even induce “negative transfer,” degrading performance and jeopardizing trustworthy AI goals through misleading scientific conclusions \citep{zhang2022survey,gibney2022ai}. By contrast, transfer learning in supervised settings has been thoroughly analyzed \citep{fregier2021mind2mind,baxter2000model,maurer2016benefit,tripuraneni2020theory}, underscoring the need for principled study in the generative realm. Complementing diffusion and flow research, Generative Adversarial Networks (GANs) provide a mature toolkit for domain adaptation: feature‑level alignment via Domain‑Adversarial Training \citep{Ganin2015}; unpaired image‑to‑image translation with CycleGAN \citep{Zhu2017}; and recent multi‑domain or data‑efficient extensions such as StarGAN \citep{Choi2020} and ADA \citep{Karras2020}. These methods demonstrate that adversarial alignment—whether in latent or pixel space—remains an effective paradigm for cross‑domain generation.

We now review the relevant literature on the accuracy of two advanced generative models, diffusion models and normalizing flows. In diffusion models, \cite{oko2023diffusion} derives convergence rates for unconditional generation for smooth densities, while \cite{chen2023score} investigates distribution recovery over a low-dimensional linear subspace. Although a conditional diffusion model has shown effectiveness \citep{batzolis2021conditional}, its theoretical foundation remains underexplored. Recently,  \cite{fu2024unveil} investigated conditional diffusion models under a smooth density assumption. By comparison, the study of generation accuracy for flows remains sparse, with limited exceptions on universal approximation \citep{koehler2021representational}. 

This paper develops a comprehensive theoretical framework for transfer learning that
addresses the accuracy of target generation. This generation accuracy, measured by the excess risk, induces several valuable metrics such as the Kullback-Leibler (KL) divergence to assess the distribution closeness. To the best of our knowledge, this study is the first to outline the bounds of generation accuracy in the context of transfer learning. The contributions of this paper are as follows:

{\bf 1).  Generation accuracy theory.}
We introduce the concept of the "Shared Embedding" condition (SEC) to quantify the similarities between the latent representations of source and target learning. The SEC distinguishes between conditional and unconditional generation by featuring nonlinear dimension reduction for the former while capturing shared latent representations through embeddings for the latter. Our theoretical framework establishes generation error bounds for conditional and unconditional models. These bounds incorporate factors such as complexity measures and approximation errors while leveraging the transferability principle via shared structures. This theory offers statistical guarantees for the efficacy of generative models through knowledge transfer while demonstrating that such models can achieve rapid convergence rates for the target task under metrics stronger than commonly used total variation TV-norm. Achieving this involves leveraging the common structures for dimension reduction. 

{\bf 2). Diffusion models and normalizing flows via transfer learning.} We leverage the general theoretical framework to unveil new insights into the precision of both conditional and unconditional generation. This exploration examines conditional generation with the KL divergence and the TV-norm for
smooth target distributions and unconditional generation with the
dimension-scaling Wasserstein distance, specifically in diffusion and coupling flows, 
as detailed in Theorems \ref{thm_diff}-\ref{cor-cp-nt2}. Our focus is on the prevalent practices with smooth distributions through continuous embeddings. 
The analysis reveals that utilizing transfer learning strategies—grounded in the shared embedding structures within the lower-dimensional manifold that bridges the source and target learning—holds the potential to elevate performance over non-transfer methodologies.

{\bf 3). Non-transfer diffusion models and normalizing flows.}
This paper investigates non-transfer generative models, an area attracting considerable interest. Our results demonstrate that diffusion models structured with the SEC framework achieve a faster KL rate than their non-transfer analogs in the TV-norm for conditional generation with a smooth density \citep{fu2024unveil},
where \cite{fu2024unveil} aligns with the minimax rate in \cite{oko2023diffusion} without dimension reduction capabilities, albeit with a logarithmic factor. In unconditional generation, our method exhibits a faster rate under the Wasserstein distance relative to that under the TV-norm \citep{oko2023diffusion}. Crucially, our analysis of coupling flows reveals its competitiveness compared to diffusion models in both conditional and unconditional generation; 
see Section \ref{sec_flows} for details. These results enrich our understanding of these models' complexities and strengths.

This article comprises seven sections. Section \ref{sec:enhancing} outlines the transfer learning framework for generative tasks. Section \ref{sec_diffusion} applies the supplementary theory to diffusion models, deriving new results to illustrate knowledge transfer. Section \ref{sec_flows} introduces a novel finding for normalizing flows. Section~\ref{sec-general} presents the core proof strategy that establishes accuracy guarantees for generative models enhanced by transfer learning.
Section \ref{sec: numerical} illustrates the core theory through numerical examples. Finally, Section \ref{sec: conlusion} concludes the article. The Appendix contains technical details and experiment details.

\section{Enhancing generation accuracy and knowledge transfer}
\label{sec:enhancing}
Within the framework of synthesizing random samples that approximate a target distribution, the transfer learning approach leverages a pre-trained generative model trained on a source domain. This method fine-tunes the target generative model using the source model and training data from the target distribution, thereby enabling sample generation that accurately reflects the target distribution.

As a starting point, we adopt a basic independence assumption between the source and target datasets.

\begin{assumption}
\label{A-independent}
The source and target data are assumed to be independent.
\end{assumption}
To facilitate transfer learning between source and target tasks, we next introduce the procedures and necessary conditions for both conditional and unconditional generation.

\subsection{Conditional generation}

In the target task, we train a conditional generator for $\X_t$ given $\Z_t$ using a target training sample $\smc{D}_t=\{\x_t^i, \z_t^i\}_{i=1}^{n_t}$, whereas source training occurs separately with an independent source training sample
$\smc{D}_s=\{\x_s^i, \z_s^i\}_{i=1}^{n_s}$. 

\noindent\textbf{SEC for conditional generation.} We introduce the "Shared Embedding" condition for conditional generation.
Denote the target and source covariate vectors by $\X_t$ and $\X_s$, which are allowed to differ in dimensionality.
To sample from the conditional distribution of the target covariates given an auxiliary vector $\Z_t$, denoted
by $P_{\X_t\mid\Z_t}$, we transfer information from the source task to improve target‑side estimation. Decompose the auxiliary vectors as
\begin{eqnarray*}
 \Z_t=(\Z,\Z_{t^c}), \quad \Z_s=(\Z,\Z_{s^c}), 
\end{eqnarray*}
where the common block $\Z \in \mathbb{R}^{d_c}$ is shared across tasks and $\Z_{j^c}$ contains the task‑specific 
remainder; $j \in \{s,t\}$, and $d_c$ is the dimension of $\Z$.

\paragraph{Shared Embedding Condition (SEC).} Assume there exists a latent representation $h(\Z)$ that is common to both tasks such that the conditional laws factor through task‑specific decoders $P_t$ and $P_s$:
\begin{eqnarray}
\label{c-shared}
P_{\x_t|\z_t}(\cdot|\z_t) = P_{t}(\cdot, h_t(\z_t)), \quad P_{\x_s|\z_s}(\cdot|\z_s) = P_{s}(\cdot, h_s(\z_s)). 
\end{eqnarray}
where $h_j(\z_j)=(h(\z),\z_{j^c})$ and $P_j$ is a suitable probability function; $j \in \{s,t\}$. 
For an explicit illustration of SEC, see Figure \ref{cg}, which highlights the shared‑embedding architecture within target and source diffusion models. This design mirrors the practical fine‐tuning strategy in text‐to‐image pipelines, where the text‐embedding module is frozen while adapting the diffusion backbone.

The SEC in \eqref{c-shared} presents a dimension reduction framework, indicating that $P_{\x_j|\z_j}$ depends on a shared manifold mapping $h(\z)$, generally of lower dimension; $j \in \{s,t\}$. For instance, a source task of text prompt-to-image ($\Z_s$ to $\X_s$) and a target task of text prompt-to-music ($\Z_t$ to $\X_t$) may share common elements $\Z$ based on a latent semantic representation $h(\Z)$ and task-specific elements $\Z_{j^c}$; $j \in \{s,t\}$. This framework broadens the scope of dimension reduction from linear subspaces \citep{li1991sliced, li2018sufficient} to nonlinear manifolds, where $\X_j$ is conditionally independent of $\Z_j$ given $(h(\Z),\Z_{j^c})$.

Here, our primary focus is on understanding the interplay between the source distribution of $\X_s$ given $\Z_s$, 
$P_{\x_s|\z_s}(\x|\z_s)=P_s(\x, h_s(\z_s))$, and the target distribution of $\X_t$ given $\Z_t$, 
$P_{\x_t|\z_t}(\x|\z_t) =P_t(\x, h_t(\z_t))$, through the shared SEC component $h$ between $h_s$ and $h_t$. 

    To learn $P_{\x_j|\z_j}$, we parameterize it as $P_{\x_j|\z_j}(\x,\z_j) = P_{j}(\x, h_j(\z_j); \theta_j)$ with $h$ from $\Theta_h$, the space of latent embeddings, and $\theta_j$ from $G_j$, either parametric or nonparametric; $j = s, t$. This approach defines the true distribution $P^0_{\x_j|\z_j}(\x,\z_j) = P_{j}(\x, h^0_j(\z_j); \theta^0_j)$ through true parameters $h^0_j(\z_j)=(h^0(\z),\z_{j^c})$ and $(\theta^0_j, h^0) = \argmin_{\theta_j \in G_j, h \in \Theta_h} \mathrm{E}_{\x_j,\z_j} l_j(\X_j, \Z_j; \theta_j, h)$, minimizing the expected loss $l_j$; $j = s, t$. Here, $\mathrm{E}_{\x_j,\z_j}$ is the expectation of $(\X_j,\Z_j)$ and we use $\Theta_j$ (a class of neural networks) as the action parameter space approximating the parameter space $G_j$ (a class of candidate functions); $j = s, t$. For the source task, we minimize its empirical loss $L_s(\theta_s, h)=\sum_{i=1}^{n_s} l_s(\x_s^i, \z_s^i; \theta_s, h)$ on a source training sample to yield 
\begin{eqnarray}
\label{hat-h}
(\hat{\theta}_s, \hat{h}) = \argmin_{\theta_s \in \Theta_s, h \in \Theta_h} L_s(\theta_s, h), 
\end{eqnarray}
where $\Theta_h$ ensures latent structure identifiability. With $\hat{h}$, we minimize the target empirical loss $L_t(\theta_t, \hat{h}) = \sum_{i=1}^{n_t} l_t(\x_t^i, \z_t^i; \theta_t, \hat{h})$ to yield $\hat{\theta}_t = \argmin_{\theta_t \in \Theta_t} L_t(\theta_t, \hat{h})$. The estimated distribution is $\hat{P}_{\x_j|\z_j}(\x|\z_j) = P_{\x_j|\z_j}(\x, \hat{h}_j(\z_j); \hat{\theta}_j)$, where $\hat{h}_t(\z_j)=(\hat{h}(\z),\z_{j^c})$. The distribution discrepancy is controlled by the excess risk $\E_{\x_j,\z_j} (l_j(\X_j, \Z_j; \theta_j, h) - l_j(\X_j, \Z_j; \theta_j^0, h^0))$.
For example, the negative log-likelihood loss yields an error bound in the excess risk, implying that in the KL divergence.

\begin{assumption}[Transferability for conditional models]
\label{transferability}
For some positive constant $c_1>0$ and $h\in \Theta_h$,
$
|\delta_t(h)-\delta_t(h^0)| \leq c_1 |\delta_s(h)-\delta_s(h^0)|,
 $
where $\delta_j(h)=\inf_{\theta_j\in \Theta_j}\E_{\x_j,\z_j} [l_j(\X_j,\Z_j;\theta_j,h)-l_j(\X_j,\Z_j;\theta_j^0,h^0)]$; $j \in \{s,t\}$. 
\end{assumption}   

Assumption \ref{transferability} characterizes the transitions of the excess risk for the latent structural representation $h$ from source to target tasks. A similar condition has been in a different context \citep{tripuraneni2020theory}.

Denote the excess risk as $\rho^2_j(\gamma_j^0, \gamma_j)=\E_{\x_j,\z_j} [l_j(\X_j,\Z_j;\theta_j,h)-l_j(\X_j,\Z_j;\theta_j^0,h^0)]$ with $\gamma_j=(\theta_j,h)$; $j \in \{s,t\}$. The following assumption specifies the generation error bound of the source for $\X_s$ given $\Z_s$, facilitating the target learning through transfer learning.

\begin{assumption}[Source error]
\label{A-error} There exists a sequence $\varepsilon_s$ indexed by $n_s$, such that the source generation error satisfies, for any $\varepsilon\geq \varepsilon_s$,
$
P(\rho_{s}(\gamma_s^0, \hat \gamma_s) \geq \varepsilon)\leq  \exp\left(-c_2 n_s^{1-\xi}\varepsilon^2\right)
$,
where $c_2>0$  and $\xi>0$ are constants, $\hat \gamma_s=(\hat \theta_s, \hat h)$ is defined in \eqref{hat-h}, and $n_s^{1-\xi}\varepsilon_s^2\rightarrow \infty$ as $n_s \rightarrow \infty$.
\end{assumption}

Because the source error for $h$ is typically intertwined with that of $\theta_s$, any estimation error in $\theta_s$ carries over to $h$ through Assumption \ref{transferability} and, in turn, affects the error in conditional generation.

\subsection{Unconditional generation}

\noindent\textbf{SEC for unconditional generation.} To sample from the marginal target distribution $P_{\X_t}$, we transfer a latent representation learned on the source task. The SEC postulates that the source and target variables, $\X_s$ and $\X_t$, arise from a \emph{shared} latent vector $\U$ through task‑specific decoders. Let $g_t$ and $g_s$ map the latent space to the target and source observation spaces, respectively, so that $\X_t = g_t(\U)$ and $\X_s = g_s(\U)$.
Consequently,
\begin{eqnarray}
\label{shared}
P_{\x_t}(\cdot)=P_{\bm u}(g_{t}^{-1}(\cdot)), \quad P_{\x_s}(\cdot)=P_{\bm u}(g_{s}^{-1}(\cdot)), 
\end{eqnarray}
where $P_{\bm u}$ denotes the probability distribution of $\bm{u}$ and $g^{-1}$ denotes the inverse image of $g$, or
$\{\bm u: g(\bm u)=x_t\}$. Figure \ref{ucg} gives a concrete illustration of this shared‑embedding architecture for source and target diffusion models. This configuration parallels the practical fine‐tuning workflow for latent diffusion models, in which the diffusion backbone is frozen and only the decoder is adapted.

This SEC in \eqref{shared} highlights that the source and target distributions share a common latent distribution within low-dimensional manifolds, defined by latent representations $g_{s}$ and $g_{t}$. 
For example, consider a source task of French text generation $\X_s$ from English and a target task of Chinese text generation $\X_t$ from English. Initially, the numerical embedding of a textual description $\U$ in English is transformed into  French and Chinese using the transformations $g_s$ and $g_t$, respectively.

Given a latent representation $\{\bm{u}_s^i\}_{i=1}^{n_s}$ for $\{\x_s^i\}_{i=1}^{n_s}$ in the source training sample $\smc{D}_s$ encoded by an encoder, we first estimate the latent distribution $P_{u}$ using a generative model parameterized by $\Theta_{u}$ as $\hat{P}_{u} = P_{u}(\cdot; \hat{\theta}_{u})$.
Here, $\hat{\theta}_{u} = \argmin_{\theta_{u} \in \Theta_{u}} L_u(\theta_{u}) = \argmin_{\theta_{u} \in \Theta_{u}} \sum_{i=1}^{n_s} l_u (\bm{u}_s^i; \theta_{u})$, 
where $\sum_{i=1}^{n_s} l_u (\bm{u}_s^i; \theta_{u})$ represents an empirical loss to estimate
$\theta_{u}$. With the estimated latent distribution $\hat{P}_{u}$, $g_t$ is estimated as
$\hat{g}_{t} = \argmin_{g_{t} \in \Theta_{g_{t}}} L_{g_{t}}(g_{t}) = \argmin_{g_{t} \in \Theta_{g_{t}}} \sum_{i=1}^{n_t} l_{g_t}(\bm{u}_t^i, \x_t^i; g_t)$ based on a target training sample
$\smc{D}_t=\{\bm{u}_t^i, \x_t^i\}_{i=1}^{n_t}$. Then $P_{\x_t}$ is estimated by $\hat P_{\x_t}(\cdot) = \hat{P}_{u}(\hat{g}_t^{-1}(\cdot))$.

Assumption \ref{transferability} is not required for unconditional generation due to different shared structures in \eqref{shared}, with a separable loss function for $\theta_u$ and $g_t$.   
As in the conditional setting, $\theta^0_u$ is defined as the minimizer of the population loss, given by $\theta^0_u=\argmin_{\theta_u\in G_u} \mathrm{E}_u l_u (\bm{u}; \theta_{u})$, where $G_u$ represents the parameter space of $\theta_u$. Let $\rho^2_{u}(\theta_u^0,\theta_u)=\mathrm{E}_{u}[l_u (\bm{U}; \theta_u)-l_u (\bm{U}; \theta^0_u)]$.

Moreover, we assume an analogous condition to Assumption~\ref{A-error} for the source error of $\rho_{u}(\theta_u^0, \hat\theta_u)$.

\begin{assumption}[Source error for $\U$]
\label{U-error} 
There exists a sequence $\varepsilon_s$ indexed by $n_s$,
such that the source generation error for $\U$ by estimating $\theta_u$
from $\{\bm{u}_s^i\}_{i=1}^{n_s}$ satisfies for any $\varepsilon\geq \varepsilon_s^u$, 
$
P(\rho_{u}(\theta_u^0, \hat\theta_u)\geq \varepsilon )\leq  \exp\left(-c_3 n_s^{1-\xi}\varepsilon^2\right)
$,
where $c_3>0$ and $\xi> 0$ are constants and $n_s^{1-\xi}(\varepsilon^u_s)^2\rightarrow \infty$ as $n_s \rightarrow \infty$.
\end{assumption}

The error bound $\varepsilon_s$, $\varepsilon^u_s$, $\xi$, $c_2$ and $c_3$ in Assumptions \ref{A-error} and \ref{U-error} can be determined in a
specific source model; cf. Lemmas \ref{thm_diff_source}, \ref{thm_pu}, \ref{thm_floW_source}, and \ref{thm_pu-flow}.

Before detailing the diffusion and flow frameworks, we present a concise overview of our theoretical guarantees in Table \ref{tab:theory_summary}. This table lists each generative model, indicates whether the task is conditional or unconditional, specifies the transfer regime, enumerates the key assumptions, cites the corresponding theorem, and identifies the performance metric.

\begin{table}[H]
\centering
\caption{Key assumptions, formal results, and distance metrics underlying our theoretical guarantees.}
\label{tab:theory_summary}
\begin{tabularx}{\textwidth}{@{}l p{2.1cm} l p{4.4cm} l l@{}}
\toprule
Model     & Task                           & Regime       & Assumptions                                                  & Theorem                          & Metric                     \\
\midrule
\multirow{4}{*}{Diffusion}
          & \makecell[l]{Conditional\\generation}
                                         & Transfer     & \makecell[l]{SEC; transferability; \\density smoothness; \\source error}      & \ref{thm_diff}                   & TV, KL                    \\
          &                                 & Non-transfer & SEC; density smoothness                                        & \ref{cor-nt}                     & TV, KL                    \\
          &                                 & General      & Density smoothness                                             & \ref{thm_diff_general}           & TV, KL                    \\
\cmidrule(lr){2-6}
          & \makecell[l]{Unconditional\\generation}
                                         & Transfer     & \makecell[l]{SEC; source error; \\transformation smoothness}            & \ref{thm_ug}                     & Wasserstein                     \\
          &                                 & Non-transfer & SEC; transformation smoothness                                 & \ref{cor-nt2}                    & Wasserstein                     \\
\midrule
\multirow{4}{*}{Flow}
          & \makecell[l]{Conditional\\generation}
                                         & Transfer     & \makecell[l]{SEC; transferability; \\transformation smoothness; \\source error} & \ref{thm_flow}                   & KL                        \\
          &                                 & Non-transfer & SEC; transformation smoothness                                 & \ref{cor-cp-nt}                  & KL                        \\
          &                                 & General      & Transformation smoothness                                      & \ref{thm_cp_general}             & KL                        \\
\cmidrule(lr){2-6}
          & \makecell[l]{Unconditional\\generation}
                                         & Transfer     & \makecell[l]{SEC; source error; \\transformation smoothness}           & \ref{thm_flow2}                  & Wasserstein                     \\
          &                                 & Non-transfer & SEC; transformation smoothness                                 & \ref{cor-cp-nt2}                 & Wasserstein                     \\
\bottomrule
\end{tabularx}
\end{table}

\section{Diffusion models} \label{sec_diffusion}

This section considers diffusion models, following the setup in Section \ref{sec:enhancing}.

\subsection{Forward and Backward Processes\label{sec_3-1}}

A diffusion model incorporates both forward and backward diffusion processes.

\noindent \textbf{Forward process. } The forward process systematically transforms a random vector $\X(0)$ into pure white noise by progressively
injecting white noise into a differential equation defined with the Ornstein-Uhlenbeck process, leading to diffused distributions from the initial state $\X(0)$:
\begin{equation}
\label{forward}
\mathrm{d}\X(\tau)=-{b}_{\tau} \X(\tau)\mathrm{d}\tau+\sqrt{2{b}_\tau}\mathrm{d}W(\tau),\quad \tau \geq 0,
\end{equation}
where $\X(\tau)$ has a probability density $p_{\x(\tau)}$, $\{W(\tau)\}_{\tau\geq 0}$ represents a standard Wiener process and ${b}_t$ is a non-decreasing weight function. Under \eqref{forward}, $\X(\tau)$ given $\X(0)$ follows $N(\mu_{\tau}\X(0),\sigma^2_{\tau}\bm I)$, where $\mu_{\tau}=\exp(-\int_0^\tau {b}_s\mathrm{d} s)$ and $\sigma^2_{\tau}=1-\mu_{\tau}^2$. 
By setting $b_s = 1$, we simplify the model to $\mu_{\tau} = \exp(-\tau)$ and $\sigma^2_{\tau} = 1 - \exp(-2\tau)$. In practice, the diffusion process terminates at a sufficiently large $\overline{\tau}$, ensuring the distribution of $\X(\tau)$, which is a mixture of the original state $\X(0)$ and white noise, approximates a standard Gaussian vector.

\noindent \textbf{Backward process.} Given $\X(\overline{\tau})$ in \eqref{forward}, a backward process is employed for sample generation for $\X(0)$. Assuming \eqref{forward} satisfies certain conditions \citep{anderson1982reverse}, the backward process $\bm V(\tau)=\X(\overline{\tau}-\tau)$, starting with $\X(\overline{\tau})$, is derived as:
\begin{equation}
\label{reverse}
\mathrm{d}\bm V(\tau)={b}_{\overline{\tau}-\tau}(\bm V(\tau)+2\nabla\log p_
{\x(\overline{\tau}-\tau)}(\X(\overline{\tau}-\tau))\mathrm{d}\tau+\sqrt{2{b}_{\overline{\tau}-\tau}}\mathrm{d}W(\tau); \quad \tau \geq 0,
\end{equation}
where $\nabla\log p_{\x}$ is the score function which represents the gradient of $\log p_{\x}$.

\noindent \textbf{Score matching.} To estimate the unknown score function, we minimize a matching loss between the score and its approximator $\theta$:
$\int_{0}^{\overline{\tau}}\mathrm{E}_{\x(\tau)}\|\nabla \log p_{\x(\tau)}(\X(\tau))-\theta(\X(\tau),\tau)\|^2\mathrm{d}\tau$,
where $\|\x\|=\sqrt{\sum^{d_x}_{j=1}\x_j^2}$ is the Euclidean norm, which is equivalent to minimizing the following loss \cite{oko2023diffusion},
\begin{equation}
\label{loss_2}
\int_{\underline{\tau}}^{\overline{\tau}}\mathrm{E}_{\x(0)}\mathrm{E}_{\x(\tau)|\x(0)}\|\nabla \log p_{\x(\tau)|\x(0)}(\X(\tau)|\X(0))-\theta(\X(\tau),\tau)\|^2\mathrm{d}\tau,
\end{equation}
with $\underline{\tau}=0$. In practice, to avoid score explosion due to $\nabla \log p_{\x(\tau)|\x(0)} \rightarrow \infty$ as $\tau\rightarrow 0$ and to ensure training stability, we restrict the integral interval to $\underline{\tau}>0$ \citep{oko2023diffusion,chen2023improved} in the loss function.
Then, both the integral and $\mathrm{E}_{\x(0)}$ can be precisely approximated by sampling $\tau$ from a uniform distribution on $[\underline{\tau},\overline{\tau}]$ and a sample of $\X(0)$ from the conditional distribution of $\X(0)$ given $\Z$.

\noindent \textbf{Generation.} To generate a random sample of $\bm V(\tau)$, we replace the score $\nabla\log p_{\x(\overline{\tau}-\tau)}$ by its estimate $\hat \theta$ in \eqref{reverse} to yield $\bm V(\tau)$ in the backward equation. For implementation, we may utilize a discrete-time approximation of the sampling process, facilitated by numerical methods for solving stochastic differential equations, such as Euler-Maruyama and stochastic Runge-Kutta methods \citep{song2020denoising}.

%For a conditional generation, we start with the conditional distribution of $\X(0)$ given a specific event of interest, as discussed in \cite{song2020denoising,batzolis2021conditional}.

\subsection{Conditional diffusion via transfer learning \label{sec_3-2}}

To generate a target sample from $\X_t$ given $\Z_t$, we use a conditional diffusion model to learn the conditional probability density $p_{\x_t|\z_t}$, as described in \eqref{forward}-\eqref{reverse}. 

In this approach, we assign $\X(0)=\X_t$ to our target task in \eqref{forward}. After deriving an estimated latent structure $\hat{h}$ from the pre-trained diffusion model (source), we employ a conditional diffusion model (target), transferring $\hat h$ to improve the synthesis task of generating $\X_t$ given $\Z_t$.

Given a target training sample $(\x_t^i,\z_t^i)_{i=1}^{n_t}$, we follow \eqref{forward}-\eqref{reverse} to construct an empirical score matching loss $L_t(\theta_t,\hat{h})=\sum_{i=1}^{n_t} l_t(\x_t^i,\z_t^i;\theta_t,\hat{h})$ 
in \eqref{loss_2} with  
\begin{align}
\label{loss-diffusion}
l_t(\x_t^i,\z_t^i;\theta_t,\hat{h})=
\int_{\underline{\tau}_t}^{\overline{\tau}_t} \mathrm{E}_{\x(\tau)|\x(0)}\|\nabla \log p_{\x(\tau)|\x(0)}(\X(\tau)|\x_t^i)-\theta_t(\X(\tau),\hat{h}_t(\z_t^i),\tau)\|^2 \mathrm{d}\tau,
\end{align}
where $(\underline{\tau}_t, \overline{\tau}_t)$ denotes early stopping 
for $(0,+\infty)$ and $\hat{h}_t(\z_t)=(\hat{h}(\z),\z_{t^c})$. The estimated score $\hat \theta_t(\x(\tau),\hat h(\z),\tau)=\argmin_{\theta_t\in\Theta_t}L_t(\theta_t,\hat{h})$.
We will use the neural network for $\Theta_t$.

\noindent \textbf{Neural network. } An $\mathbb{L}$-layer network $\Phi$ is defined by a composite function
$
\Phi(\x)=(\bm{\mathrm{A}}_\mathbb{L}\sigma(\cdot)+\bm{b}_\mathbb{L})\circ\cdots(\bm{\mathrm{A}}_2\sigma(\cdot)+\bm{b}_2)\circ (\bm{\mathrm{A}}_1\x+\bm{b}_1),
$
where $\bm{\mathrm{A}}_i\in \R^{d_{i+1}\times d_i}$ is a weight matrix and $\bm b_i \in \R^{d_{i+1}}$ is the bias of a linear transformation of the $i$-th layer, and $\sigma$ is the ReLU activation function, defined as $\sigma(\x)=\max(\x,0)$.
Then, the parameter space $\Theta_t$ is set as $\mathrm{NN}(\mathbb{L}_t,\mathbb{W}_t,\mathbb{S}_t,\mathbb{B}_t,\mathbb{E}_t)$ 
 with $\mathbb{L}_t$ layers, a maximum width of $\mathbb{W}_t$, effective parameter number $\mathbb{S}_t$, the sup-norm $\mathbb{B}_t$, and parameter bound $\mathbb{E}_t$:
\begin{align}
\label{p-space}
\mathrm{NN}&(\mathbb{L}_t,\mathbb{W}_t,\mathbb{S}_t,\mathbb{B}_t,\mathbb{E}_t)=\{ \Phi: \mathbb{R}^{d_{x_t}+d_{h_t}+1}\rightarrow \mathbb{R}^{d_{x_t}}, \max_{1\leq i\leq \mathbb{L}_t}d_i\leq\mathbb{W}_t, \sum_{i=1}^{\mathbb{L}_t}(\|\bm{\mathrm{A}}_i\|_0+ \|\bm{b}_i\|_0)\leq \mathbb{S}_t, \nonumber \\
&\sup_{\x\in\mathbb{R}^{d_{x_t}+d_{h_t}}}\|\Phi(\x,\tau)\|_{\infty}\leq \mathbb{B}_t(\tau),
\sup_{\tau}\mathbb{B}_t(\tau)\leq\mathbb{B}_t,
\max_{1\leq i\leq \mathbb{L}_t}(\|\bm{\mathrm{A}}_i\|_{\infty}, \|\bm{b}_i\|_{\infty})\leq \mathbb{E}_t
\},
\end{align}
where $d_{x_t}$ and $d_{h_t}$ denote the dimensions of $\X_t$ and the output of $h_t$, respectively, $\|\cdot\|_{\infty}$ denotes the maximum absolute value of the entries, and $\|\cdot\|_0$ denotes the number of nonzero entries.

\textbf{Conditional generation.} We approximate \eqref{reverse} by substituting the score $\nabla \log p_{\x(\tau)|\z_t}$ with its estimate $\hat{\theta}_t$, yielding:
\begin{equation}
\label{s-reverse}
\mathrm{d}\hat{\V}(\tau)={b}_{\overline{\tau}_t-\tau}(\hat{\V}(\tau)+2 \hat{\theta}_t(\hat{\V}(\tau),\z_t,\overline{\tau}_t-\tau) )\mathrm{d}\tau+\sqrt{2{b}_{\overline{\tau}_t-\tau}}\mathrm{d}W(\tau), \tau\in[0,\overline{\tau}_t-\underline{\tau}^*_{t}],
\end{equation}
where we start the backward process from an initial state $\hat{\bm V}(0) \sim N(\bm 0, \bm I)$ and terminate the process at $\tau=\overline{\tau}_t-\underline{\tau}^*_{t}$ with $0\leq \underline{\tau}^*_{t}\leq \underline{\tau}_t$, which will be determined later based on the density smoothness. We then utilize $\hat{\bm V}(\overline{\tau}_t-\underline{\tau}^*_{t})$ as a generated sample to approximate $\X(0)$. The resulting conditional density estimate $\hat{p}_{\x_t|\z_t}$ corresponds to the distribution $p_{\hat{\bm v}(\overline{\tau}_t-\underline{\tau}^*_{t})|\z_t}$.
Note that, because we apply early stopping at \(\underline{\tau}_t\) during training of \eqref{loss-diffusion}, we need to extend the reverse--time interval from $\tau \in [0,\,\overline{\tau}_t-\underline{\tau}_t]$ to $\tau \in [0,\,\overline{\tau}_t-\underline{\tau}^*_{t}]$. For the extended segment $\tau \in [\overline{\tau}_t-\underline{\tau}_t,\,\overline{\tau}_t-\underline{\tau}^*_{t}]$, we freeze the 
estimator, replacing
$\hat{\theta}_t\!\bigl(\hat{\bm v}(\tau),\z_t,\overline{\tau}_t-\tau\bigr)$ by
$\hat{\theta}_t\!\bigl(\hat{\bm v}(\overline{\tau}_t-\underline{\tau}_t),\z_t,\underline{\tau}_t\bigr)$.

Next, we introduce assumptions specific to diffusion models.

\noindent \textbf{Smooth class. } Let $\bm \alpha$ be multi-index with $|\bm \alpha| \leq \lfloor 
r\rfloor$, where $\lfloor r\rfloor$ is the integer part of $r>0$.  A H\"older ball $\smc{C}^{r}(\smc D,\R^m,B)$ of radius $B$ with the degree of smoothness $r$ from domain $\smc D$ to $\R^m$ is defined by: 
\begin{equation*}
\label{eq-holder}
 \left\{(g_1,\cdots,g_m): \max_{1\leq l\leq m}\left(\max_{|\bm \alpha| \leq \lfloor r\rfloor}\sup_{\substack{\x}} |\partial^{\bm \alpha} g_l(\x)| + 
\max_{|\bm \alpha|=\lfloor r\rfloor}\sup_{\substack{\x \neq \y}}\frac{|\partial^{\bm \alpha} g_l(\x) - \partial^{\bm \alpha} g_l(\y)|}{\|\x - \y\|^{r-\lfloor r\rfloor}}\right) < B \right\}.  
\end{equation*}

\begin{assumption}[Target density]
\label{A_p0}
Assume that the true conditional density of $\X_t$ given $\Z_t$ is expressed as $p^0_{\x_t|\z_t}(\x|\z_t) = \exp(-c_4\|\x\|^2 / 2) \cdot f_t(\x,h_t^0(\z_t))$, where $f_t$ is a non-negative function and $c_4>0$ is a constant. Additionally, $f_t$ is lower bounded by a positive constant and belongs to
a H\"older ball $\smc{C}^{r_t}(\mathbb{R}^{d_{x_t}} \times [0,1]^{d_{h_t}},\R, B_t)$ 
with $r_t>0$ and $B_t>0$, where $d_{h_t}$ is the dimension of $(h(\z),z_{t^c})$. 
\end{assumption}

Assumption \ref{A_p0} posits that the density ratio between the target and a Gaussian density falls within a H\"older class, bounded by upper and lower limits. This condition for $p^0_{\x_t|\z_t}$ has been cited in the literature for managing approximation errors in diffusion models, as shown in \cite{fu2024unveil, oko2023diffusion}. This condition, introduced in \cite{fu2024unveil}, relaxes the restricted support condition in \cite{oko2023diffusion}, leading to some smooth characteristics of the score function used in \cite{chen2023sampling,chen2023improved,chen2023score}.

Next, we present the error bounds for conditional diffusion generation via transfer learning. The network $\Theta_t$ and the stopping criteria are set with specified parameters: 
\begin{align}
\label{eq-diff-set}
&\mathbb{L}_t = c_L \log^4 K, \mathbb{W}_t = c_W K \log^7 K, \mathbb{S}_t = c_S K \log^9 K, \log \mathbb{B}_t = c_B \log K, \log \mathbb{E}_t = c_E \log^4 K,\nonumber \\
&\log\underline{\tau}_t = -c_{\underline{\tau}_t}\log K, \overline{\tau}_t = c_{\overline{\tau}_t}\log K, \\underline{\tau}^*_{t} = \mathbb{I}_{\{r_t\le 1\}}\underline{\tau}_t,
\end{align}
for sufficiently large constants $c_L-c_{\underline{\tau}_t}$,
with $K$ a tuning parameter for its complexity, depending on the training size $n_t$, the smoothness degree $r_t$, and the dimensions of $\X_t$ and $h_t$,
$d_{x_t}$ and $d_{h_t}$. Note that the configuration $\mathbb{W}_t\gg\mathbb{L}_t$ corresponds to a wide network. 

The choice of \(\underline{\tau}^*_{t}\) enables us to extend the backward SDE all the way to \(\tau = \overline{\tau}_t\), thereby approximating \(\X_t(0)\) by \(\widehat V(0)\). This extension is valid under the smoothness condition \(r_t > 1\), which ensures the desired continuity of the score function and allows its behavior over the tail interval \([0,\underline{\tau}]\) to be well represented by the score at \(\underline{\tau}\).

To establish Theorems \ref{thm_diff}--\ref{cor-nt2}, we address several technical challenges involved in integrating source and target tasks. Specifically, we ensure model transferability for conditional generation and develop suitable latent representations for unconditional generation, while effectively controlling the source error as detailed in Assumption \ref{A-error}. Central to our approach is the utilization of the lower-dimensional manifold structure defined by the shared embedding condition (SEC), which significantly enhances distribution estimation accuracy. Additionally, we capitalize on structural properties specific to diffusion models to achieve our results.

\begin{theorem}[Conditional diffusion via transfer learning]
\label{thm_diff}
Under Assumptions \ref{A-independent}-\ref{A-error} and \ref{A_p0}, with setting \eqref{eq-diff-set} by $K\asymp n_t^{\frac{d{x_t} + d_{h_t}}{d{x_t} + d_{h_t} + 2r_t}}$, the generation error of conditional transfer diffusion models is
\begin{eqnarray*}
\mathrm{E}_{\smc{D}_t,\smc{D}_s}\mathrm{E}_{\z_t}[\mathrm{TV}(p^0_{\x_t|\z_t},\hat p_{\x_t|\z_t})]=O(n_t^{-\frac{r_t}{d_{x_t}+d_{h_t}+2r_t}}\log^{m_t} n_t+\varepsilon_s), \text{ if } r_t>0;\\
\mathrm{E}_{\smc{D}_t,\smc{D}_s}\mathrm{E}_{\z_t}[\smc{K}^{\frac{1}{2}}(p^0_{\x_t|\z_t},\hat p_{\x_t|\z_t})]=O(n_t^{-\frac{r_t}{d_{x_t}+d_{h_t}+2r_t}}\log^{m_t} n_t+\varepsilon_s), \text{ if } r_t>1.
\end{eqnarray*}
Here, $m_t = \max(\frac{19}{2}, \frac{r_t}{2}+1)$, $\asymp$ means mutual boundedness, and $\mathrm{TV}$ and $\smc{K}$ denote the TV-norm and the KL divergence.
\end{theorem}

A formal non-asymptotic bound is provided in Theorem \ref{thm_diff-detail} in Appendix \ref{appendix}.

To compare transfer conditional and non-transfer diffusion generations, we adopt the framework of the transfer model while omitting source learning. Specifically, we define the conditional distribution  without leveraging the source knowledge of $h$ from the
estimated score function 
\begin{equation}
\tilde\theta_t(\x_t,\tilde h_t(\z_t),\tau)=\argmin_{\theta_t\in \tilde\Theta_t, h\in \Theta_h} L_t(\theta_t,h).
\end{equation}

We impose a smoothness constraint on the neural network, explicitly defining:
\begin{equation}
\tilde\Theta_t={\theta_t\in \mathrm{NN}(\mathbb{L}_t,\mathbb{W}_t,\mathbb{S}_t,\mathbb{B}_t,\mathbb{E}_t,\lambda_t):\mathbb{R}^{d{x_t}+d{h_t}+1}\rightarrow \mathbb{R}^{d{x_t}}},
\end{equation}
where this set represents a ReLU neural network analogous to equation~\eqref{p-space}, supplemented by an additional Hölder-norm bound :
\begin{equation}
\lambda_t=\sup_{\x \in\mathbb{R}^{d_x},\z\neq \z',\tau\in [\underline\tau,\overline{\tau}]}
\frac{\|\theta_t(\x,\z',\tau)-\theta_t(\x,\z,\tau)\|_{\infty}}{\|\z-\z'\|^{\alpha_t}_2},
\end{equation}
where $\alpha_t=r_t$ if $r_t\leq 1$ and $\alpha_t=\min(1,r_t-1)$ if $r_t>1$.

\begin{theorem}[Non-transfer conditional diffusion]
\label{cor-nt}
Suppose $\tilde\Theta_t$ has the same configuration as $\Theta_t$ from Theorem \ref{thm_diff} and an additional constraint on $\tilde\Theta_t$: $\lambda_t=c_{\lambda}$ for $r>1$ and $\lambda_t=c_{\lambda}/\sigma_{\underline{\tau}}$ for $r\leq 1$, provided that $c_{\lambda}$ is sufficiently large. Under Assumption \ref{A_p0},  the generation error of the non-transfer conditional diffusion model, adhering to the same stopping criteria from Theorem \ref{thm_diff}, is given by:
\begin{eqnarray*}
\label{c-rate2}
\mathrm{E}_{\smc{D}_t}\mathrm{E}_{\z_t}[\mathrm{TV}(p^0_{\x_t|\z_t},\tilde p_{\x_t|\z_t})]=O( n_t^{-\frac{r_t}{d_{x_t}+d_{h_t}+2r_t}}\log^{m_t} n_t + \varepsilon^h_t), \text{ if } r_t>0; \\
\mathrm{E}_{\smc{D}_t}\mathrm{E}_{\z_t}[\smc{K}^{1/2}(p^0_{\x_t|\z_t},\tilde p_{\x_t|\z_t})]= O( n_t^{-\frac{r_t}{d_{x_t}+d_{h_t}+2r_t}}\log^{m_t} n_t + \varepsilon^h_t), \text{ if } r_t>1.
\end{eqnarray*} 
Here, $\varepsilon^h_t$ is the error rate for estimating $h_t(\z_t)$ as defined by
\eqref{eq-entropy-h} in Lemma \ref{l-composite-entroty}. 
\end{theorem}

The results in Theorems \ref{thm_diff} and \ref{cor-nt} provide valuable insights into conditional generation.

\noindent
\textbf{Dimension reduction via $h(\z)$ and large pre-trained models.} Theorem \ref{thm_diff} indicates that generation accuracy is influenced by the source error $\varepsilon_s$, which depends on the size of the pre-training sample $n_s$. Larger pre-training datasets effectively reduce $\varepsilon_s$. Specifically, using a diffusion model configured similarly for the source task yields $\varepsilon_s = O(n_s^{-\frac{r_s}{d_{x_s}+d_{h_s}+2 r_s}}\log^{m_s} n_s + \varepsilon_s^h)$, as derived in Lemma \ref{thm_diff_source}, where $r_s$ denotes the smoothness degree of $p_{\x_s|\z_s}$ and $m_s = \max(\frac{19}{2}, \frac{r_s}{2}+1)$. When pre-training models are significantly large such that $n_s$ substantially exceeds $n_t$, the source error $\varepsilon_s$ becomes minimal, making the term $n_t^{-\frac{r_t}{d{x_t}+d{h_t}+2r_t}}\log^{m_t} n_t$ the dominant factor influencing the accuracy of the transfer model. This rate generally surpasses the conventional estimation rate $n_t^{-\frac{r_t}{d_{x_t}+d_{z_t}+2r_t}}$ for conditional densities over $(\x_t,\z_t)$ within $[0,1]^{d{x_t}+d{z_t}}$, as indicated in \citep{shen1994convergence, wong1995probability}, assuming a smoothness degree of $r_t$. This improvement occurs because the effective dimension $d_{x_t}+d_{h_t}$ is smaller than $d_{x_t}+d_{z_t}$, attributed to the compact latent representation $h_t(\z_t)$ ($d{h_t} < d_{z_t}$). Thus, $\varepsilon_t$ achieves an accelerated convergence rate through (nonlinear) dimensionality reduction.

\noindent\textbf{Transfer versus nontransfer conditional diffusion.} Building on Theorem\ref{cor-nt}, we contrast the excess--risk behavior of transfer and non\textendash transfer conditional diffusion models to quantify the benefit of pre\textendash training. When extensive source data yield a high-quality latent representation $\hat h$, the target learner can regard $\hat h$ as fixed, thereby reducing the effective complexity of the target problem. Specifically, the excess risk of the transfer model satisfies whereas the nontransfer counterpart incurs with $\varepsilon_s \ll \varepsilon^{h}_t$ whenever the source pre-training set is large. Thus the leading term of the transfer bound is strictly smaller, formalizing the systematic performance gains achievable with transfer learning in data rich pre-training regimes. When the source and target tasks are weakly related, pre-training may instead degrade performance, a phenomenon known as negative transfer. Section~\ref{sec: numerical} demonstrates both the gains and the hazards via a controlled numerical study. In practice, simple model-selection heuristics such as cross-validation can reliably detect and alleviate negative transfer \citep{HuZhang2023OptimalTransfer}.

\noindent \textbf{Connection to \cite{fu2024unveil} for non-transfer conditional generation.} The study by \cite{fu2024unveil} investigates conditional diffusion generation without transfer learning with the density $p_{\x_t|z_t}$ over dimension $d_{x_t}$ and sample size $n_t$. They establish a TV-norm rate $O(n_t^{-\frac{r_t}{d_{x_t}+d_{z_t}+2r_t}}\log^{m_t}n_t)$ under their Assumption 3.3, paralleling Assumption \ref{A_p0} but on $\mathbb{R}^{d_{x_t}} \times [0,1]^{d_{z_t}}$ rather than a manifold $\mathbb{R}^{d_{x_t}} \times [0,1]^{d_{h_t}}$. This rate aligns with the minimax rate presented in \cite{oko2023diffusion} for unconditional generation, excluding a logarithmic term. Conversely, Theorem \ref{cor-nt} describes a KL divergence rate of $O(n_t^{-\frac{r_t}{d_{x_t}+d_{h_t}+2r_t}})$, incorporating a logarithmic factor for non-transfer conditional generation when $r_t > 1$. Given that KL divergence provides a stronger metric compared to the TV-norm by Pinsker's inequality, the reduced dimension $d_{x_t}+d_{h_t}$ (generally smaller due to $d_{h_t}< d_{z_t}$ and the nonlinear manifold structure $h(\z_t)$) enhances convergence. Thus, leveraging latent representations guided by SEC typically leads to accelerated convergence rates.

\subsection{Unconditional diffusion via transfer learning\label{sec_3-3}}

For an unconditional generation of $\U$, we use the diffusion model to learn the latent distribution of $\U$ from the target data $\{\bm{u}_s^i\}_{i=1}^{n_s}$, with $\X(0)=\U$ in
\eqref{forward}. Then, the empirical score matching loss in \eqref{loss_2} $L_{u}$ is
%\begin{align*}
$L_{u}(\theta_u)
=\sum_{i=1}^{n_s} \int_{\underline{\tau}}^{\overline{\tau}} \mathrm{E}_{\x(\tau)|\x(0)}\|\nabla \log p_{\x(\tau)|\x(0)}(\X(\tau)|
\bm{u}_s^i)-\theta_u(\X(\tau),\tau)\|^2 \mathrm{d}\tau$,
%\end{align*}
where $\theta_u\in \Theta_{u}=\mathrm{NN}(\mathbb{L}_u,\mathbb{W}_u,\mathbb{S}_u,\mathbb{B}_u,\mathbb{E}_u)$.  
The estimated score function for the target task is
$\hat{\theta}_u=\argmin_{\theta_u\in\Theta_{U}}L_{U}(\theta_u)$.
Then, we set $l_{g_t}(\bm u,\x;g_t)$ as the reconstruction squared $L_2$ loss and minimize the loss to yield
$
\hat{g}_{t}=\argmin_{g_{t}\in \Theta_{g_{t}}}\sum_{i=1}^{n_t} l_{g_t}(\bm{u}_t^i, \x_t^i;g_{t})=\argmin_{g_{t}\in \Theta_{g_{t}}}\sum_{i=1}^{n_t}\|g_{t}(\bm{u}_t^i)-\x_t^i\|^2_2,
$
where $\Theta_{g_{t}}=\mathrm{NN}(\mathbb{L}_g,\mathbb{W}_g,\mathbb{S}_g,\mathbb{B}_g,\mathbb{E}_g)$. Finally, $P_{\x_t}$ is estimated by $\hat{P}_{\x_t}=\hat{P}_{u}(\hat{g}_{t}^{-1})$.

Next, we introduce specific assumptions for unconditional generation.

\begin{assumption}[Smoothness of $g^0_t$]
\label{A_g}
Assume that $g_t^0 \in \smc{C}^{r_g}(\R^{d_u},
[0,1]^{d_{x_t}},B_g)$ with radius $B_g>0$ and degree of smoothness $r_g>0$.
\end{assumption}

The network class $\Theta_g$ is set with specified parameters: 
\begin{align}
\label{eq-diff-set2}
&\mathbb{L}_g=c_L L\log L, \mathbb{W}_g= c_W W\log W, \mathbb{S}_g= c_S W^2 L\log^2W \log L,\nonumber\\ &\mathbb{B}_g= c_B, \log\mathbb{E}_g= c_E\log(WL).
\end{align}
Here, $c_L, c_W, c_S, c_B, c_E$ are sufficiently large constants, and $(W,L)$ 
are the parameters to control the complexity of the estimator class and dependent on $d_u$, $n_t$ and $r_t$. This configuration allows for flexibility in the network's architecture, enabling the use of either wide or deep structures.

\begin{theorem}[Unconditional diffusion via transfer learning]
\label{thm_ug}
Under Assumptions \ref{A-independent}, \ref{U-error} and \ref{A_g}, with setting \eqref{eq-diff-set2} by $WL\asymp n_t ^{\frac{d_u}{2(d_u+2r_g)}}$,  the generation error for unconditional transfer diffusion models in the Wasserstein distance is
\begin{equation*}
\mathrm{E}_{\smc{D}_t,\smc{D}_s}\smc{W}(P^0_{\x_t}, \hat{P}_{\x_t})=O( n_t ^{-\frac{{r_g}}{d_u+2{r_g}}}  \log^{m_g} n_t+\varepsilon_s^u),
\end{equation*}
where $m_g=\max(\frac{5}{2},\frac{r_g}{2})$.
\end{theorem}

 To compare transfer and non-transfer diffusion generation, we define the non-transfer estimation as 
\[
\tilde{P}_{\x_t} = \tilde{P}_{u}(\hat{g}_{t}^{-1}),
\]
where $\tilde{P}_u$ is estimated using diffusion models characterized by the score function $\tilde{\theta}_u$. Specifically, $\tilde{\theta}_u$ is obtained by substituting the sample set in $L_u$ with $\{\bm{u}_t^i\}_{i=1}^{n_t}$.

\begin{theorem}[Non-transfer via unconditional diffusion]
\label{cor-nt2}
Suppose there exists a sequence $\varepsilon_t^u$ indexed by $n_t$ such that $n_t^{1-\xi}(\varepsilon_t^u)^2\rightarrow\infty$ as $n_s\rightarrow\infty$ and
$P(\rho_u(\theta_u^0,\tilde\theta_u)\geq \varepsilon)\leq \exp(-c_3 n_t^{1-\xi} \varepsilon^2)$ for any $\varepsilon\geq \varepsilon_t^u$ and some constants $c_3, \xi>0$. Under Assumption \ref{A_g} and the same settings for $\Theta_{g_t}$ in Theorem \ref{thm_ug}, 
the generation error of the non-transfer unconditional diffusion model is 
\begin{equation*}
\mathrm{E}_{\smc{D}_t}\smc{W}(P^0_{\x_t}, \tilde{P}_{\x_t})=O( n_t ^{-\frac{{r_g}}{d_u+2{r_g}}}\log^{m_g} n_t+\varepsilon_t^u).
\end{equation*}
\end{theorem}

For the unconditional case, we derive the generation error expressed in the Wasserstein distance $\smc{W}(\hat{P}_{\x},P^0_{\x})=\sup_{\|f\|_{Lip}\leq 1} 
|\int f(\x)(\mathrm{d}\hat{P}_{\x}-\mathrm{d}P^0_{\x})|$, where $\|f\|_{Lip}\leq 1$ indicates that $f$ is within the 1-Lipschitz class. The Wasserstein distance is appropriate for scenarios
with dimension reduction, where the input dimension $d_u$ for $g_t$ is less than the output dimension $d_{x_t}$. In contrast, the KL divergence or the TV-norm may not be appropriate when $g_t$ is not invertible.

To understand the significance of this result in unconditional generation, we explore the following aspects:

\noindent \textbf{Dimension reduction via latent structures $\U$.}
Theorem \ref{thm_ug} shows the error rate for unconditional diffusion generation of density $p_{\x_t}$ is $O(n_t^{-\frac{r_g}{d_u+2r_g}}\log^{m_g} n_t + \varepsilon_s^u)$ with $m_g=\max(\frac{5}{2},\frac{r_g}{2})$, where $\varepsilon_s^u=n_s^{-\frac{r_u}{d_u+2r_u}}\log^{m_u}n_s$ as established in Lemma \ref{thm_pu}, with $m_u=\max(\frac{19}{2},\frac{r_u}{2}+1)$ and $r_u$ denoting the smoothness degree of $p_{\bm u}$.
Given sufficient pre-training data, particularly when $n_s>>n_t$, $\varepsilon_s^u$ becomes negligible, leaving $n_t^{-\frac{r_g}{d_u+2r_g}}\log^{m_g} n_t$ as the primary determinant of accuracy. This rate differs from the $n_t^{-\frac{r_g}{d_{x_t}+2r_g}}$ rate for density estimation of 
$p_{\x_t}$, where $r_g$ represents the smoothness of $p_{\x_t}$ over $[0,1]^{d_{x_t}}$. The improved rate
$\varepsilon_t$ results from dimension reduction since $d_u < d_{x_t}$,  facilitated by the latent structures $\U$,
leading to enhanced generation accuracy relative to methods lacking dimension reduction.

\noindent \textbf{Advantages of transfer learning.} 
Estimating the transferred latent distribution $P_{\bm u}$ from the source task significantly improves generative performance in the target task. 
A comparative analysis reveals a notable difference between transfer and non-transfer models in terms of generation error rates:
$n_t^{-\frac{r_g}{d_u+2 r_g}}\log^{m_g} n_t+\varepsilon_s^u$ for the transfer model and $n_t^{-\frac{r_g}{d_u+2 r_g}}\log^{m_g} n_t+ \varepsilon_t^u$ for the non-transfer model, with $\varepsilon_j^u$; $j \in \{s,t\}$ representing the generation errors in source and target learning, respectively. Here, $\varepsilon_s^u$ is significantly lower than $\varepsilon_t^u$ when the source model is extensively pre-trained, evidenced by $n_s \gg n_t$. 
This highlights the efficiency and effectiveness of transfer learning when leveraging well-prepared source models.

\noindent \textbf{Comparison with \citet{oko2023diffusion} and \citet{chen2023score} in non-transfer unconditional generation.}
The work by \cite{oko2023diffusion} on unconditional diffusion generation for a $d_{x_t}$-dimensional density $p_{\x_t}$ with a sample size $n_t$ sets a TV-norm upper bound at $O(n_t^{-\frac{r_t}{d_{x_t}+2r_t}}\log^{\frac{5d_{x_t}+8r_t}{2d_{x_t}}}n_t)$ with $r_t$ indicating the smoothness
degree of $p_{\x_t}$. This rate, nearly minimax, requires the density $p_{\x_t}$ to be smoothly within a Besov ball $\mathbb{B}^{r}_{p,q}([0,1]^{d_{x_t}})$ and assumes infinite smoothness near boundaries as per Assumption 2.4. Meanwhile, \cite{chen2023score} establishes a TV-norm upper bound at $O(n_t^{-\frac{1}{2(d_{x_t}+5)}})$ for low-dimensional linear subspaces with Lipschitz continuous score functions, a rate slower than the previous and suboptimal when $r_t=2$. In contrast, Theorem \ref{cor-nt2} and Theorem \ref{cor-w} derive a Wasserstein distance bound of $O(n_t^{-\frac{r_g}{d_u+2r_g}}\log^{m_g} n_t+n_t^{-\frac{r_u}{d_u+2r_u}}\log^{m_u} n_t)$. With equal smoothness degrees ($r=r_g=r_u$), this
bound becomes $O(n_t^{-\frac{r_t}{d_u+2r_t}}\log^{\max(\frac{19}{2},\frac{r_t}{2}+1)} n_t)$. Given $d_u<d_{x_t}$ due to dimension reduction, this suggests a faster rate despite different metrics.

\section{Normalizing flows}\label{sec_flows}

\subsection{Coupling}\label{flows-sec1}
Normalizing flows transform a random vector $\X$ into a base vector $\V$ with known density
$p_{\bm v}$, through a diffeomorphic mapping $T(\X)$, which is invertible and differentiable. The composition of these mappings, $T = \phi_K \circ \ldots \circ \phi_1$, with each $\phi_j$ modeled by a neural network, estimates $T$. The density of $\X$ is expressed as $p_{\x}(\x) = p_{\bm v}(T(\x)) \left| \det \frac{\partial T(\x)}{\partial \x} \right|$, with the determinant indicating the volume change under $T$. The maximum likelihood approach is used to estimate $T$, enabling the generation of new $\X$ samples by inverting $T$ on samples from $p_{\bm v}$.

\noindent\textbf{Coupling flows.} Coupling flows partition $\x$ into two parts $\x=[\x_1,\x_2]$. Each flow employs a transformation $\phi_j(\x_1, \x_2) = (\x_1, q(\x_2, \omega_j(\x_1)))$; $j=1,\ldots,K$. The function $q$ modifies $\x_2$ based on the output of $\omega$, where $q$ ensures that $\phi_j$ is invertible.

\noindent\textbf{Conditional coupling flows. }To add an additional condition input of $\z$ to the
coupling layer, we adjust $\phi_j(\x_1, \x_2,\z) = (\x_1, 
{q}(\x_2, \omega(\x_1,\z)))$.

\subsection{Conditional flows via transfer learning}\label{flows-sec2}

This section constructs coupling flows to model the conditional density $p_{\x_t|\z_t}$. We use the three-layer coupling flows defined in \eqref{eq_p1} with $q(\x,\y)=\x+\y$, denoted by
$\mathrm{CF}(\mathbb{L},\mathbb{W},\mathbb{S},\mathbb{B},\mathbb{E},\lambda)$, where $\phi_1$ is defined by a neural network $\omega_1\in \mathrm{NN}_t(\mathbb{L},\mathbb{W},\mathbb{S},\mathbb{B},\mathbb{E},\lambda)$,  with the maximum depth $\mathbb{L}$, the maximum width $\mathbb{W}$, the number of effective parameters $\mathbb{S}$, the supremum norm of the neural network $\mathbb{B}$, the parameter bound $\mathbb{E}$ and the Lipschitz norm of the neural network $\lambda$; $\phi_2$ is a permutation mapping; $\phi_3$ is defined by $\omega_3=-\omega_1^{-1}$. For the target learning task, we define the parameter space as $\Theta_t = \mathrm{CF}_t(\mathbb{L}_t,\mathbb{W}_t,\mathbb{S}_t,\mathbb{B}_t,\mathbb{E}_t,\lambda_t)$.
\begin{align}
\label{l-theta-cp}
\Theta_t=\{\theta_t(\x,\hat{h}_t(\z)):[\phi_3\circ\phi_2\circ\phi_1((\x,\bm 0),\hat{h}_t(\z))]^{1:d_{x_t}}, 
\omega_1\in \mathrm{NN}_t(\mathbb{L}_t,\mathbb{W}_t,\mathbb{S}_t,\mathbb{B}_t,\mathbb{E}_t,\lambda_t)
\}.   
\end{align}
Here, $(\x,\bm 0)$ is a zero padding vector of dimension $2d_{x_t}$ and $[\cdot]^{1:d_{x_t}}$ denotes the first $d_{x_t}$ elements. 
Given a target training sample $\{\x_t^i,\z_t^i\}_{i=1}^{n_t}$ and an estimated latent structures $\hat{h}$ derived from the pre-trained flow model based on
a source training sample $(\x_s^i,\z_s^i)_{i=1}^{n_s}$, we define the loss function as $l_{t}=-\log p_{\x_t|\z_t}(\x_{t},\z_t;\theta_t,\hat{h})$, and then estimate the target flow by minimizing the negative log-likelihood as follows:
\begin{align*}
    \hat{{T}}_t=\hat{\theta}_{t}(\x_t,\hat{h}_t(\z_t))
=\arg\min_{\theta_t\in\Theta_t}\sum_{i=1}^{n_t}-\log p_{\bm v}(\theta_t(\x_t^i,\hat{h}_t(\z_t^i)))-\log \left|\det( \nabla_{\x}\theta_t(\x_t^i,\hat{h}_t(\z_t^i)))\right|.
\end{align*}
Here $\nabla_{\x}f(\x)=\frac{\partial f(\x)}{\partial \x}$.
This allows for conditional generation via flows using  $\x_t=\hat{T}_t^{-1}(\bm v,\z_t)$ and estimating the conditional target density as $\hat{p}_{\x_t|\z_t}(\x,\z) = 
p_{\bm v}(\hat{T}_t(\x,\z)) \left|\det \nabla_{\x}\hat{T}_t(\x,\z) \right|$.

We make several assumptions about the distribution of $\X_t$ given $\Z_t$. 

\begin{assumption}[Transformation] \label{A_BL}
Suppose that there exists $T_t^0(\x,\z)=\theta_t^0(\x,h_t^0(\z))$ such that $\V=T_t^0(\X_t,\Z_t)$.
The true transform $\theta_t^0(\x_t,h_t^0(\z_t))$ and its inverse belong to a H\"older ball $\smc{C}^{r_t+1}([0,1]^{d_{x_t}+d_{h_t}},[0,1]^{d_{x_t}}, B_t)$ of radius $B_t>0$, while $|\det \nabla_{\x}\theta^0_t|$ is lower bounded by some positive constant. Moreover, the base vector $\bm V$ has a known smooth density in $\smc{C}^{\infty}([0,1]^{d_{x_t}},\R,B_v)$ with a lower bound.
\end{assumption}

This condition, a generalized version of the bi-Lipschitz condition used in \cite{jin2024approximation}, enables the constructed invertible network to approximate $T^0$ while satisfying invertibility. 
The smoothness condition is critical for controlling the approximation error associated with the Jacobian matrix during the approximation process. 

The network class $\Theta_t$ is set with specified parameters: 
\begin{align}
\label{eq-cp-set}
&\mathbb{L}_t = c_L L\log L, \mathbb{W}_t = c_W W\log W, \mathbb{S}_t = c_S W^2 L\log^2W \log L,\nonumber\\ &\mathbb{B}_t = c_B, \log\mathbb{E}_t = c_E\log(WL),\lambda_t = c_{\lambda}.
\end{align}

To derive Theorems \ref{thm_flow}--\ref{cor-cp-nt2}, we address essential technical challenges related to constructing invertible neural networks for flow approximation, as well as those previously outlined for diffusion models. Furthermore, we capitalize on the distinct structural properties inherent to flow models.

\begin{theorem} [Conditional flows via transfer learning]
\label{thm_flow}
Under Assumptions \ref{A-independent}-\ref{A-error} and \ref{A_BL}, with $WL\asymp n_t^{\frac{d_{x_t}+d_{h_t}}{2(d_{x_t}+d_{h_t}+2r_t)}} $ in the setting \eqref{eq-cp-set}, the generation error of conditional transfer flow models is  
\begin{equation*}
\mathrm{E}_{\smc{D}_t,\smc{D}_s}\mathrm{E}_{\z_t}[\smc{K}^{1/2}(p^0_{\x_t|\z_t},\hat p_{\x_t|\z_t})]=O(n_t^{-\frac{r_t}{d_{x_t}+d_{h_t}+2r_t}}\log^{\frac{5}{2}}  n_t+\varepsilon_s).
\end{equation*}
\end{theorem}

This theorem introduces the first results delineating the generation accuracy bounds for flow models, particularly in the context of transfer learning. It extends the existing approximation literature for invertible neural networks, as detailed in \cite{jin2024approximation} on approximation. Importantly, we establish an error bound that simultaneously addresses the mapping and the Jacobian matrix.  Although these results are theoretically significant, practical implementation of such neural networks with constraints as per \eqref{eq-linear-con} may require considerable effort, as discussed in the appendix.

Regarding the non-transfer case, the transformation function $T$ is estimated by
$
 \tilde{\theta}_{t}(\x_t,\tilde{h}_t(\z_t))=\arg\min_{\theta_t\in \Theta_t,h\in \Theta_h}\sum_{i=1}^{n_t} -\log p_{\x_t|\z_t}(\x_t^i,\z_t^i;\theta_t;h).
$

\begin{theorem}[Non-transfer conditional flows]
\label{cor-cp-nt}
Under Assumption \ref{A_BL} and the same configurations in Theorem \ref{thm_flow}, the generation error of the non-transfer conditional flow model is
\begin{equation*}
\mathrm{E}_{\smc{D}_t}\mathrm{E}_{\z_t}[\smc{K}^{1/2}(p^0_{\x_t|\z_t},\tilde p_{\x_t|\z_t})]= O(n_t^{-\frac{r_t}{d_{x_t}+d_{h_t}+2r_t}}\log^{\frac{5}{2}}n_t + \varepsilon^h_t ).
\end{equation*}
Here, $\varepsilon^h_t$ is the error rate for estimating $h_t(\z_t)$ as defined by
\eqref{eq-entropy-h} in Lemma \ref{l-composite-entroty}. 
\end{theorem}

\noindent\textbf{Conditional flow generation: transfer versus non-transfer models.}
The conditional flow generation rate via transfer learning is $n_t^{-\frac{r_t}{d_{x_t}+d_{h_t}+2r_t}}\log^{5/2} n_t + \varepsilon_s$. For the 
pre-trained (source) flow model, $\varepsilon_s$ is expressed as $n_s^{-\frac{r_s}{d_{x_s}+d_h+2r_s}}\log^{5/2} n_s + \varepsilon^h_s$, by Lemma \ref{thm_floW_source}, where $\varepsilon^h_s$ satisfies the integral entropy condition in Lemma \ref{l-composite-entroty}. In comparison to the non-transfer model's error bound of $n_t^{-\frac{r}{d_{x_t}+d_h+2r}}\log^{5/2} n_t + \varepsilon^h_t$, the transfer flow model exhibits superior performance in contexts where the latent structure complexity is substantial and the source sample size $n_s$ is considerably larger than the target sample size $n_t$, such that $\varepsilon^h_s \ll \varepsilon^h_t$.

\subsection{Unconditional flows via transfer learning}\label{flows-sec3}

In unconditional generation, we estimate the latent distribution $p_u$ using coupling flows
$\Theta_u=\mathrm{CF}_u(\mathbb{L}_u,\mathbb{W}_u,\mathbb{S}_u,\mathbb{B}_u,\mathbb{E}_u,\lambda_u)$, based on a source sample $\{\bm{u}^i_s\}_{i=1}^{n_s}$. The model $\theta_u$ is obtained by: 
\begin{align*}
    \hat\theta_u=\arg\min_{\theta_u\in \Theta_u}\sum_{i=1}^{n_s} l_u(\bm{u}_s^i;\theta_u) 
    =\arg\min_{\theta_u\in \Theta_u}
    \sum_{i=1}^{n_s}-\log p_{\bm v}(\theta_u(\bm{u}_s^i))-\log \left|\det\nabla\theta_u(\bm{u}_s^i)\right|.
\end{align*}
Then, we minimize the reconstruction error loss on a target sample $\{\bm{u}_t^i,\x_t^i\}_{i=1}^{n_t}$ to obtain
$
\hat{g}_t=\argmin_{g_t\in\Theta_{g_t}}\sum_{i=1}^{n_t}\|g_t(\bm{u}_t^i)-\x_t^i\|^2_2,
$
where $\Theta_{g_t}=\mathrm{NN}_g(\mathbb{L}_g,\mathbb{W}_g,\mathbb{S}_g,\mathbb{B}_g,\mathbb{E}_g)$.
The estimation for $P_{\x_t}$ can be derived as $\hat{P}_{\x_t}=\hat{P}_u(\hat{g}_t^{-1})$. Next, we restate Assumption \ref{A_g} for the true function $g^0_t$. 

\begin{assumption}[Smoothness of $g^0_t$]
\label{A_g2}
Assume that $g_t^0 \in \smc{C}^{r_g}([0,1]^{d_u},
[0,1]^{d_{x_t}},B_g)$ with radius $B_g>0$ and degree of smoothness $r_g>0$.
\end{assumption}

The network class $\Theta_g$ is set with specified parameters: 
\begin{align}
\label{eq-cp-set2}
&\mathbb{L}_g=c_L L\log L, \mathbb{W}_g= c_W W\log W, \mathbb{S}_g= c_S W^2 L\log^2W \log L, \nonumber\\
&\mathbb{B}_g= c_B, \log\mathbb{E}_g= c_E\log(WL).
\end{align}
\begin{theorem} [Unconditional flows via transfer learning]
\label{thm_flow2}
Under Assumptions \ref{A-independent}, \ref{U-error} and \ref{A_g2}, with setting \eqref{eq-cp-set2} by $WL\asymp n_t ^{\frac{d_u}{2(d_u+2r_g)}}$,
the error for unconditional flow generation via transfer learning is,
\begin{equation*}
\mathrm{E}_{\smc{D}_t,\smc{D}_s}\smc{W}(P^0_{\x_t}, \hat{P}_{\x_t})=O(n_t^{-\frac{r_g}{d_u+2r_g}}\log^{\frac{5}{2}} n_t+\varepsilon_s^u).
\end{equation*}
\end{theorem} 

In the non-transfer case, we define the estimation $\tilde p_u$ with $\tilde \theta_u$ using the target data $\{\bm{u}_t^i\}_{i=1}^{n_t}$. The distribution estimation for $\X_t$ is given by $\tilde P_{\x_t}=\tilde P_{u}(\hat{g}_t^{-1})$.

\begin{theorem}[Non-transfer unconditional flows]
\label{cor-cp-nt2}
Suppose there exists a sequence $\varepsilon_t^u$ indexed by $n_t$ such that $n_t^{1-\xi}(\varepsilon_t^u)^2\rightarrow\infty$ as $n_s\rightarrow\infty$ and
$P(\rho_u(\theta_u^0,\tilde\theta_u)\geq \varepsilon)\leq \exp(-c_3 n_t^{1-\xi} \varepsilon^2)$ for any $\varepsilon\geq \varepsilon_t^u$ and some constants $c_3, \xi>0$. Under Assumption \ref{A_g2} and the same configurations of Theorem \ref{thm_flow2}, 
  the error in non-transfer flow generation is 
  \begin{equation*}
\mathrm{E}_{\smc{D}_t}\smc{W}(P^0_{\x_t},\tilde{P}_{\x_t})= O(n_t^{-\frac{r_g}{d_u+2r_g}}\log^{\frac{5}{2}} n_t+\varepsilon_t^u).
 \end{equation*}
\end{theorem}

\noindent{\bf Comparison of diffusion and flows}

\noindent \textbf{Generation accuracy. }Flow models learn probability densities by transforming the target distribution to a known base distribution. On the other hand, Gaussian diffusion models learn the target density by bridging the standard Gaussian and target distributions through score matching. Both approaches utilize round-trip processes and achieve similar error rates in both conditional and unconditional generation, as discussed in Theorems \ref{cor-nt}, \ref{cor-nt2}, \ref{cor-cp-nt} and \ref{cor-cp-nt2}, except for the evaluation metrics and network architectures.

\noindent \textbf{Limit of assumptions. }These models require specific assumptions about the target density. Notably, coupling flows operate under slightly less stringent conditions than those for diffusion models. The assumptions for diffusion models primarily relate to the complexities of matching the score function and the necessity of approximating the target density by its smoothed version by Gaussian noise.

\noindent \textbf{Network architecture.} Theorem \ref{thm_diff} suggests that diffusion generation requires a wide network for sufficient approximation. In contrast, Theorem \ref{thm_flow} indicates that the architecture for flow models can vary in either network depth or width, offering greater flexibility. However, designing flows for specific tasks can be computationally challenging.

\section{Core proof strategy}
\label{sec-general}

The proofs for our main theoretical results must overcome three principal technical hurdles:

\textbf{Propagating error from source to target under the SEC condition.} We first derive a model-agnostic risk-decomposition bound that requires no distributional assumptions. It splits the target risk into three components: (i) the source error, (ii) an approximation error, and (iii) an estimation error. This result hinges on extending the classical convergence theorem of  \cite{shen1994convergence} to the conditional setting needed for transfer learning.

\textbf{Controlling approximation and estimation errors in diffusion models.} Adopting the Gaussian-control framework of \cite{fu2024unveil}, we inherit their guarantees on the score-matching loss, which recover the known total-variation bounds. To upgrade these guarantees to Kullback–Leibler divergence, we perform a refined analysis of the tail segment ($[0,\underbar{t}]$) of the reverse diffusion, converting score-matching accuracy into KL guarantees.

\textbf{ Simultaneously approximating mappings and their derivatives in flow models.} We introduce two key tools: (i) a new coupling-structure argument that links errors in function values and gradients, and (ii) the simultaneous-approximation theory for ReQU-activated neural networks of \cite{belomestny2023simultaneous}, which provides joint control over both approximation and derivative errors.

\subsection{Theory for transferred risk control}

This section establishes a theoretical framework to quantify the excess risk associated with estimating high-dimensional distributions that lie on lower-dimensional manifolds, both in the context of transfer learning and its non-transfer counterpart. This setting significantly differs from classical supervised transfer learning frameworks due to the potential degeneracy inherent in high-dimensional distributions. Leveraging the unique structural properties of diffusion and flow models, we derive generation error bounds directly linked to the excess risk in estimation accuracy. We specify precise conditions regarding the variance, sub-Gaussian properties of the loss functions, and the mechanisms governing knowledge transfer between source and target tasks.

As mentioned in Section \ref{sec:enhancing}, we parametrize the loss $l_j$ by $\gamma_j=(\theta_j,h) \in \Gamma_j=\Theta_j\times\Theta_h$ for conditional generation; $j \in \{s,t\}$. This setup includes task-specific parameters $\theta_j$'s and shared latent parameters $h$. In what follows,
$\E$ and $\Var$ represent the expectation and variance concerning the associated randomness. We make the following assumptions.

\begin{assumption}(Variance) 
\label{Variance} 
There exist constants $c_{vj}>0$ such that for all small
$\varepsilon>0$,
$$
 \sup_{\{\rho_j(\gamma^0_j,\gamma_j) \leq
\varepsilon: \gamma_j \in \Gamma_j\}}{\rm Var} (l_j(\cdot;\gamma_j)
-l_j(\cdot;\gamma^0_j)) \leq c_{vj}\varepsilon^2; \quad j \in \{s,t\}, 
$$
where $\rho_j^2(\gamma_j^0,\gamma_j)=\mathrm{E} [l_j(\cdot;\gamma_j)-l_j(\cdot;\gamma^0_j)]$ is the excess risk; $j \in \{s,t\}$.
\end{assumption}

This assumption specifies a connection between the variance and the mean of the loss function.

The assumption next ensures that the loss function exhibits an exponential tail behavior.
A random variable from a function class $\cal F$ is said to satisfy Bernstein's condition with parameter $b>0$ if
$
\sup_{f \in {\cal F}}\E(|f(\X) - \E[f(\X)]|^k) \leq \frac{1}{2} k! v b^{k-2}
$
and $k \geq 2$, where $\sup_{f \in \cal F} \Var(f(X)) \leq v^2$.

\begin{assumption}(Bernstein)
\label{sub-Gaussian} 
Assume that $l_j(\cdot;\gamma_j) -l_j(\cdot;\gamma_j^0)$ satisfies Bernstein's condition
for $\gamma_j \in \Gamma_j$ with parameter $0<c_{bj} <4 c_{vj}$; $j \in \{s,t\}$,
where $c_{vj}$ is defined in Assumption \ref{Variance}.
\end{assumption}

 To measure the complexity of a function class, we define the bracketing 
$L_2$-metric entropy of $\smc{F}=\{f\}$. For any $u>0$, a finite set of function pairs ${(f_{j}^{L},f_{j}^{U})}_{j=1}^{N}$ constitutes a $u$-bracketing of $\smc{F}$ if, for each $j=1,\ldots,N$, the condition $[\mathrm{E}(f_{j}^{L}(\cdot)-f_{j}^{U}(\cdot))^2]^{1/2}\leq u$ is satisfied. Furthermore, for any function $f \in \smc{F}$, there exists an index $j$ such that $f_{j}^{L} \leq f \leq f_{j}^{U}$. The bracketing $L$-metric entropy of $\smc{F}$, denoted as $H_B(\cdot,\smc{F})$, is then defined by the logarithm of the cardinality of the smallest-sized $u$-bracketing of $\smc{F}$.

Next, we derive some results for the transfer and non-transfer generation error under the conditional SEC setting. Let $\delta_j(h)=\inf_{\gamma_j=(\theta_j,h): \theta_j \in \Theta_j} \rho_j^2(\gamma_j^0,\gamma_j)$ represent the approximation error given the true shared parameter $h$. Then $\delta_j(h^0)$ renders as an upper bound for the overall approximation error for the source and target tasks when $h^0\in\Theta_h$ and $j \in \{s,t\}$. Define $\smc{F}_s=\{l_s(\cdot;\gamma_s)-l_s(\cdot;\pi\gamma^0_s):
\gamma_s \in\Gamma_s\}$ and $\smc{F}_{t}(h)=\{l_t(\cdot;\gamma_t)-l_t(\cdot;\pi\gamma^0_t): \gamma_t=(\theta_t,h), \theta_t \in\Theta_t\}$ as candidate loss function classes induced by the parameters, where $\pi \gamma^0_j \in \Gamma_j$ is an approximate point of $\gamma^0_j$ within $\Gamma_j$.

Theorem \ref{theorem1} establishes the excess risk associated with estimating high-dimensional distributions 
in the context of transfer learning.

\begin{theorem}[Transfer Learning]
\label{theorem1}
Under Assumptions \ref{A-independent}, \ref{transferability}, \ref{Variance}, and \ref{sub-Gaussian}, the error $\varepsilon_s$ and $\varepsilon_t$ satisfy: $\varepsilon_s\geq\sqrt{2\delta_s(h^0)}$, 
and $\varepsilon_t \geq \sqrt{2\delta_t(h^0)}$, as well as the following entropy bounds:
\begin{eqnarray}
\label{entropy-t3}
\int_{k_s\varepsilon_s^2/16}^{4 c^{1/2}_{vs} \varepsilon_s}
H_B^{1/2}(u,\smc{F}_s) du \leq c_{h_s} n_s^{1/2} \varepsilon_s^2, \text{ and }
\int_{k_t\varepsilon_t^2/16}^{4 c^{1/2}_{vt} \varepsilon_t}
\sup_{h\in \Theta_h}H_B^{1/2}(u,\smc{F}_t(h)) du \leq c_{h_t} n_t^{1/2} \varepsilon_t^2,
\end{eqnarray}
where $k_j$ and $c_{h_j}$ are constants depending on $c_{vj}$ and $c_{bj}$ for $j = s, t$. Then, the error bound for the target generation via $\hat{\gamma}_t$ is given by:
\begin{equation}
\label{bound-t3}
P\left(\rho_t(\gamma_t^0, \hat{\gamma}_t) \geq x(\varepsilon_t + \sqrt{3c_1} \varepsilon_s)\right) \leq \exp(-c_{e_t} n_t (x\varepsilon_t)^2) + \exp(-c_{e_s} n_s (x\varepsilon_s)^2),
\end{equation}
for any $x\geq 1$ and some positive constants $c_{e_s}$ and $c_{e_t}$. This implies that $\rho_t(\gamma_t^0, \hat{\gamma}_t) = O_p(\varepsilon_t + \varepsilon_s)$ provided that $\min(n_t \varepsilon_t^2, n_s \varepsilon_s^2) \rightarrow \infty$ as $\min(n_s, n_t) \rightarrow \infty$.
\end{theorem}

The target generation error errors $\varepsilon_t$ and $\varepsilon_s$, established in Theorem \ref{theorem1}, conform to the entropy constraints detailed in \eqref{entropy-t3} and are subject to the lower bounds $\varepsilon_j \geq \sqrt{2\delta_j(h^0)}$, reflecting the variance-bias trade-off inherent in a learning process. Notably, due to the utilization of a large pre-trained model,  $\varepsilon_s$ is significantly less than $\varepsilon_t$, making $\varepsilon_t$ the predominant factor in transfer learning scenarios. Moreover, Theorem \ref{non-transfer-c} suggests that the generation error in transfer learning is relatively smaller than its non-transfer counterpart, owning to shared parameters learned from the source task. The shared learning facilitates dimension reduction, thus enhancing the generation accuracy.

In the absence of knowledge transfer, let $\tilde \gamma_t=\argmin_{\gamma_t\in \Gamma_{t}} L_{t}(\gamma_t)$ be the counterpart of $\hat{\gamma_t}$. Define the function class $\tilde{\smc{F}}_t=\{l(\cdot;\gamma_t)-l(\cdot;\pi \gamma^0_t)):\gamma_t\in \Gamma_{t}\}$, where $\pi \gamma^0_t \in \Gamma_t$ is an approximating point of $\gamma^0_t$ within $\Gamma_t$.

 Theorem \ref{theorem2} establishes the excess risk associated with estimating high-dimensional distributions 
in the context of non-transfer learning.

\begin{theorem}[Non-transfer] 
\label{theorem2}
Under Assumptions \ref{Variance}--\ref{sub-Gaussian}, let $\tilde\varepsilon_t$ satisfy the conditions
\begin{equation}
\label{entropy-t4}
\int_{k_t\tilde\varepsilon_t^2/16}^{4 c^{1/2}_{vt} \tilde\varepsilon_t}
H_B^{1/2}(u,\tilde{\smc{F}}_t) du \leq c_{h_t} n_t^{1/2} \tilde\varepsilon_t^{2},
\quad \tilde\varepsilon_t \geq \sqrt{2 \delta_t(h^0)}.
\end{equation}
Then, the probability bound for target generation is as follows:
\begin{align}
\label{bound-t4}
 P\Big(\rho(\gamma_t^0,\tilde\gamma_t) \geq  
\tilde\varepsilon_t\Big) \leq
 \exp(-c_{e_t} n_t \tilde \varepsilon_t^2),
\end{align}
for some constant $c_{e_t}>0$. 
\label{non-transfer-c}
\end{theorem}

Theorem \ref{non-transfer-c} demonstrates a situation where the generation error of a transfer model remains lower compared to its non-transfer counterpart. This situation is evident when the source training size $n_s$ is significantly higher than that of the target $n_t$. Consequently, leveraging the shared representations for the source and target in transfer learning leads to a lower generation error than the non-transfer approach.

\section{Numerical experiments}
\label{sec: numerical}
\subsection{Simulations}

We conduct simulation studies to demonstrate the effectiveness of transfer generative learning and validate our theoretical findings. Two simulation models are considered for the conditional and unconditional cases:

\noindent\textbf{Conditional generation.} The source variable is defined as $X_s = \sin{Z_1} + \cos{Z_2} + Z_3^2 + e_s$, where $\bm{Z} = (Z_1, \ldots, Z_3) \sim \text{Unif}(-2,2)$ is a random vector, and \( e_s \sim N(0,1) \) is independent noise. The target variable is $X_t = \sin{Z_1} + \cos{Z_2} + Z_3^2 + \exp{e_t}$, where $e_t \sim \text{Unif}(-1,1)$ is independent noise.

\noindent  \textbf{Unconditional generation.} The source variable is $X_s=(\sin{U_1} + \cos{U_2}, U_1^2 + U_2^2, \tanh{(U_1U_2)},\\
\exp{(U_1 - U_2)},\log(|U_1|+1)+\log(|U_2|+1))$, where $\bm{U} = (\sin \varepsilon_1, \cos \varepsilon_2)$ and  $\varepsilon_j \sim N(0,1) $ for $j = 1, 2 $. The target variable is $X_t = (\sin{U_1} + \tanh{U_2}, U_1^2 + U_2, \exp{(U_1 - U_2)})$.

\begin{figure}[H]
\centering
  \includegraphics[width=0.4\textwidth]{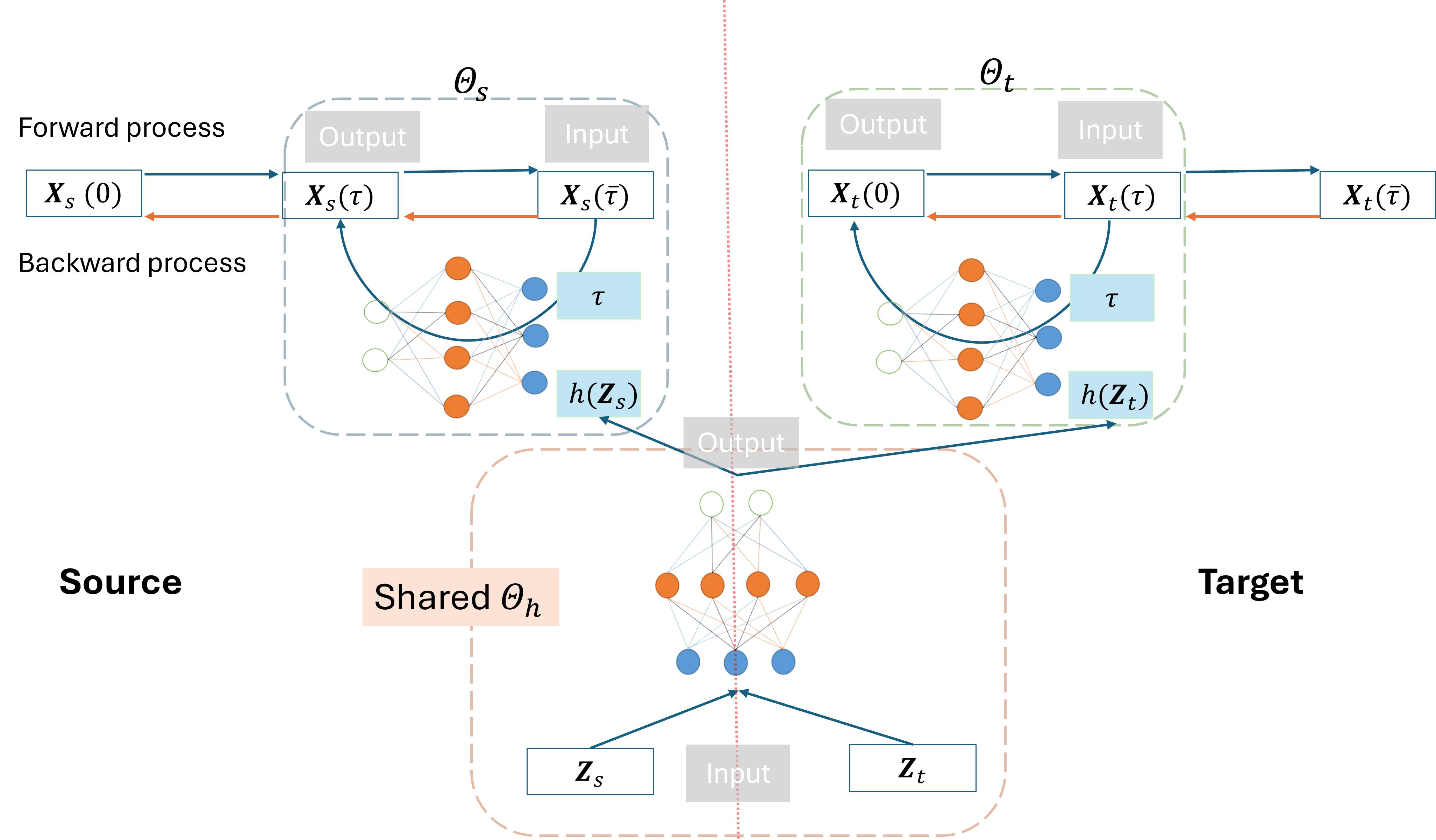}
\caption{Shared architecture for conditional diffusion generation. A common backbone $\Theta_h$ is first trained on the source data and subsequently fed into the target diffusion model $\Theta_t$ to transfer knowledge.}
\label{cg}
\end{figure}

\begin{figure}[H]
\centering
  \includegraphics[width=0.4\textwidth]{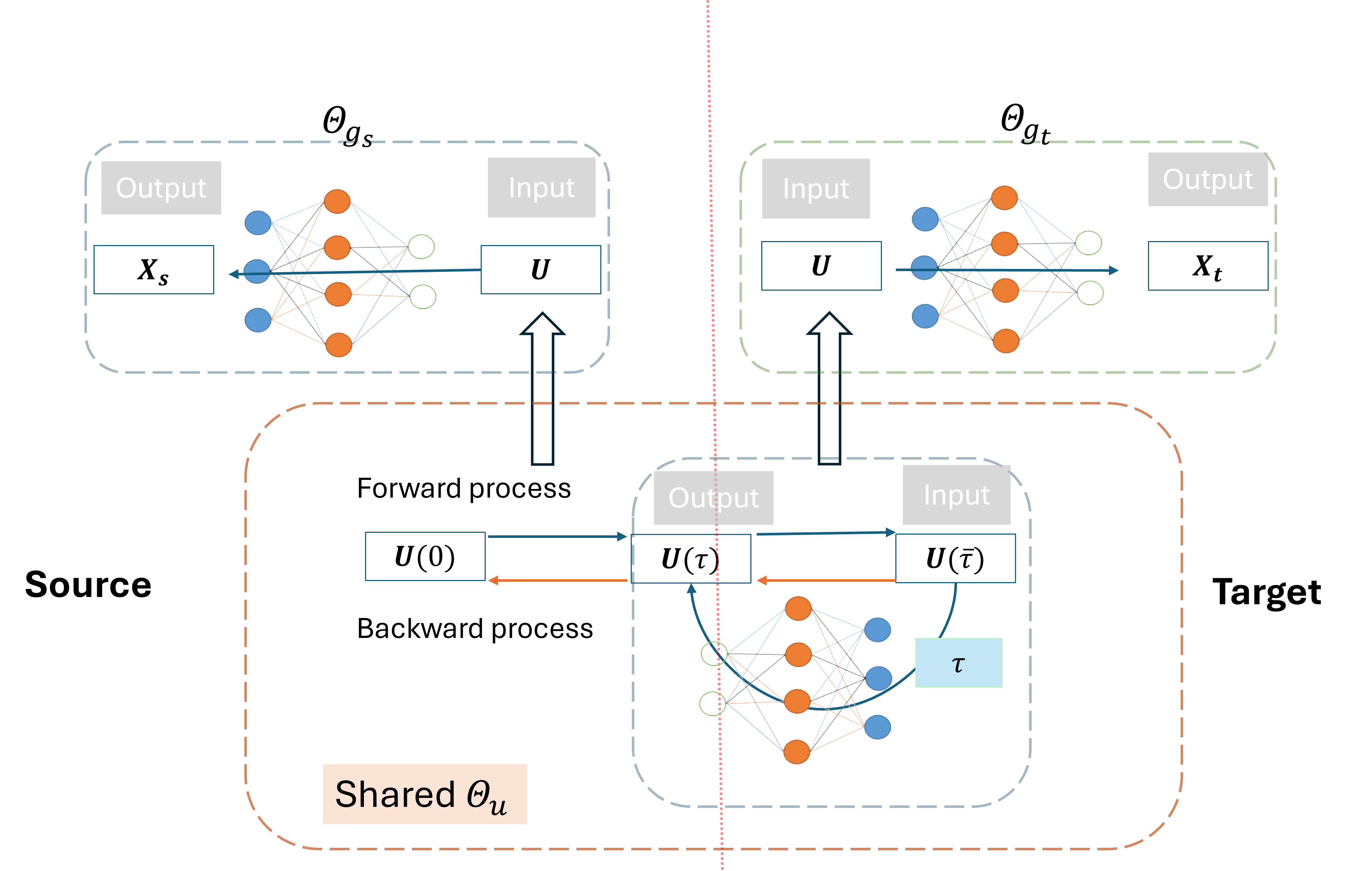}
\caption{Shared architecture for unconditional diffusion generation. A common backbone $\Theta_u$ is first trained on the source data and subsequently fed into the target diffusion model $\Theta_u$ to transfer knowledge.}
\label{ucg}
\end{figure}

We employ diffusion models to learn the underlying distribution and generate synthetic samples closely following the original data's distribution. For the conditional task, the parameters $\Theta_t$, $\Theta_s$, and $\Theta_h$ are implemented as feedforward neural networks with three hidden layers, each comprising 128 units. For the unconditional task, $\Theta_u$ and $\Theta_{g_t}$ are set with the same NN architecture in the conditional models. The detailed frameworks for both cases are illustrated in Figures \ref{cg} and \ref{ucg}.

\vspace{-5mm}

\vspace{-5mm}

\vspace{-5mm}
To illustrate the impact of the source data size $n_s$ on the accuracy of target generation, we select $n_s = \lfloor\exp(i)\rfloor$ for $i \in \{8.0, 8.5, 9.0, 9.5, 10.0, 10.5, 11.0\}$, while keeping the target sample size $n_t$ fixed at $5,000$. In this situation, the source and target simulation models differ in error structures. To compute the TV-norm between raw and synthetic samples, we approximate the empirical density via binning and then sum half the absolute differences between the approximated empirical densities. For the Wasserstein distance, we apply the Sinkhorn algorithm \citep{cuturi2013sinkhorn} which calculates the distributional distance by solving an optimal transport problem using suitable metrics.

\begin{figure}[H]
\centering
  \includegraphics[width=.6 \textwidth]{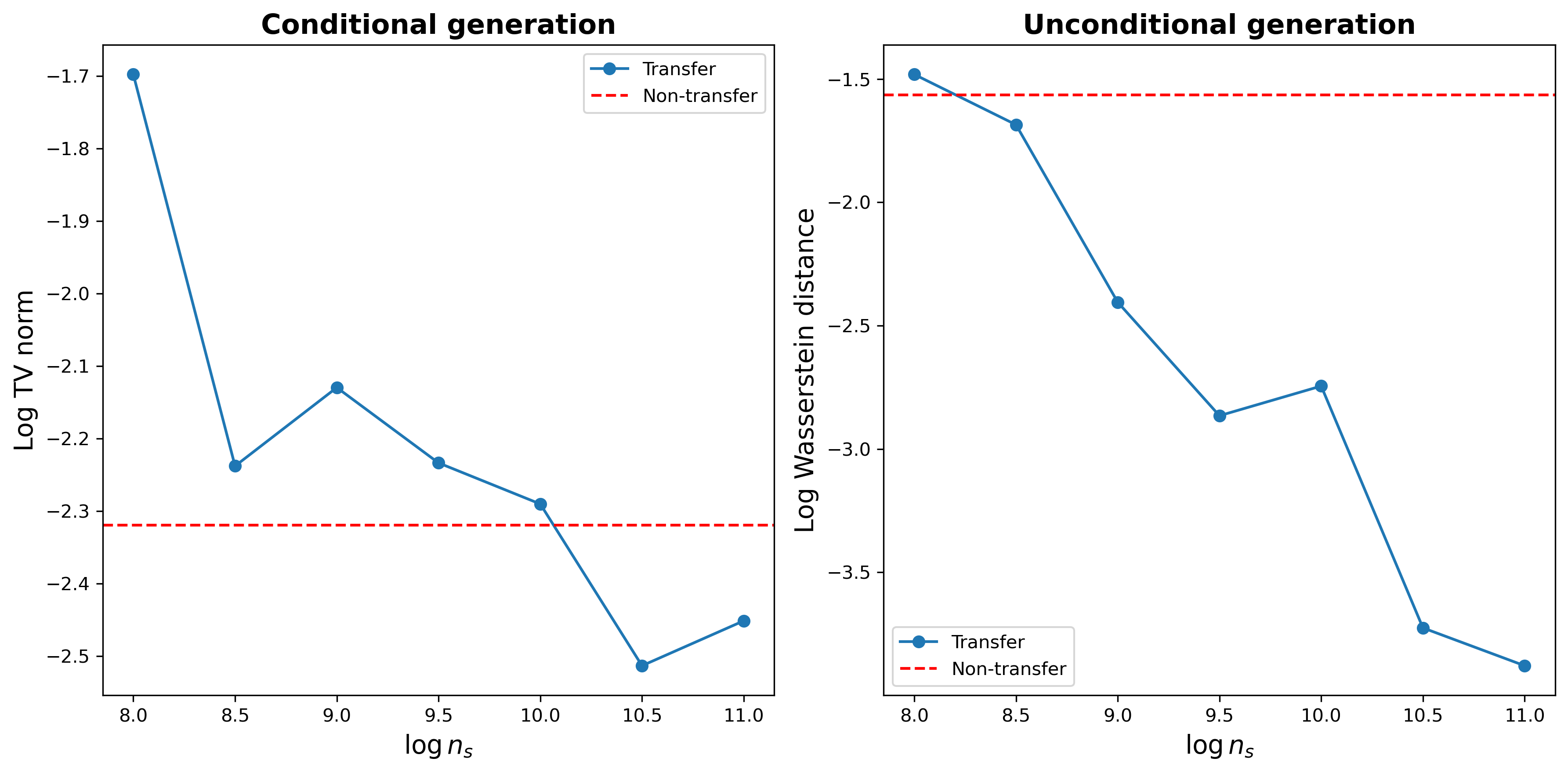}
\caption{Log-scale generation errors as a function of source sample size $n_s$, with the target sample size fixed at $n_t = 5,000$. The left panel illustrates generation errors in the TV-norm for conditional generation, while the right panel shows generation errors in the Wasserstein distance for unconditional generation. The red dash line represents the errors associated with non-transfer methods.}
\label{pic1}
\end{figure}

As shown in Figure \ref{pic1}, generation error for both conditional and unconditional transfer diffusion models declines as the source sample size $n_s$ increases, ultimately outperforming their non‑transfer counterparts once $n_s$ passes a modest threshold. Although small $n_s$ can occasionally trigger negative transfer, enlarging $n_s$, particularly when leveraging large pre-trained models, consistently yields positive transfer. In practice, negative transfer can be guarded against via cross‑validation. These empirical trends closely mirror the theoretical guarantees of Theorems \ref{thm_diff} and \ref{thm_ug} and reinforce the interpretations of Theorems \ref{cor-nt} and \ref{cor-nt2}.

\subsection{Benchmark example: MNIST–USPS Digit Images}
\label{benchmarl}

We investigate image generation on the MNIST–USPS benchmark \citep{carlucci2019hallucinating}, a challenging transfer-learning task because the two handwritten-digit corpora differ substantially in resolution, stroke style, and intra-class variability. We study both conditional and unconditional generation. The experiment details are given in Appendix~\ref{appendix-B}.

\textbf{Conditional generation.}  
We use the MNIST dataset with varying training sample sizes, \(n_s \in \{1{,}000,\;5{,}000,\;10{,}000,\;20{,}000,\;35{,}000,\;60{,}000\}\), to train a UNet model from the \texttt{Diffusers} library, augmented with a class-embedding layer for digit label conditioning. To synthesize USPS digits $(\{0,\cdots,9\})$, we fine-tune this MNIST-pre-trained model on \(n_t = 5{,}103\) USPS training images (approximately 70\% of the dataset), while keeping the class-embedding layer frozen. Generation quality is evaluated on a held-out test set of 2,188 USPS images (about 30\% of the total) using the 1-Wasserstein distance between real and generated distributions.

\begin{figure}[H]
  \centering
  \begin{subfigure}[b]{0.7\textwidth}
    \centering
    \includegraphics[width=\textwidth]{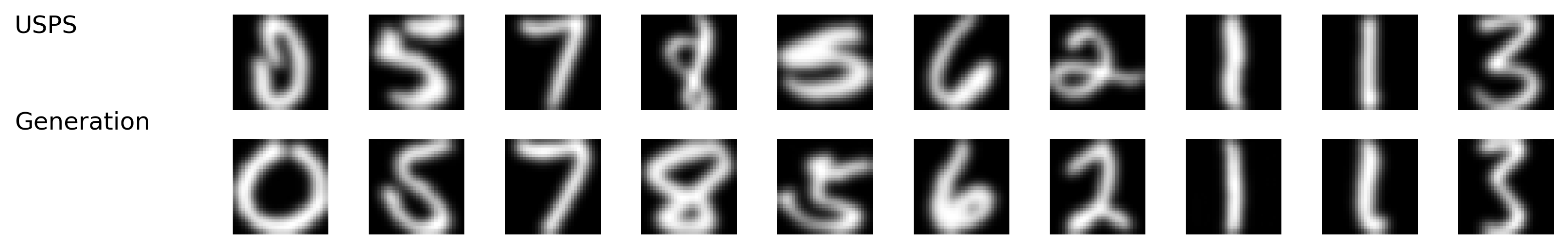}
    \caption{Conditional generation example. Real USPS examples (top) and generated samples (bottom).}
    \label{fig-con-ex}
  \end{subfigure}\\[1em]
  \begin{subfigure}[b]{0.7\textwidth}
    \centering
    \includegraphics[width=\textwidth]{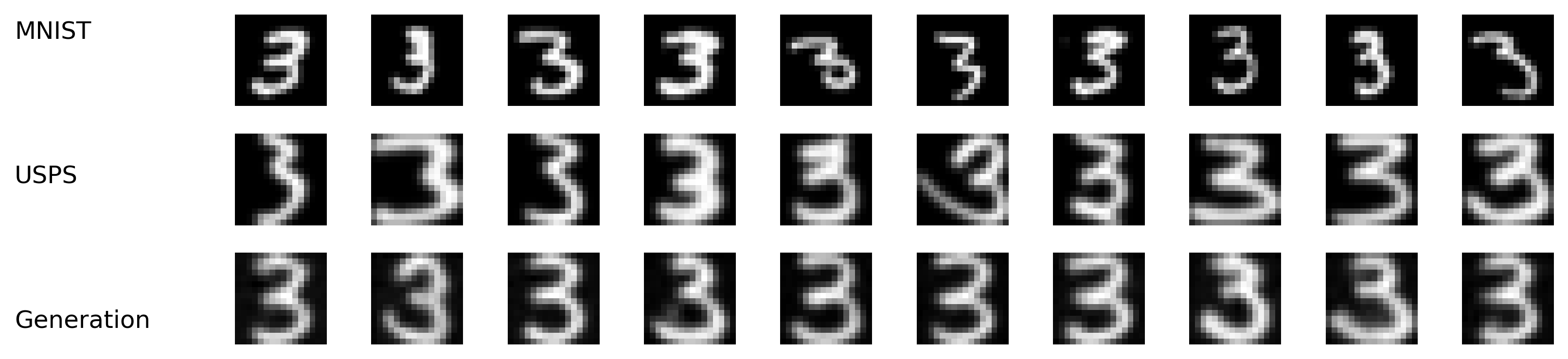}
    \caption{Unconditional generation example: Digit “3” examples—MNIST (top), USPS (middle), and USPS-style image generated by a diffusion model pre-trained on MNIST and adapted to USPS (bottom).}
    \label{fig-un-ex}
  \end{subfigure}
  \caption{Examples of image generation: conditional and unconditional generation.}
  \label{fig:cond-uncond-examples}
\end{figure}

\textbf{Unconditional generation.} 
We restrict the task to digit ``3'' images—an appropriate subset given that the MNIST–USPS benchmark is intended for conditional generation. We start from a diffusion model pre-trained on MNIST and fine-tune it on $n_t = 460$ USPS digit-3 samples, capitalizing on the larger MNIST corpus despite its stylistic gap (see Figure~\ref{fig-un-ex}). During pre-training we vary the MNIST source size, $n_s \in \{100, 500, 1{,}000, 2{,}000, 3{,}500, 6{,}000\}$. The UNet denoiser and auto-encoder backbone are implemented with the \texttt{Diffusers} library. Generation quality is measured by the 1-Wasserstein distance on an independent test split of 198 USPS digit-3 images (approximately 30\% of the dataset).

\begin{figure}[H]
\centering
\includegraphics[width=0.7\textwidth]{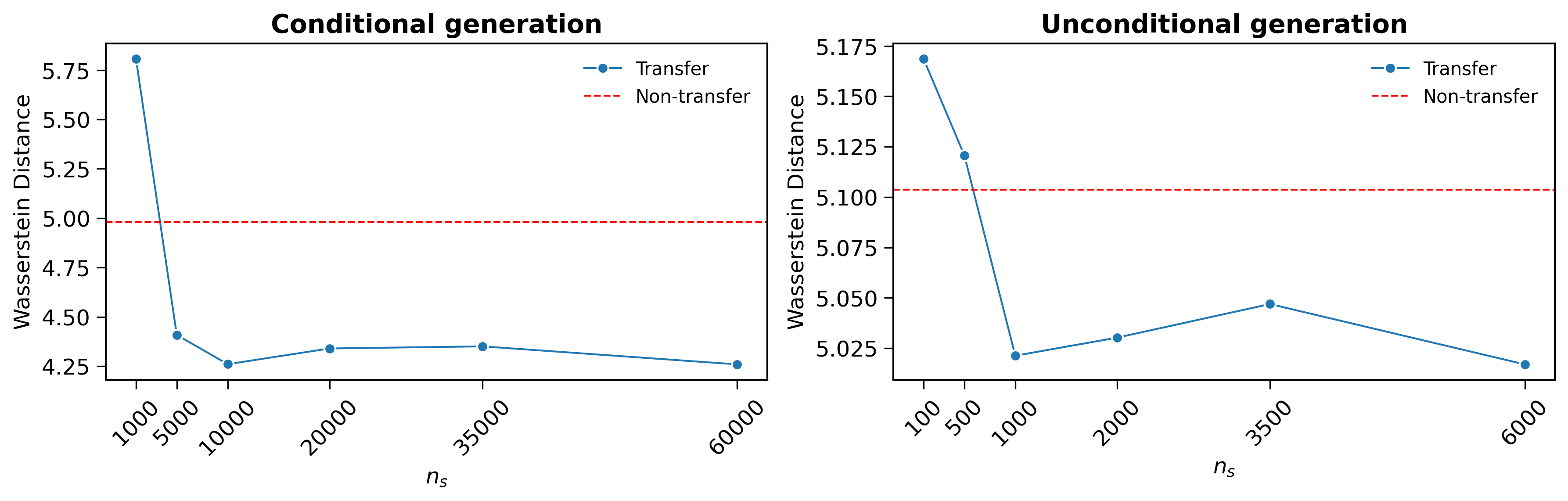}
\caption{Conditional (all digits, $n_t = 5103$)  and unconditional (digit “3”, $n_t = 100$)  generation accuracy on USPS, measured by the Wasserstein distance and plotted against the MNIST pre-training size $n_s$.The red dashed line indicates the USPS-only baseline without transfer.}
\label{fig-image-un}
\end{figure}

Figure \ref{fig-image-un} shows that, in both conditional and unconditional settings, the Wasserstein error of the transfer-diffusion model decreases monotonically as the MNIST pre-training set grows. Holding the USPS fine-tuning size fixed at $n_t$, the transfer model outperforms the USPS-only baseline once the source sample size $n_s$ surpasses a critical threshold. This phenomenon mirrors the trends reported in Section \ref{sec: numerical} and supports our theoretical claim that richer source data reduce target-domain error. The findings suggest that leveraging shared latent structure boosts generation fidelity, whereas an insufficient source corpus risks negative transfer. Latent-space interpolations (Figure \ref{fig:cond-uncond-examples}) further reveal a smooth stylistic transition from MNIST to USPS.

\vspace{-1cm}

\section{Conclusion}
\label{sec: conlusion}
This paper presents a comprehensive theoretical analysis of transfer learning for generative models, emphasizing the transformative potential of these models from a transfer learning perspective. We introduce a shared embedding framework to illustrate how the knowledge transfer between source and target domains facilitates synthetic data generation through shared embeddings. This research provides novel insights into the conditions under which transferred models can achieve enhanced generative performance by systematically quantifying generation errors. We apply our theory to two leading-edge generative models, diffusion models and coupling flows,
yielding new results that address unresolved challenges in the field. Additionally, we develop a theory on non-transfer generation accuracy for these models, establishing a standalone benchmark with independent significance.

% Acknowledgements and Disclosure of Funding should go at the end, before appendices and references

\acks{This work was supported in part by NSF Grant DMS-1952539 and NIH Grants R01AG069895, R01AG065636, R01AG074858, and U01AG073079. It was also partially supported by the Minnesota Supercomputing Institute (MSI) at the University of Minnesota.}
%All acknowledgements go at the end of the paper before appendices and references.
%Moreover, you are required to declare funding (financial activities supporting the
%submitted work) and competing interests (related financial activities outside the submitted work).
%More information about this disclosure can be found on the JMLR website.}


\appendix

\section{Proofs}
\label{appendix}
\subsection{Proofs for Section \ref{sec-general}}

\begin{proof}[Theorem \ref{theorem1}]
To bound $P(\rho_t(\gamma_t^0,\hat\gamma_t) \geq \varepsilon)$, note that 
\begin{align}
\label{eq-dec-p}
    P(\rho_t(\gamma_t^0,\hat\gamma_t) \leq  \varepsilon)\geq P(\rho_t(\gamma_t^0,\hat\gamma_t) \leq  \varepsilon|(2 \delta_t(\hat{h}))^{1/2}\leq \varepsilon)P((2 \delta_t(\hat{h}))^{1/2}\leq \varepsilon),
\end{align}
where $P(\cdot|2 \delta_t(\hat{h}))^{1/2}\leq \varepsilon)$ denotes the conditional probability given event $(2 \delta_t(\hat{h}))^{1/2}\leq \varepsilon)$. Here
$\varepsilon$, as defined in Theorem \ref{theorem1}, is $\varepsilon_t+\sqrt{3c_1}\varepsilon_s$, with $\varepsilon_t\geq\sqrt{2\delta_t(h^0)}$, $\varepsilon_s\geq\sqrt{2\delta_s(h^0)}$, 
\begin{eqnarray}
\label{entropy-Tz}
 \int_{k_t\varepsilon^2/16}^{4 c^{1/2}_{vt} \varepsilon_t}
\sup_{h\in \Theta_h}H_B^{1/2}(u,\smc{F}_t(h)) du \leq c_{h_t} n_t^{1/2} \varepsilon_t^{2},
\end{eqnarray}
and 
\begin{eqnarray}
\label{eq-entropy-source}
 \int_{k_s\varepsilon_s^2/16}^{4 c^{1/2}_{vs} \varepsilon_s}
H_B^{1/2}(u,\smc{F}_s) du \leq c_{h_s} n_s^{1/2} \varepsilon_s^{2}.
\end{eqnarray}

The first conditional probability is bounded by Proposition \ref{prop1} (cf. Remark 1). 
Note that 
$\varepsilon\geq \varepsilon_t $ satisfies when $\varepsilon_t$ satisfies \eqref{entropy-Tz}:
\begin{eqnarray}
\label{entropy-Tz2}
 \int_{k_t\varepsilon^2/16}^{4 c^{1/2}_{vt} \varepsilon}
\sup_{h\in \Theta_h}H_B^{1/2}(u,\smc{F}_t(h)) du \leq c_{h_t} n_t^{1/2} \varepsilon^{2}.
\end{eqnarray}
Hence, by Proposition \ref{prop1}, if Assumptions \ref{A-independent}, \ref{Variance} and \ref{sub-Gaussian} hold, the conditional probability given the condition $\sqrt{2\delta_t(\hat{h})}\leq \varepsilon$ is bounded:
\begin{equation}
\label{bound-t1z}
P(\rho_t(\gamma_t^0,\hat\gamma_t) \leq  \varepsilon|\sqrt{2\delta_t(\hat{h})}\leq \varepsilon) \geq 
1-\exp(-c_{e_t} n_t \varepsilon^2).
\end{equation}

For the second probability, we will use Assumption \ref{transferability} to show $\{\rho_s(\gamma_s^0,\hat\gamma_s) \leq \varepsilon_s\}\subset \{\sqrt{2\delta_t(\hat{h})}\leq \varepsilon\}$, where $\gamma_s=(\hat{\theta},\hat{h})$.
By the definition of $\delta_s(h)=\inf_{\{\gamma_s=(\theta_s,h): \phi_s\in \Theta_s,h\in \Theta_h\}}\rho^2_s(\gamma_s^0,\gamma_s)$, $\delta_s(\hat{h}) \leq
\rho_s^2(\gamma_s^0,\hat\gamma_s)$.
Under the event $\left\{\rho_s(\gamma_s^0,\hat\gamma_s) \leq \varepsilon_s\right\}$,
\begin{equation*}
|\delta_s(h)-\delta_s(h^0)|\leq \delta_s(h)+\delta_s(h^0)\leq 
\rho^2_s(\gamma_s^0,\hat\gamma_s)+\delta_s(h^0)\leq \frac{3}{2}\varepsilon_s^2.
\end{equation*}
The last inequality uses the assumption that
$\sqrt{2\delta_s(h^0)}\leq\varepsilon_s$. %This implies $\hat{h}\in \{h \in \Theta_h: {|\delta_s(h)-\delta_s(h^0)|}\leq \varepsilon_0\}$.  By Assumption \ref{transferability}, we have
    \begin{align*}
    \delta_t(\hat{h})\leq \delta_t(h^0)+c_1|\delta_s(\hat{h})-\delta_s(h^0)|\leq \delta_t(h^0)+c_1\rho_s^2(\gamma_s^0,\hat\gamma_s)+c_1\delta_s(h^0).
    \end{align*}
On the event that $\rho_s(\gamma_s^0,\hat\gamma_s) \leq \varepsilon_s$ with $\varepsilon_j\geq\sqrt{2\delta_j(h^0)}$, $j \in \{s,t\}$,
\begin{equation*}
\delta_t(\hat{h})\leq \delta_t(h^0)+c_1\varepsilon^2_s+c_1\delta_s(h^0)
\leq \frac{1}{2}\varepsilon_t^2+c_1\frac{3}{2}\varepsilon^2_s\leq \frac{1}{2}(\varepsilon_t+\sqrt{3c_1}\varepsilon_s)^2\leq \frac{1}{2}\varepsilon^2.   
\end{equation*}
This shows  $\{\rho_s(\gamma_s^0,\hat\gamma_s) \leq \varepsilon_s\}\subset \{\sqrt{2\delta_t(\hat{h})}\leq \varepsilon\}$.

Let $\delta_s=\inf_{\gamma_s\in \Theta_s\times \Theta_h}\rho_s^2(\gamma_s^0,\gamma_s)$.
Then, $\delta_s \leq 
\delta_s(h^0)$ when $h^0\in \Theta_h$. Thus, $\varepsilon_s>\sqrt{2\delta_s(h^0)}\geq\sqrt{2\delta_s}$. Together with \eqref{eq-entropy-source}, we derive through Proposition \ref{prop1} that there exists a constant $c_{e_s}>0$ such that,
\begin{equation}
\label{bound-t2z}
P((2 \delta_t(\hat{h}))^{1/2}\leq \varepsilon)\geq P(\rho_s(\gamma_s^0,\hat\gamma_s) \leq \varepsilon_s)\geq 1- 
\exp(-c_{e_s} n_s (\varepsilon_s)^2).
\end{equation}
Plugging \eqref{bound-t1z} and \eqref{bound-t2z} into \eqref{eq-dec-p} leads to the final result. This completes the proof. 
\end{proof}

\begin{lemma}
\label{large-d}
Assume that $f(\Y) \in \cal F$ satisfies the Bernstein condition with some constant $c_b$ for an i.i.d. sample $\Y^1,\cdots,\Y^n$.  Let $\varphi (M,v^2,{\cal F})=\frac{M^{2}}{2[4v^2+ M c_b /3 n^{1/2}]}$, 
where ${\rm Var}(f(\Y)) \leq v^2$. Assume that 
\begin{eqnarray}
\label{mean-var}
M \leq k n^{1/2} v^2/4c_b, 
\end{eqnarray}
with $0<k<1$ and
\begin{eqnarray}
\label{entropy0}
\int_{k M/(8 n^{1/2})}^{v} H_B^{1/2}(u,{\cal F}) du \leq M k^{3/2}/2^{10},
\end{eqnarray}
then 
\[  P^{*}(\sup_{\{f \in {\cal F}\}} n^{-1/2} \sum_{i=1}^n (f(\Y^i) -\E f(\Y^i)) \geq M) \leq  3
\exp(-(1-k) \varphi(M,v^2,n)),    \]
where $P^{*}$ denotes the outer probability measure for $\Y^1,\cdots,\Y^n$.
\end{lemma}

\begin{proof}[Lemma \ref{large-d}]
The result follows from the same arguments as in the proof of Theorem 3 in \citep{shen1994convergence} with $\Var(f(X)) \leq v^2$. 
Note that Bernstein's condition replaces the upper boundedness condition
there, and (4.5) there is not needed here. 
\end{proof}

Suppose that the class of loss functions is indexed by $\gamma\in\Gamma$ and $\smc{F}=\{l(\cdot,\pi\gamma^0)-l(\cdot,\gamma)\}$, where $\pi \gamma^0 \in \Gamma$ is an approximate point of $\gamma^0_t$ within $\Gamma$. Suppose $\hat{\gamma}=\argmin_{\gamma\in \Gamma} \sum_{i=1}^{n} l(\Y^i,\gamma)$. Then the following general results are established for $\rho^2(\gamma^0,\hat{\gamma})=\E (l(\Y,\gamma)-l(\Y,\gamma^0))$.

\begin{prop}
\label{prop1}
If $l(\Y,\gamma)-l(\Y,\gamma^0)$ satisfies the variance condition (Assumption \ref{Variance}) with a constant $c_v$ and the Bernstein condition (Assumption \ref{sub-Gaussian}) with a constant $c_v$, there exists a
small constant $k$ ($0<\frac{c_{b}}{4c_{v}} \leq k < 1$) 
such that for $\varepsilon>0$ satisfying 
\begin{equation}
\label{entropy}
\int_{k\varepsilon^2/16}^{4 c^{1/2}_{v} \varepsilon}
H_B^{1/2}(u,{\cal F}) du \leq c_{h} n^{1/2} \varepsilon^2, \quad 
\varepsilon^2 \geq 2 \inf_{\gamma\in\Gamma}\rho^2(\gamma^0,\gamma), 
\end{equation}
for $c_{h}=\frac{k^{3/2}}{2^{11}}$ and $c_e =\frac{(1-k)}{(8c^2_{v}+\frac{1}{24})}$, we have 
\begin{eqnarray*}
P(\rho(\gamma^0,\hat{\gamma}) \geq \varepsilon) 
\leq 4 \exp(- c_e n \varepsilon^2).
\end{eqnarray*}
\end{prop}

\begin{remark}
   This proposition holds for the source estimation. In the transfer case, the entropy and the approximation error for target learning may depend on the $\hat{h}$ derived from the source. The probability bound here can be extended to the condition probability given the condition of $\hat{h}$ by the independence between source and target training samples assumed in Assumption \ref{A-independent}.
\end{remark}
Theorem \ref{theorem2} is a direct result obtained from Proposition \ref{prop1}.
\begin{proof}[Proposition\ref{prop1}]
     Let $\nu_{n}(l(\gamma)- l(\pi \gamma^0))=n^{-1/2} \sum_{i=1}^{n} (l(\Y^i,\gamma)-l(\Y^i,\pi \gamma^0)-\E (l(\Y^i,\gamma)-l(\Y^i,\pi \gamma^0)))$ be an empirical process indexed by $\gamma \in \Gamma$, and $L(\gamma)=\sum_{i=1}^{n} (l(\Y^i,\gamma)-\E l(\Y^i,\gamma))$. 

 For $l=0,\cdots,$ let $A_l=\{\gamma \in \Gamma: 2^l \varepsilon^2 \leq \rho^2(\gamma^0, \gamma) < 2^{l+1} \varepsilon^2\}$. Note that $\sup_{A_l} \Var(l(\Y^i,\gamma)- l(\Y^i,\gamma^0)) \leq c_{v} 2^{l+1} \varepsilon^2$
(Assumption \ref{Variance}) and $\inf_{A_l} E(l(\Y^i,\gamma)- l(\Y^i,\pi \gamma^0)) \geq (2^l-\frac{1}{2}) \varepsilon^2$ when $\inf_{\gamma\in\Gamma}\rho^2(\gamma^0,\gamma) \leq \rho^2(\gamma^0,\pi\gamma^0) \leq \frac{1}{2} \varepsilon^2$; for $i=1,\cdots,n$. By Assumption \ref{Variance}, $\sup_{A_l} Var(l(\Y^i,\gamma)- l(\Y^i,\pi \gamma^0)) 
\leq 4 \sup_{A_l} Var(l(\Y^i,\gamma)- l(\Y^i, \gamma^0))+ 4 \sup_{A_l} Var(l(\Y^i,\gamma)- l(\Y^i,\pi\gamma^0)) \leq 8 c_{v} 2^{l+1} \varepsilon^2$.  Then
$P(\rho(\gamma^0,\hat{\gamma}) \geq \varepsilon)$ is bounded by 
\begin{eqnarray*}                          
& & P^{*} (\sup_{\{\rho(\gamma^0,\gamma) \geq \varepsilon, \gamma \in
\Gamma\}} (L(\gamma)-L(\pi\gamma^0))) \geq 0)
\leq \sum_{l=0}^{\infty} P^{*} (\sup_{A_l} (L(\gamma)-L(\pi\gamma^0))) \geq 0)\\
& = &  \sum_{l=0}^{\infty} P^{*}(\sup_{A_l} 
\nu_{n}((l(\gamma)-l(\pi\gamma^0)))  \geq n^{1/2} (2^l-1/2) \varepsilon^{2})).
\end{eqnarray*}

To apply Lemma \ref{large-d} to the empirical process over each $A_l$ ($l=0,\cdot$), we set $M=M_l=n^{1/2}(2^l-1/2) \varepsilon^2$ and $v^2=v_l^2=8 c_{v} 2^{l+1} \varepsilon^{2}$ there. Then, $\frac{M}{n^{1/2}v^2} \leq \frac{1}{16 c_{v}} \leq \frac{k}{4c_{bj}}$, given that $k\geq \frac{c_{bj}}{4c_{v}}$ 
according to Assumption \ref{Variance}, leading to \eqref{mean-var}. Consequently, $\varphi(M_l,v_l^2,{\cal F}) = \frac{M_l^{2}}{8c_{v} v_l^2+ 2 M_l c_{b}/3n^{1/2}} \geq \frac{M_l^2}{(8c_{v}+ 1/24c_{v}) v_l^2}$. Furthermore, for any $\varepsilon$ meeting \eqref{entropy}, it also fulfills \eqref{entropy0} with 
$(M_l,v^2_l)$ for $l \geq 1$ by examining the least favorable scenario of $l=0$. By Assumption \ref{sub-Gaussian}, 
\begin{eqnarray*}
& & \sum_{l=0}^{\infty} P^{*}(\sup_{A_l}
\nu_{n}(l(\gamma)-l(\pi \gamma^0))  \geq n^{1/2} (2^l-1/2) \varepsilon^{2})) \\
 & \leq & 3 \sum_{l=0}^{\infty} \exp(-(1-k)\frac{(2^l -1/2)^2 n \varepsilon_j^2}{c_{v} 2^{l+1}}
\frac{1}{(8c_{v}+1/24c_{v})}) 
  \leq  4 \exp(-c_e n_j \varepsilon^{2}),
\end{eqnarray*}
where $c_e =\frac{(1-k)}{(8c^2_{v}+\frac{1}{24})}$.
This completes the proof. 
\end{proof}

\subsection{Proofs for Section \ref{sec_diffusion}}

\subsubsection{Error for diffusion generation}

This subsection presents a general theory for the generation accuracy of conditional and unconditional diffusion models in a generic situation without non-transfer and dimension reduction. Then, we will modify the general results tailored to situations of transfer learning and dimension reduction subsequently.

Consider the conditional generation task for a $d_x$-dimensional vector $\X$ given a $d_z$-dimensional vector $\Z$.  Following the generation
process described in Section \ref{sec_diffusion}, we use \eqref{forward}-\eqref{reverse} to construct an empirical score matching loss based on a training
set $(\x^i,\z^i)_{i=1}^{n}$ of size $n$, $L(\theta)=\sum_{i=1}^{n} 
l(\x^i,\z^i;\theta)$ in \eqref{loss_2} with 
\begin{align}
\label{loss-diffusion-g}
l(\x,\z;\theta)=
\int_{\underline{\tau}}^{\overline{\tau}} \mathrm{E}_{\x(\tau)|\x,\z}\|\nabla \log p_{\x(\tau)|\x,\z}(\X(\tau)|\x,\z)-\theta(\X(\tau),\z,\tau)\|^2 \mathrm{d}\tau,
\end{align}
where the estimated score $\hat \theta(\x(\tau), \z,\tau)=\argmin_{\theta\in\Theta}L(\theta)$. Here, the parameter space $\Theta$ is defined as:
$\Theta=\{\theta\in\mathrm{NN}(\mathbb{L},\mathbb{W},\mathbb{S},\mathbb{B},\mathbb{E}):\mathbb{R}^{d_{x}+d_z+1}\rightarrow \mathbb{R}^{d_{x}}\}$,
representing a ReLU network with $\mathbb{L}$ layers, a maximum width of $\mathbb{W}$, the number of effective parameters $\mathbb{S}$, the output sup-norm $\mathbb{B}$ and the parameter bound $\mathbb{E}$.

Following the sampling scheme of the reverse process \eqref{s-reverse} and its alternative described in Section \ref{sec_3-2}, we derive the density of the generation sample $\hat{p}_{\x|\z}$. 

Next, we make some assumptions about the true conditional density $p^0_{\x|\z}$.
\begin{assumption}
\label{A_d_c}
Assume that $p^0_{\x|\z}(\x|\z) = \exp(-c_f\|\x\|^2 / 2) \cdot f(\x,\z)$, where $f$ belongs to $\smc{C}^{r}(\mathbb{R}^{d_x} \times [0,1]^{d_z},\R,B)$ for a constant radius $B>0$ and $c_f>0$ is a constant. Assume that $f$ is lower bounded away from zero with
$f \geq \underline{c}$.
\end{assumption}

Next, we give the results of the generation accuracy for conditional diffusion models.

\begin{theorem}[Generation error of diffusion models]
\label{thm_diff_general}
Under Assumption \ref{A_d_c}, setting the neural network's structural hyperparameters of $\Theta=\mathrm{NN}(\mathbb{L},\mathbb{W},\mathbb{S},\mathbb{B},\mathbb{E})$ as follows: $\mathbb{L}=c_L \log^4K$, $\mathbb{W}= c_W K\log^7K$, $\mathbb{S}= c_S K\log^9K$, 
$\log\mathbb{B}= c_B\log K$, $\log\mathbb{E}= c_E\log^4K$,  with stopping criteria from \eqref{forward}-\eqref{reverse} 
as $\log \underline{\tau}= -c_{\underline{\tau}}\log K$, $\overline{\tau}=c_{\overline{\tau}}\log K$, and $\underline{\tau}^*=\mathbb{I}_{\{r\le 1\}} \underline{\tau}$,  where $\{c_L,c_W, c_S, c_B,c_E,c_{\underline{\tau}},c_{\overline{\tau}}\}$ are sufficiently large constants, yields the error in diffusion generation via transfer learning: 
\begin{eqnarray}
\label{c-rate-general}
& P(\mathrm{E}_{\z}[\mathrm{TV}(p^0_{\x|\z},\hat{p}_{\x|\z})]\geq x(\beta_n + \delta_n))\leq \exp(-c_e n^{1-\xi} (x(\beta_n + \delta_n))^2), \text{ if $r>0$;} \nonumber \\
& P(\mathrm{E}_{\z}[\smc{K}^{1/2}(p^0_{\x|\z},\hat{p}_{\x|\z})]\geq x(\beta_n + \delta_n))\leq \exp(-c_e n^{1-\xi} (x(\beta_n + \delta_n))^2), \text{ if $r>1$,}\nonumber
\end{eqnarray}
for any $x\geq 1$, 
some constant $c_e>0$ and a small $\xi>0$. Here, $\beta_n$ and $\delta_n$ represent the estimation and approximation errors, given by:
$ \beta_n\asymp 
    \sqrt{\frac{K\log^{19}K}{n}},\quad \delta_n\asymp K^{-\frac{r}{d_x+d_z}}\log^{\frac{r}{2}+1}K$.
Setting $\beta_n = \delta_n$ to solve for $K$, and neglecting the logarithmic term, 
leads to 
$\beta_n = \delta_n \asymp n^{-\frac{r}{d_x+d_z+2r}}\log^m n$ with 
the optimal $K \asymp n^{\frac{d_x+d_z}{d_x+d_z+2r}}$, with $m = \max(\frac{19}{2}, \frac{r}{2}+1)$. Consequently, 
this provides the best bound $n^{-\frac{r}{d_x+d_z+2r}}\log^m n$.

Moreover, we extend this error bound to unconditional diffusion generation with $\Z= \emptyset$ and $d_z=0$,
$
\mathrm{TV}(p^0_{\x},\hat{p}_{\x})=O_p(n_{t}^{-\frac{r}{d_x+2r}}\log^m n_{t})
$
for $r>0$
and
$
\smc{K}^{1/2}(p^0_{\x},\hat{p}_{\x})=O_p(n_{t}^{-\frac{r}{d_x+2r}}\log^m n_{t})
$
for $r>1$.
\end{theorem}

To simplify the notation,
we write $p_{\x(\tau)|\z}(\x|\z)$ as $p_{\tau}(\x|\z)$ and
$p_{\x(\tau)|\x(0)}(\x|\x(0))$ as $p_{\tau}(\x|\x(0))$ in subsequent proofs.

\begin{lemma}[Approximation error of $\Theta$]
\label{thm_approx}
Under Assumption \ref{A_d_c}, there exists a ReLU network $\Theta=\mathrm{NN}(\mathbb{L},\mathbb{W},\mathbb{S},\mathbb{B},\mathbb{E})$ with depth $\mathbb{L}=c_L\log^4K$, width $\mathbb{W}=c_W K\log^{7} K$, the number of effective parameters $\mathbb{S}=c_S K\log^9K$, the parameter bound $\log \mathbb{E}=c_E\log^4K$, and $\mathbb{B}=\sup_{\tau}\mathbb{B}(\tau)= \sup_{\tau}c_{B}\sqrt{\log K}/\sigma_{\tau}$, such that for $\theta^0$ there exists $\pi \theta^0 \in \Theta$ with
\begin{eqnarray}
\label{app-error1}
\rho(\theta^0,\pi \theta^0)= O(K^{-\frac{r}{d_x+d_z}}
\log^{\frac{r}{2}+1}K),
\end{eqnarray}
provided that $\log \underline \tau=-c_{\underline\tau} \log K$ and $\overline \tau=c_{\overline\tau}\log K$ with sufficiently large constants $c_{\underline\tau}>0$ and $c_{\overline\tau}>0$.
Furthermore, there exists a subnetwork $\mathrm{NN}(\mathbb{L},\mathbb{W},\mathbb{S},\mathbb{B},\mathbb{E},\lambda)$, which has $\alpha$ H\"older continuity 
with $\lambda=c_{\lambda}$, $\alpha=\min(1,r-1)$ when $r>1$ and $\lambda=c_{\lambda}/\sigma_{\underline{\tau}}$, $\alpha=r$ when $r\leq 1$, where $c_\lambda$ is a sufficiently large positive constant. For $\theta^0$, there exists $\pi \theta^0 \in \tilde \Theta=\mathrm{NN}(\mathbb{L},\mathbb{W},\mathbb{S},\mathbb{B},\mathbb{E},\lambda)$ such that
\eqref{app-error1} holds. 
\end{lemma}

The approximation error for $\Theta=\mathrm{NN}(\mathbb{L},\mathbb{W},\mathbb{S},\mathbb{B},\mathbb{E})$ in Lemma \ref{thm_approx} directly follows from \cite{fu2024unveil}, which outlines this network's architecture and the input domain of the neural network can be limited in $[-R,R]^{d_x}\times[0,1]^{d_z}\times[\underline\tau,\overline{\tau}]$ where $R=c_x\sqrt{\log K}$ with $c_x$ is a positive constant dependent on the parameters in the true density. The approximation error for $\tilde \Theta=\mathrm{NN}(\mathbb{L},\mathbb{W},\mathbb{S},\mathbb{B},\mathbb{E},\lambda)$ is obtained by applying an $\alpha$ H\"older constraint to the network, where we choose $\lambda$ by the following lemma on the smooth property of the true score function.

The subsequent lemma elucidates the degree of smoothness of the gradient.

\begin{lemma}[Gradient H\"older continunity]
\label{l_lip}
Under Assumption \ref{A_d_c}, for any $\x \in \R^{d_x}$, $\tau>0$, and $\z,\z'\in [0,1]^{d_z}$,
          \begin{equation*}
          \begin{split}
   \sup_{\x}\frac{\|\nabla \log p_\tau(\x|\z)-\nabla \log p_\tau(\x|\z')\|_{\infty}}{\|\z-\z'\|^{\alpha}} \leq 
                \begin{cases}
    c^h_1,& \text{with } \alpha=\min(r-1,1) \text{ if } r>1, \\
    c_{2}^h/\sigma_\tau, & \text{with } \alpha=r \text{ if } r \leq 1, 
                \end{cases}
                \end{split}
                \end{equation*}
where $c^h_1=(B/\underline{c}+B^2\sqrt{d_z}^{1-\alpha}/\underline{c}^2)\max(1,1/c_f)$, $c_{2}^h=\sqrt{\frac{\pi}{2}}\left(B/\underline c+B^2/\underline c^2\right)\max(1,\sqrt{1/c_f})$, and $B,\underline c,c_f$ are specified in Assumption \ref{A_d_c}. 
\end{lemma}
\begin{proof} Consider any $\x \in \R^{d_x}$, $\tau>0$, and $\z,\z'\in [0,1]^{d_z}$ in what follows.

 By \eqref{forward}, the density of $\X(\tau)$ given $\Z$ can be expressed
through a mixture of the Gaussian and initial distribution: 
\begin{eqnarray}
\label{mixture}
p_\tau(\x|\z)=\int p_{N}(\x; \mu_\tau\y,\sigma_\tau) p_{\x(0)|\z}(\y|\z)\mathrm{d}\y,
\end{eqnarray}
where $p_{N}(\cdot;\mu_\tau\y,\sigma_\tau)$ is the Gaussian density of $N(\mu_\tau\y, \sigma^2_\tau\bm I)$ with $\mu_{\tau}=\exp(-\tau)$, $\sigma^2_{\tau}=1-\exp(-2 \tau)$, and $p_{\x(0)|\z}(\x|\z) = \exp(-c_f\|\x\|^2 / 2) \cdot f(\x,\z)$ is given in Assumption \ref{A_d_c}.

Direct calculations from \eqref{mixture} yield that
\begin{equation}
\label{eq-dec-score}
    \nabla \log p_{\tau}(\x|\z) = -\frac{c_f \x}{(\bar \mu_{\tau}^2 + c_f \bar \sigma_{\tau}^2)} + \frac{\nabla g(\x, \z, \tau)}{g(\x, \z, \tau)},
\end{equation}  
where $\nabla g(\x,\z,\tau)=\frac{\partial g(\x, \z, \tau)}{\partial\x}$ and
\begin{align}
\label{eq-bound-g}
    g(\x, \z, \tau)=\int f(\y, \z) p_{N}(\y;\bar{\mu}_\tau\x,\bar\sigma_\tau) \mathrm{d}\y\geq \underline c.
\end{align} 
The lower bound holds with $f \geq \underline c$. 
Here $p_{N}(\cdot;\bar \mu_\tau\x,\bar \sigma_\tau)$ is the Gaussian density of $N(\bar \mu_\tau\x,\bar \sigma^2_\tau\bm I)$ with $\bar \mu_\tau=\frac{\mu_{\tau}}{\mu_{\tau}^2 + c_f\sigma_{\tau}^2}$ and $\bar \sigma_\tau=\frac{\sigma_\tau}{\sqrt{\mu_{\tau}^2 + c_f \sigma_{\tau}^2}}$. Note
that $\bar \sigma_\tau\rightarrow 0$ and $\bar \mu_\tau \rightarrow 1$ as $\tau 
\rightarrow 0$. Furthermore, $\bar\mu_\tau\leq \max(1,1/c_f)$ and $\bar\sigma_\tau\geq \sigma_\tau\min(1,1/c_f^{1/2})$ since $\min(1,c_f)\leq\mu_{\tau}^2 + c_f \sigma_{\tau}^2\leq \max(1,c_f)$. Hence, 
\begin{align}
\label{eq-dec-lg}
&\|\nabla \log p_{\tau}(\x|\z)-\nabla \log p_{\tau}(\x|\z')\|_{\infty} 
\leq
\left\|\frac{\nabla g(\x, \z, \tau)}{g(\x, \z, \tau)}-\frac{\nabla g(\x, \z', \tau)}{g(\x, \z', \tau)}\right\|_{\infty}\nonumber
\\ 
&\leq \left\|\frac{\nabla g(\x, \z, \tau)-\nabla g(\x, \z', \tau)}{g(\x, \z', \tau)}\right\|_{\infty}+ \left\|\nabla g(\x, \z', \tau)\right\|_{\infty}\left|\frac{g(\x, \z, \tau)-g(\x, \z', \tau)}{g(\x, \z, \tau)g(\x, \z', \tau)}\right|.  
\end{align}

 To bound \eqref{eq-dec-lg}, we first consider the case of $r>1$ ($\alpha=\min(r-1,1)$). By integration by parts, 
\begin{align}
\label{eq-sg}
 \nabla g(\x,\z,\tau)&=\int_{R^{d_x}}  f(\y,\z) \nabla_{\x}p_{N}(\y;\bar{\mu}_\tau\x,\bar\sigma_\tau)\mathrm{d}\y
={\bar\mu_\tau}
 \int_{R^{d_x}} \nabla f(\y,\z) p_{N}(\y;\bar{\mu}_\tau\x,\bar\sigma_\tau)\mathrm{d}\y.
\end{align}
By Assumption \ref{A_d_c},  $\|\nabla f(\y,\z)\|_{\infty}\leq B$ since $f \in \smc{C}^{r}(\mathbb{R}^{d_x} \times [0,1]^{d_z},\R,B)$. Then 
\begin{align}
\label{eq-bound-sg}
    \|\nabla g(\x, \z, \tau)\|_{\infty} \leq {B}{\bar{\mu}_\tau}.
\end{align}
By Assumption \ref{A_d_c} with $r>1$, $\|\nabla f(\y, \z)- \nabla f(\y, \z')\|_{\infty} \leq B\|\z-\z'\|^{\alpha}$. By \eqref{eq-sg}, 
\begin{align}
\label{eq-bound-sgz}
\|\nabla  g(\x,\z,\tau)-\nabla g(\x,\z',\tau)\|_{\infty}
\leq \bar\mu_{\tau}\int B\|\z-\z'\|^{\alpha}p_{N}(\y;\bar{\mu}_\tau\x,\bar\sigma_\tau)\mathrm{d}\y  \leq B\bar\mu_{\tau}\|\z-\z'\|^{\alpha}. 
\end{align} 
Similarly as in \eqref{eq-bound-sgz}, with $\alpha=\min(r-1,1)$,
\begin{align}
\label{eq-bound-gz}
    \|g(\x, \z, \tau)-g(\x, \z', \tau)\|_{\infty}
   \leq B\|\z-\z'\| \leq \sqrt{d_z}^{1-\alpha} B \|\z-\z'\|^{\alpha}.
\end{align}

Plugging \eqref{eq-bound-g} and \eqref{eq-bound-sg}--\eqref{eq-bound-gz} into \eqref{eq-dec-lg} yields that
\begin{align*}
\|\nabla \log p_{\tau}(\x|\z)-\nabla \log p_{\tau}(\x|\z')\|_{\infty}\leq  \left(B\bar\mu_{\tau}/\underline{c}+B^2\bar\mu_{\tau}\sqrt{d_z}^{1-\alpha}/\underline{c}^2\right)\|\z-\z'\|^{\alpha}\leq  c^h_1 \|\z-\z'\|^{\alpha}
\end{align*}
for some constant $c^h_1=\left(B/\underline{c}+B^2\sqrt{d_z}^{1-\alpha}/\underline{c}^2\right)\max(1,1/c_f)$ since $\bar\mu_\tau\leq \max(1,1/c_f)$.

Next, we consider the case of $r\leq 1$, with $\alpha=r$. For any $j=1,2,\ldots,d_x$, the partial derivative of $g$ in the $j$-th element $\x_j$ of $\x$ $\nabla_{\x_j}  g$ 
satisfies: 
\begin{align}
\label{eq-bound-sgz2}
|\nabla_{\x_j}  g(\x,\z,\tau)-&\nabla_{\x_j} g(\x,\z',\tau)|\leq\int | f(\y, \z)- f(\y, \z')||\nabla_{\x_j}p_{N}(\y;\bar{\mu}_\tau\x,\bar\sigma_\tau)|\mathrm{d}\y  \nonumber \\
&\leq B\|\z-\z'\|^{\alpha} \int |\nabla_{\x_j}p_{N}(\y;\bar{\mu}_\tau\x,\bar\sigma_\tau)|\mathrm{d}\y  
= B\|\z-\z'\|^{\alpha}\sqrt{\frac{\pi}{2}}\frac{\bar{\mu}_\tau}{\bar\sigma_\tau},
\end{align} 
since
 $\int |\nabla_{\x_j}p_{N}(\y;\bar{\mu}_\tau\x,\bar\sigma_\tau)|\mathrm{d}\y=\bar{\mu}_\tau\int \frac{|\y_j-\bar{\mu}_\tau\x_j|}{\sigma_\tau}p_{N}(\y_j;\bar{\mu}_\tau\x_j,\bar\sigma_\tau)\mathrm{d}\y_j=\sqrt{\frac{\pi}{2}}\frac{\bar{\mu}_\tau}{\bar\sigma_\tau}$.
As in \eqref{eq-bound-gz},  
\begin{align}
\label{eq-bound-gz2}
    \|g(\x, \z, \tau)-g(\x, \z', \tau)\|_{\infty}
    \leq B\|\z-\z'\|^{\alpha} \int p_{N}(\y;\bar{\mu}_\tau\x,\bar\sigma_\tau)\mathrm{d}\y = B \|\z-\z'\|^{\alpha}.
\end{align}
By \eqref{eq-bound-sgz2} and the fact that $0<f(\y, \z) \leq B$, 
\begin{align}
\label{eq-bound-sg2}
    \|\nabla g(\x, \z, \tau)\|_{\infty}\leq \sup_{1 \leq j\leq d_x}\int f(\y, \z) |\nabla_{\x_j}p_{N}(\y;\bar{\mu}_\tau\x,\bar\sigma_\tau)|\mathrm{d}\y\leq B\frac{\bar{\mu}_\tau}{\bar\sigma_\tau}\sqrt{\frac{\pi}{2}}.
\end{align}

Plugging the bounds from \eqref{eq-bound-g} and \eqref{eq-bound-sgz2}--\eqref{eq-bound-sg2} into \eqref{eq-dec-lg} yields that
\begin{equation*}
\|\nabla \log p_{\tau}(\x|\z)-\nabla \log p_{\tau}(\x|\z')\|_{\infty}\leq \sqrt{\frac{\pi}{2}}\left(B/\underline c+B^2/\underline c^2\right)\frac{\bar \mu_\tau}{\bar\sigma_\tau} \leq \frac{c_{2}^h}{\sigma_\tau}\|\z-\z'\|^{\alpha}, 
\end{equation*}
for some constant 
$c_{2}^h=\sqrt{\frac{\pi}{2}}\left(B/\underline c+B^2/\underline c^2\right)\max(1,\sqrt{1/c_f})$ since $\bar\mu_\tau\leq \max(1,1/c_f)$. 
This completes the proof. 
\end{proof}

\begin{lemma}[Metric entropy]
\label{thm_entropy}
For the network $\Theta=\mathrm{NN}(\mathbb{L},\mathbb{W},\mathbb{S},\mathbb{B},\mathbb{E})$ defined in Lemma \ref{thm_approx}, 
the metric entropy of $\smc{F}=\{l(\cdot;\theta)-l(\cdot;\pi \theta^0): \theta \in\Theta\}$ 
is bounded, 
$
H_B(u,\smc{F})= O(K\log^{16}K\log(\frac{K}{u})).
$
\end{lemma}
\begin{proof}[Lemma \ref{thm_entropy}] 
For $\theta_j \in \Theta$; $j=1,2$, consider the case
$\sup_{\x,\z,\tau}\|\theta_1(\x,\z,\tau)-\theta_2(\x,\z,\tau)\|_{\infty} 
\leq u$. By \eqref{loss-diffusion-g}, for any $\x(0),\z$, 
\begin{align}
\label{eq3}
    &|l(\x(0),\z;\theta_1)-l(\x(0),\z;\theta_2)| \nonumber \\  \leq&\int_{{\underline{\tau}}}^{\overline{\tau}}\int_{\R^{d_x}} (\|\theta_1(\x,\z,\tau)-\nabla \log p_\tau(\x|\x(0),\z)\|+\|\theta_2(\x,\z,\tau)-\nabla \log p_\tau(\x|\x(0),\z)\|) \nonumber \nonumber \\
    &\|\theta_1(\x,\z,\tau)-\theta_2(\x,\z,\tau)\|p_{\tau}(\x|\x(0),\z)\mathrm{d}{\x}\mathrm{d}\tau \nonumber \\
   \leq& 2d^{1/2}_x u  \int_{{\underline{\tau}}}^{\overline{\tau}}\int_{\R^{d_x}}
( \sup_{\theta\in\Theta,\x,\z}\|\theta(\x,\z,\tau)\| + \|\nabla \log p_\tau(\x|\x(0),\z)\|) p_{\tau}(\x|\x(0),\z)\mathrm{d}{\x}\mathrm{d}\tau.
\end{align}
Moreover, note that $ p_{\tau}(\x|\x(0),\z)= p_{\tau}(\x|\x(0))=p_N(\x;\mu_{\tau}\x(0),\sigma^2_{\tau})$, and
\begin{align}
\label{ide}
\int_{\R^{d_x}}\|\nabla \log p_{\tau}(\x|\x(0))\|^2p_{\tau}(\x|\x(0))\mathrm{d}\x=\int_{\R^{d_x}} \frac{\|\x-\mu_\tau \x(0)\|^2}{\sigma^4_\tau} p_{\tau}(\x|\x(0))\mathrm{d}\x=\frac{d_x}{\sigma^2_\tau}.
\end{align}
By the assumption of Lemma 2, $\sup_{\theta\in\Theta,\x,\z}\|\theta(\x,\z,\tau)\| \leq \sqrt{d_x}c_{B}\frac{\sqrt{\log K}}{\sigma_\tau}$. By Cauchy-Schwartz inequality and \eqref{ide}, \eqref{eq3} is bounded by   
\begin{align*}
   \eqref{eq3}\leq & 2 d_x u\int_{{\underline{\tau}}}^{\overline{\tau}}(c_{B}\sqrt{\log K}
+1)/\sigma_\tau\mathrm{d}\tau \leq  2 d^{1/2}_x u \int_{\underline{\tau}}^{\overline{\tau}}
 \frac{c_{B}\sqrt{\log K}+1}{\sqrt{1-e^{-1}}\min(1,\sqrt{2\tau})}\mathrm{d}\tau\\
   \leq& \frac{2 d_x(c_{B}\sqrt{\log K}+1)c_{\overline{\tau}}\log K}{\sqrt{1-e^{-1}}}u \leq c^* 
(\log^{3/2} K) u,
\end{align*}
with $c^*=\frac{2 d_x(c_{B}+1)c_{\overline{\tau}}}{\sqrt{1-e^{-1}}}$ when $K >e$, where the forth inequality is because $\sigma_{\tau}=\sqrt{1-\exp(-2 \tau)} \geq \sqrt{1-e^{-1}}\sqrt{2\tau}$ when $\tau\leq \frac{1}{2}$ and $\sigma_{\tau} \geq \sqrt{1-e^{-1}}$ when $\tau> \frac{1}{2}$. 

Then, $(\mathbb{E}_{\x,\z}|l(\x(0),\z;\theta_1)-l(\x(0),\z;\theta_2)|^2)^{1/2}\leq c^* 
(\log^{3/2} K) u
$.
Consequently, by Lemma 2.1 in \cite{ossiander1987central}, we can bound the bracketing $L_2$ metric entropy by the $L_{\infty}$ metric entropy $H(u,\Theta)$, the logarithm of the number of $u$ balls in the sup norm needed to cover $\Theta$,
$H_B(u,\smc{F})\leq  H((2c^*\log^{3/2} K)^{-1} u,\Theta)$ for small $u>0$,
where $\Theta$ is defined in Lemma \ref{thm_approx} same as in \cite{fu2024unveil}. 

By Lemma C.2 \cite{oko2023diffusion} and Lemma D.8 \cite{fu2024unveil}, $H(\cdot,\Theta)$
is bounded by the hyperparameters of depth $\mathbb{L}$, width $\mathbb{W}$, number of parameters $\mathbb{S}$, parameter bound $\mathbb{E}$ and the diameter of the input domain, $H(u,\Theta) \leq O(\mathbb{S}\mathbb{L}\log(\mathbb{E}\mathbb{W}\mathbb{L}\max(R,\overline{\tau})/u))=O(K(\log^{13}K)(\log^4K-\log u))$ given the approximation error in Lemma \ref{thm_approx} with
$\mathbb{L}\asymp\log^4K$, $\mathbb{W}\asymp K\log^{7} K$, $\mathbb{S}\asymp K\log^9K$, $\log\mathbb{E}\asymp\log^4K$, $R\asymp \sqrt{\log K}$ and $\overline{\tau}\asymp\log K$. Thus, 
$H(\delta,\smc{F})\leq   H((c^*\log^{3/2} K)^{-1} u,\Theta)=O\left(K\log^{16}K\log\frac{K}{u}\right)$.
This completes the proof. 
\end{proof}

\begin{lemma}
\label{l-lowertau}
Let $\alpha=\min(1,r-1)$ when $r>1$. Under Assumption \ref{A_d_c}, for any 
$\z\in [0,1]^{d_z}$ and small $\underline \tau>0$, 
there exists a constant $c>0$, as given in \eqref{c-lemma6}, such that
\begin{align}
\label{gradient}
    \int_{0}^{\underline \tau}\mathrm{E}_{\x(\underline \tau),\x(\tau)|\z}\|\nabla\log p_{\underline \tau}(\X(\underline \tau)|\z)-\nabla\log p_{\tau}(\X(\tau)|\z)\|^2\mathrm{d}\tau\leq c\underline\tau^{1+\alpha}.
\end{align}
\end{lemma}
\begin{proof}
Consider any $0\leq \kappa< \underline \tau$. First,
\begin{align*}
  & \mathrm{E}_{\x(\underline\tau),\x(\kappa)|\z}\|\nabla\log p_{\underline\tau}(\X(\underline\tau)|\z)-\nabla\log p_{\kappa}(\X(\kappa)|\z)\|^2   \leq  I_3 + 
I_4, \nonumber  \\
 & I_3  = \mathrm{E}_{\x(\underline\tau),\x(\kappa)|\z}\|\nabla\log p_{\underline\tau}(\X(\underline\tau)|\z)-\nabla\log p_{\underline\tau}(\X(\kappa)|\z)\|^2,  \nonumber \\
 & I_4  =\mathrm{E}_{\x(\kappa)|\z}\|\nabla\log p_{\underline\tau}(\X(\kappa)|\z)-\nabla\log p_{\kappa}(\X(\kappa)|\z)\|^2. 
\end{align*}
By \eqref{gradient-lemma7} in Lemma \ref{l-lg-c}, for
some constants $c_1^g>0$ and $c_{2}^g>0$, 
\begin{align*}
    I_3 &\leq  2d_x (c_1^g)^2(\mathrm{E}_{\x(\underline\tau),\x(\kappa)|\z}\|\X(\underline\tau)-\X(\kappa)\|^2
+2 d_x(c_{2}^g)^2\mathrm{E}_{\x(\underline\tau),\x(\kappa)|\z}\|\X(\underline\tau)-\X(\kappa)\|^{2\alpha}).
\end{align*}

 For the first term of $I_3$, by \eqref{forward} with $b_\tau=1$,
\begin{align*}
& \mathrm{E}_{\x(\underline\tau),\x(\kappa)|\z}\|\X(\underline\tau)-\X(\kappa)\|^2 = \mathrm{E}_{\x(\tau)|\z,w(\tau)}  \left\|\int_{\kappa}^{\underline\tau} -b_\tau \X(\tau) \mathrm{d}\tau  
+ \sqrt{2b_\tau }  \int_{\kappa}^{\underline\tau} d W(\tau) \right\|^2 \\
&\leq 2\mathrm{E}_{\x(\tau)|\z,w(\tau)} \left(\left\|\int_{\kappa}^{\underline\tau}  \X(\tau) \mathrm{d}\tau\right\|^2  + 2\left\| \int_{\kappa}^{\underline\tau} d W(\tau) \right\|^2\right)  \\
& \leq 2(\underline\tau-\kappa) \left( \int_{\kappa}^{\underline\tau} \E_{\x(\tau)|\z}\left\| \X(\tau) \right\|^2 \mathrm{d}\tau  + 2d_x\right),
\end{align*}
where the last inequality uses the fact that $\|\int_{\kappa}^{\underline\tau}  \X(\tau) \mathrm{d}\tau\|^2\leq (\underline\tau-\kappa)\int_{\kappa}^{\underline\tau}\|\X(\tau)\|^2\mathrm{d}\tau$ by the Cauchy-Schwartz inequality, and $\int_{\kappa}^{\underline\tau} d W(\tau)\sim N(0,\underline\tau-\kappa)$.

Similarly, for the second term in $I_3$, by Jensen's inequality, $\mathrm{E}_{\x(\underline\tau),\x(\kappa)|\z}\|\X(\underline\tau)-\X(\kappa)\|^{2\alpha}\leq 
\left(\mathrm{E}_{\x(\underline\tau),\x(\kappa)|\z}\|\X(\underline\tau)-\X(\kappa)\|^{2}\right)^{\alpha}
\leq 2(\underline\tau-\kappa)^{\alpha}\left( \int_{\kappa}^{\underline\tau} \E_{\x(\tau)|\z}\left\| \X(\tau) \right\|^2 \mathrm{d}\tau  + 2d_x\right)^{\alpha}$ for $0<\alpha \leq 1$. Moreover, 
note that $\X(\tau) \sim N(\mu_\tau \X(0), \sigma_\tau^2\bm I)$ given $\X(0)$. Direct computation yields that 
\begin{eqnarray}
\label{eq-2moment}
& \E_{\x(\tau)|\z}\left\| \X(\tau) \right\|^2=\E_{\x(0)|z}\E_{\x(\tau)|\x(0)}\left\| \X(\tau) \right\|^2
\leq  \mathbb{E}_{\x(0)|z}(\mu_\tau^2 \|\X(0)\|^2+d_x) \nonumber\\
&\leq 
\E_{\x(0)|z}\left\| \X(0) \right\|^2+d_x\leq c_M, 
\end{eqnarray}
where $c_M=\sqrt{2\pi}Bd_x/c^{3/2}_f+d_x$ is derived from that
fact that $p_{\x(0)|z}(\x|\z)\leq B\exp(-\frac{c_f\|\x\|^2}{2})$ for any $\z$ in Assumption \ref{A_d_c}.
Hence, given sufficiently small $\underline\tau$,
combining these two bounds yields that 
$I_3 \leq c_{I_3} (\underline\tau-\kappa)^{\alpha}$ for $c_{I_3}=8d_x\max(c_1^g,c_{2}^g)^2(c_M+2d_x)$.
 
By \eqref{eq-dec-score}, $I_4\leq I_5+I_6$ with 
\begin{align*}
I_5= \mathrm{E}_{\x(\kappa)|\z}\left\|\frac{c_f \X(\kappa)}{(\mu_{\underline\tau}^2 + c_f \sigma_{\underline\tau}^2)}-\frac{c_f \X(\kappa)}{(\mu_{\kappa}^2 + c_f \sigma_{\kappa}^2)}\right\|^2,\\
I_6=\mathrm{E}_{\x(\kappa)|\z}\left\|\nabla \log g(\X(\kappa),\z,\tau)-\nabla \log g(\X(\kappa),\z,\kappa)\right\|^2.
\end{align*}
By \eqref{eq-2moment}, given $\mu_{\tau}=\exp(-\tau)$ and $\sigma^2_{\tau}=1-\exp(-2 \tau)$,
\begin{align*}
    I_5\leq \sup_{\tau \in [0,\underline\tau] } \frac{\partial}{\partial \tau} \left[\frac{c_f}{\mu^2_{\tau}+c_f\sigma^2_{\tau}}\right]^2(\underline\tau-\kappa)^2\mathrm{E}_{\x(\kappa)|\z}\|\X(\kappa)\|^2\leq c_{I_5}( \underline\tau-\kappa)^2,
\end{align*}
where $\sup_{\tau \in [0,\underline\tau] } \left[\frac{\partial}{\partial \tau} \left(\frac{c_f}{\mu^2_{\tau}+c_f\sigma^2_{\tau}}\right)\right]^2\leq 4\frac{c_f^2(1-c_f)^2}{\min(1,c_f^2)}$ and $c_{I_5}= 4\frac{c_f^2(1-c_f)^2}{\min(1,c_f^2)}c_M>0$. 

To bound $I_6$, we bound  $\|\nabla g(\X(\kappa),\z,\tau)-\nabla f(\X(\kappa),\z)\|$ and $\|g(\X(\kappa),\z,\tau)-f(\X(\kappa),\z)\|$ separately. 
By \eqref{eq-sg}, $\nabla g(\x,\z,\tau)={\bar\mu_\tau} \int_{R^{d_x}} \nabla f(\y,\z) p_{N}(\y;\bar{\mu}_\tau\x,\bar\sigma_\tau)\mathrm{d}\y$.
By the triangle inequality, $\|\nabla g(\x,\z,\tau)-\nabla f(\x,\z)\|$ is bounded by  
\begin{align}
\label{eq5}
   & \|\nabla g(\x,\z,\tau)-{\bar{\mu}_\tau}\nabla f({\bar\mu_\tau}\x,\z)\|
    +  {\bar\mu_\tau}\|\nabla f({\bar{\mu}_\tau}\x,\z)-\nabla f(\x,\z)\| 
  +\|{\bar\mu_\tau}\nabla f(\x,\z)-\nabla f(\x,\z)\| \nonumber \\
  &\leq  \bar{\mu}_\tau
    \left\|
    \int(\nabla f(\bar{\mu}_\tau\x)-\nabla f(\y))p_{N}(\y;\bar{\mu}_\tau\x,\bar\sigma_\tau) \mathrm{d}\y
    \right\|+{\bar{\mu}_\tau}B\left\|(1-\bar{\mu}_\tau)\x\right\|^{\alpha}+\left|1-{\bar{\mu}_\tau}\right|B.
\end{align}
Using the smooth property and the Cauchy-Schwartz inequality, 
the first term in \eqref{eq5} is bounded by 
$\bar{\mu}_\tau \int B\|\bar{\mu}_\tau\x-\y\|^{\alpha}p_{N}(\y;\bar{\mu}_\tau\x,\bar\sigma_\tau)\mathrm{d}\y\leq \bar{\mu}_\tau
B (d_x\bar\sigma^2_\tau)^{\alpha/2}$. 
Then, \eqref{eq5}, with $\bar \mu_\tau=\frac{\mu_{\tau}}{\mu_{\tau}^2 + c_f\sigma_{\tau}^2}$ and  $\bar \sigma_\tau=\frac{\sigma_\tau}{\sqrt{\mu_{\tau}^2 + c_f \sigma_{\tau}^2}}$,  is bounded by 
\begin{align*}
& (\bar{\mu}_\tau (d_x\bar\sigma^2_\tau)^{\alpha/2}+{\bar{\mu}_\tau}\left\|(1-\bar{\mu}_\tau)\x\right\|^{\alpha}+\left|1-{\bar{\mu}_\tau}\right|) B \\
%    & \frac{\mu_{\tau}}{(\mu_{\tau}^2 + c_f \sigma_{\tau}^2)}B\left(\frac{d_x\sigma^2_{\tau}}{\mu_{\tau}^2 + c_f \sigma_{\tau}^2}\right)^{\frac{\alpha}{2}}+\left|1-\frac{\mu_{\tau}}{(\mu_{\tau}^2 + c_f \sigma_{\tau}^2)}\right|B+
% \frac{\mu_{\tau}}{(\mu_{\tau}^2 + c_f \sigma_{\tau}^2)}\left|1-\frac{\mu_{\tau}}{(\mu_{\tau}^2 + c_f \sigma_{\tau}^2)}\right|^{\alpha} B\|\x\|^{\alpha} \\   
&\leq \frac{B(2d_x\tau)^{\alpha/2}}{\min(1,c_f^{1+\alpha/2})}
+\frac{(2c_f+1)B\tau^{\alpha}}{\min(1,c_f^{1+\alpha})}\|\x\|^{\alpha}+\frac{(2c_f+1)B\tau}{\min(1,c_f)}.
\end{align*}
The last inequality holds because $\sigma^2_\tau=1-\exp(-2\tau)\asymp 2\tau$ when $\tau \rightarrow 0$.
Combining these constants and using the bound in \eqref{eq-2moment} leads
to 
\begin{align}
\label{eq-b2-1}
\mathrm{E}_{\x(\kappa)|\z}\|\nabla g(\X(\kappa),\z,\tau)-\nabla f(\X(\kappa),\z)\|^2\leq 3\frac{\max((2c_f+1)B(c_M)^{\alpha},B(2d_x)^{\alpha/2})^2}{\min(1,c_f^{2+2\alpha})} \tau^{\alpha}.
\end{align}
Similarly, 
\begin{align*}
    \|g(\x,\z,\tau)-f(\x,\z)\|&\leq \|g(\x,\z,\tau)-f(\bar\mu_\tau\x,\z)\|+\|f(\bar\mu_\tau\x,\z)-f(\x,\z)\|\\
    &\leq \|\int(f(\bar\mu_\tau\x,\z)-f(\y,\z)))p_{N}(\y;\bar{\mu}_\tau\x,\bar\sigma_\tau) \mathrm{d}\y\|
    +B|1-\bar\mu_\tau|\|\x\|\\
    &\leq B(d_x\bar\sigma^2_\tau)^{1/2}+B|1-\bar\mu_\tau|\|\x\|
    \leq \frac{B(2d_x\tau)^{1/2}}{\min(1,c_f^{1/2})}+\frac{(2c_f+1)B}{\min(1,c_f)}\|\x\|.
\end{align*}
By \eqref{eq-2moment},  
\begin{align}
\label{eq-b2-2}
\mathrm{E}_{\x(\kappa)|\z}\|g(\X(\kappa),\z,\tau)- f(\X(\kappa),\z)\|^2\leq 2\frac{\max((2c_f+1)Bc_M,B(2d_x)^{1/2})^2}{\min(1,c_f^{2})} \tau^{\alpha}.
\end{align}
Plugging \eqref{eq-b2-1} and \eqref{eq-b2-2} into \eqref{eq-dec-lg}, we obtain
\begin{align*}
I_6&\leq \Big(\frac{2B}{\min(1,c^{\alpha+1}_f)\underline c}
+\frac{2B^2}{\min(1,c_f^2)\underline{c}^2} \Big)\mathrm{E}\|g(\X(\kappa),\z,\underline\tau)- f(\X(\kappa),\z)\|^2 \leq c_{I_6}\underline{\tau}^{\alpha},
\end{align*}
where $c_{I_6}=10\frac{\max((2c_f+1)Bc_M,B(2d_x)^{1/2})^2}{\min(\min(1,c_f^{3+3\alpha})\underline c/B,\min(1,c_f^{4})\underline{c}^2/B^2)}$. Combining the bounds for $I_3$ and $I_4$ through $I_5$ and $I_6$ yields the following result:
\begin{align}
\label{c-lemma6}
& \int_0^{\underline\tau} \mathrm{E}_{\x(\underline\tau),\x(\kappa)|\z}\|\nabla\log p_{\underline\tau}(\X(\underline\tau)|\z)-\nabla\log p_{\kappa}(\X(\kappa)|\z)\|^2 \mathrm{d}\kappa \nonumber\\
\leq& \int_0^{\underline\tau}(c_{I_3}(\underline{\tau}-\kappa)^{\alpha}+c_{I_5}(\underline{\tau}-\kappa)^2+c_{I_6}\underline{\tau}^{\alpha}\mathrm{d}\kappa 
\leq c\underline\tau^{1+\alpha},
\end{align}
with $c=\frac{1}{1+\alpha}(c_{I_3}+c_{I_5})+c_{I_6}$.
This completes the proof. 
\end{proof}

\begin{lemma}
\label{l-lg-c}
  Under Assumption \ref{A_d_c} with $r>1$, for any $\z\in [0,1]^{d_z}$, $\tau\geq 0$, and $\x,\x'\in \mathbb{R}^{d_x}$,
\begin{eqnarray}
\label{gradient-lemma7}
  \|\nabla \log p_\tau(\x|\z)-\nabla \log p_\tau(\x'|\z)\|_{\infty}
  \leq c_1^g\|\x-\x'\|^{\alpha}+c_{2}^g\|\x-\x'\|,
\end{eqnarray}
where $\alpha=\min(1,r-1)$, $c_1^g=\frac{B}{\min(1,C^{\alpha+1}_f)\underline c}$ and $c_{2}^g=\left(\frac{B^2}{\min(1,c_f^2)\underline{c}^2}+\max(1,c_f)\right)$, with $B$ and $c_f$ defined in Assumption \ref{A_d_c}.
\end{lemma}
\begin{proof}
Note that for any $\z\in [0,1]^{d_z}$, $\tau\geq 0$, and $\x,\x'\in \mathbb{R}^{d_x}$,
\begin{align}
\label{eq1}
\|\nabla \log p_{\tau}(\x|\z)&-\nabla \log p_{\tau}(\x'|\z)\|_{\infty}\leq   \left\|\frac{\nabla g(\x, \z, \tau)-\nabla g(\x', \z, \tau)}{g(\x, \z, \tau)}\right\|_{\infty} \nonumber \\
&+ \left\|\nabla g(\x, \z, \tau)\right\|_{\infty}\left|\frac{g(\x, \z, \tau)-g(\x', \z, \tau)}{g(\x, \z, \tau)g(\x', \z, \tau)}\right|+   \frac{c_f }{(\mu_{\tau}^2 + c_f \sigma_{\tau}^2)}\|\x-\x'\|.
\end{align}
By \eqref{eq-sg},
\begin{align}
\label{eq2}
&\|\nabla  g(\x,\z,\tau)-\nabla g(\x',\z,\tau)\|_{\infty} \nonumber \\
& \leq{\bar{\mu}_\tau}\int \|\nabla f(\y, \z)-\nabla f(\y-\bar\mu_{\tau} (\x-\x'), \z)\|_{\infty}p_{N}(\y;\bar{\mu}_\tau\x,\bar\sigma_\tau)\mathrm{d}\y \nonumber \\
&\leq{\bar{\mu}_\tau} B\|\bar{\mu}_\tau(\x-\x')\|^{\alpha} \leq \bar\mu_{\tau}^{\alpha+1} B\|\x-\x'\|^{\alpha}.
\end{align}
Similarly, $|g(\x, \z, \tau)-g(\x', \z, \tau)|\leq \bar\mu_{\tau} B\|\x-\x'\|$.

  Finally, \eqref{eq2} together with the fact that $\|\nabla g(\x, \z, \tau)\|_{\infty}
\leq B\frac{\bar{\mu}_\tau}{\bar\sigma_\tau}\sqrt{\frac{\pi}{2}}$ in \eqref{eq-bound-sg2}
and $g(\x, \z, \tau) \geq \underline{c}>0$
in \eqref{eq-bound-g}, leads to \eqref{gradient-lemma7}, as in the proof of Lemma \ref{l_lip}. 
This completes the proof.
\end{proof}

\begin{proof}[Theorem \ref{thm_diff_general}]
It suffices to apply Proposition \ref{prop1} 
with $l(\Y;\gamma)=l(\X,\Z;\theta)$ 
for the diffusion settings 
to the excess risk 
\begin{equation*}
\rho^2(\theta^0,\theta)=\mathrm{E}_{\x,\z}[l(\x,\z,\theta)-l(\x,\z,\theta^0)]=
\int_{\underline\tau}^{\overline{\tau}}\mathrm{E}_{\x(\tau),\z}\|\nabla \log p_\tau(\X(\tau)|\Z)-\theta(\X(\tau),\Z,\tau)\|^2\mathrm{d}\tau.
\end{equation*}
Here, $\theta^0(\x,\z,\tau)=\nabla \log p_\tau(\x|\z)$ and $\theta\in \Theta$ is used to approximate $\theta^0$.

We first show that $l(\cdot;\theta)$ is bounded, satisfying the Bernstein condition with $c_b=O(\log^2K)$. Note that 
$l(\x(0),\z;\theta)=\int_{{\underline{\tau}}}^{\overline{\tau}}\int_{\R^{d_x}}\|\nabla \log p_{\tau}(\x|\x(0),\z)-\theta(\x,\z,\tau)\|^2 p_{\tau}(\x|(\x(0),\z))\mathrm{d}\x\mathrm{d}\tau \geq 0$. Further,
$\|\nabla \log p_{\tau}(\x|\x(0),\z)-\theta(\x,\z,\tau)\|^2 \leq 2 (
\|\nabla \log p_{\tau}(\x|\x(0),\z)\|^2+\|\theta(\x,\z,\tau)\|^2)$.
By the assumption of Lemma 2 that $\sup_{\x,\z}\|\theta(\x,\z,\tau)\| 
\leq c_{B} \frac{\sqrt{\log K}}{\sigma_\tau}$ and \eqref{ide},
\begin{align} 
\label{eq-cb*}
 &l(\x(0),\z;\theta) 
    \leq \int_{{\underline{\tau}}}^{\overline{\tau}}\frac{2d_x}{\sigma_{\tau}^2}\mathrm{d}\tau+
\int_{{\underline{\tau}}}^{\overline{\tau}}\frac{2d_xc_{B}^2\log K}{\sigma_{\tau}^2}\mathrm{d}\tau \nonumber\\
& \leq 2 d_x \int_{{\underline{\tau}}}^{\overline{\tau}}\frac{(c_{B}^2\log K+1)}{\sigma_{\tau}^2}\mathrm{d}\tau
    \leq 2 d_x \int_{{\underline{\tau}}}^{\overline{\tau}}\frac{(c_{B}^2\log K+1)}{(1-e^{-1})\min(1,2\tau)}\mathrm{d}\tau \leq c^*_b\log^2 K,
\end{align}
where $c^*_b=2\frac{d_x(c_{B}^2\log K+1)(c_{\overline{\tau}}+c_{\underline\tau})}{1-e^{-1}}$. 
 
To show that $l(\x(0),\z;\theta^0)$ is bounded, where $\theta^0=\nabla \log p_{\tau}(\x|\z)$ may not necessarily belong to $\Theta$, we consider $\x(0)$ within the range $[-c_R\sqrt{\log K}, c_R\sqrt{\log K}]^{d_x}$. This is achieved by employing the truncation at $c_R\log K$ for the unbounded density $p_{\tau}(\x|\z)$. By
\eqref{eq-dec-score}, \eqref{eq-bound-g}, and \eqref{eq-bound-sg2}, 
$\|\nabla \log p_{\tau}(\x|\z)\| \leq \frac{c_f}{(\bar \mu_{\tau}^2 + c_f \bar \sigma_{\tau}^2)}\|\x\|+\sqrt{\frac{d_x \pi}{2}} \frac{B\bar{\mu}_\tau}{\bar c \bar\sigma_\tau}$. By the Cauchy–Schwarz inequality,
\begin{align}  &l(\x(0),\z;\theta^0)=\int_{{\underline{\tau}}}^{\overline{\tau}}\int_{\R^{d_x}}\|\nabla \log p_{\tau}(\x|\x(0),\z)-\theta^0(\x,\z,\tau)\|^2 p_{\tau}(\x|(\x(0),\z))\mathrm{d}\x\mathrm{d}\tau \nonumber \\
    &\leq \int_{{\underline{\tau}}}^{\overline{\tau}}\frac{2d_x}{\sigma_{\tau}^2}\mathrm{d}\tau+ 2\int_{{\underline{\tau}}}^{\overline{\tau}}
    \int_{\R^{d_x}}2\left[\left(\sqrt{d_x\frac{\pi}{2}}\frac{B\bar\mu_\tau}{\underline{c}\bar\sigma_\tau}\right)^2+\left(\frac{c_f\|\x\|}{\mu_\tau^2+c_f\sigma_\tau^2}\right)^2\right]p_{\tau}(\x|(\x(0),\z))\mathrm{d}\x\mathrm{d}\tau
     \nonumber \\
    &\leq 2\int_{{\underline{\tau}}}^{\overline{\tau}}\frac{d_x(\frac{\pi B^2}{\underline{c}^2}\max(1,1/c_f)+1)}{\sigma_{\tau}^2}\mathrm{d}\tau +4\int_{{\underline{\tau}}}^{\overline{\tau}}\max(1,c_f^2)(\mu_\tau\|\x(0)\|^2+d_x\sigma_\tau^2)\mathrm{d}\tau 
\leq c^0_b\log^2 K,
\label{l-bound}
\end{align}
where $c^0_b=2\frac{d_x(\frac{\pi B^2}{\underline{c}^2}\max(1,1/c_f)+1)(c_{\overline{\tau}}+c_{\underline\tau})}{1-e^{-1}}+4\max(1,c_f^2)d_xc_R+4d_xc_{\overline{\tau}}$.

Next, we verify the variance condition. By the Cauchy-Schwarz inequality, 
\begin{align*}
&l(\x,\z,\theta)-l(\x,\z,\theta^0)\\
 &\leq \left[\int_{{\underline{\tau}}}^{\overline{\tau}}\mathrm{E}_{\x(\tau)|\x(0)}\|2\nabla \log p_{\tau}(\X|\x(0),\z)-\theta(\X,\z,\tau)-\theta^0(\X,\z,\tau)\|^2\mathrm{d}\tau \right.\\ &\qquad \left.\int_{{\underline{\tau}}}^{\overline{\tau}}\mathrm{E}_{\x(\tau)|\x(0)}\|\theta(\X,\z,\tau)-\theta^0(\X,\z,\tau)\|^2\mathrm{d}\tau\right]\\
 &\leq  \sup_{\x,\z}(l(\x,\z,\theta)+l(\x,\z,\theta^0))\int_{{\underline{\tau}}}^{\overline{\tau}}\mathrm{E}_{\x(\tau)|\x(0)}\|\theta(\X,\z,\tau)-\theta^0(\X,\z,\tau)\|^2\mathrm{d}\tau.
\end{align*}
By \eqref{eq-cb*} and \eqref{l-bound} and the sub-Gaussian property of $p^0_{x}$ with a sufficiently large $c_R$, 
we have
\begin{align*}
    &\Var_{\x(0),\z}(l(\X(0),\Z;\theta)-l(\X(0),\Z;\theta^0))
    \leq \mathrm{E}_{\x(0),\z}(l(\X(0),\Z;\theta)-l(\X(0),\Z;\theta^0))^2 \\
\leq &
\sup_{\|\x\|_{\infty}\leq c_R\sqrt{\log K},\z} (l(\x,\z;\theta)+l(\x,\z;\theta^0))
\int_{{\underline{\tau}}}^{\overline{\tau}}\mathrm{E}_{\x,\z} \mathrm{E}_{\x(\tau)|\x}\|\theta(\X,\Z,\tau)-\theta^0(\X,\Z,\tau)\|^2\mathrm{d}\tau
\\\leq & (c^0_b+c^*_b) (\log^2K) \rho^2(\theta^0,\theta),
\end{align*}
This implies Assumption \ref{Variance} with $c_v=O(\log^2 K)$. 

By Lemma \ref{thm_approx}, the approximation error is $\delta_n=O(K^{-\frac{r}{d_x+d_z}}
\log^{\frac{r}{2}+1}K)$. By Lemma \ref{thm_entropy}, there exists a positive $c_H$ such that $H(u,\smc{F})\leq c_H K\log^{14}K\log\frac{Kn}{u}$. Then,
\begin{align*}
 \int_{k\varepsilon^{2}/16}^{4 c^{1/2}_v \varepsilon}
H_B^{1/2}(u,\smc{F}) \mathrm{d}u &\leq c_H\int_{k\varepsilon^{2}/16}^{4c^{1/2}_{v} \varepsilon} 
K\log^{16}K\log\frac{K}{u}\mathrm{d}u
 \leq 4c_H c^{1/2}_v\varepsilon \sqrt{ K\log^{16}K\log\frac{K}{\varepsilon^2}}.
\end{align*}

Solving the inequality for $\beta_n$: $4c_H c^{1/2}_v \beta_n \sqrt{ K\log^{16}K\log\frac{K}{\varepsilon^2}}\leq c_h\sqrt{n}\beta_n^2$ with $c_{v}=2(c^*_b+c^0_b)\log^2K$ yields $\beta_n=2\frac{4c_H\sqrt{2(c^*_b+c^0_b)}}{c_h}\sqrt{\frac{K\log^{19} K}{n}}$.

Let $\varepsilon_n=\beta_n+\delta_n$ with $\beta_n\asymp\sqrt{\frac{K\log^{19} K}{n}}$ and $\delta_n\asymp K^{-\frac{r}{d_x+d_z}}
\log^{\frac{r}{2}+1}K$ so that $\varepsilon_n$ satisfies the conditions of Proposition \ref{prop1}. 
Omitting the logarithmic term, let $\beta_n=\delta_n$ by $K\asymp n^{-\frac{d_x+d_z}{d_x+d_z+2r}}$ and this yields the optimal rate $\varepsilon_n\asymp n^{\frac{r}{d_x+d_z+2r}}\log^m n$ with $m=\max(\frac{19}{2},\frac{r}{2}+1)$. 
Note that $c_e\asymp\frac{1}{c^2_v}\asymp \log^4 K$ in Proposition \ref{prop1}.
 With $\log K=O(\log n)$, for some constant $c_e>0$,
$
P(\rho(\theta^0,\hat \theta) \geq \varepsilon_n )\leq 4\exp(-c_en^{1-\xi}\varepsilon_n^2 ),
$
where $0<\xi<1$ is small such that $\frac{\log^2n}{n^{\xi}}=o(1)$ and $1-\xi-\frac{2r}{d_x+d_z+2r}>0$, implying that
$\rho(\theta^0,\hat \theta)=O_p(\varepsilon_n)$ as $n^{1-\xi}\varepsilon_n^2 \rightarrow
\infty$ as $n \rightarrow \infty$. By Lemma \ref{l-KL-Score}, the desired result follows from \eqref{c-lemma5}.
This completes the proof. 
\end{proof}

\begin{lemma}[KL divergence and score matching loss]
\label{l-KL-Score}
Under the assumptions and settings in Theorem \ref{thm_diff_general}, suppose that the excess risk is bounded in that $\rho(\theta^0,\hat{\theta}) \leq \varepsilon_n$ with high probability. Then, the diffusion generation errors under the total variation distance and the square root of the Kullback-Leibler divergence are also bounded by $\varepsilon_n$ with constants $c_{TV}$ and $c_{KL}$, respectively:
\begin{eqnarray}
\label{c-lemma5}
& \mathrm{E}_{\z}[\mathrm{TV}(p^0_{\x|\z},\hat{p}_{\x|\z})]\leq c_{TV}\varepsilon_n, 
\text{ if $r>0$; }  \mathrm{E}_{\z}[\smc{K}^{1/2}(p^0_{\x|\z},\hat{p}_{\x|\z})]\leq c_{KL}\varepsilon_n, 
\text{ if $r>1$.}
\end{eqnarray}
\end{lemma}
\begin{proof}
The first inequality in \eqref{c-lemma5} follows from Lemma D.5 in
\cite{fu2024unveil}. For the second inequality, by Girsanov’s Theorem and 
Proposition C.3 in \cite{chen2023improved}, the KL
divergence can be bounded by the diffusion approximation to the standard
Gaussian and score matching:
\begin{align}
\label{loss-bound}
 &  \smc{K}(p^0_{\x|\z},\hat p_{\x|\z})\leq \smc{K}(p_{\overline{\tau}},p_{N})+I_1(\z)+I_2(\z), \nonumber \\
& I_1(\z)= \int_{{\underline{\tau}}}^{\overline{\tau}} \frac{1}{2} \mathrm{E}_{\x(\tau)|\z} \, \| \nabla \log p_{\tau}(\x|\z) - \hat{\theta}(\x,\z,\tau) \|^{2}  \mathrm{d}\tau \nonumber, \\
& I_2(\z)=
\int_0^{\underline{\tau}}
    \frac{1}{2} \mathrm{E}_{\x(\tau),\x(\underline{\tau})|\z}\,  \| \nabla \log p_{\tau}(\x|\z) -  \hat{\theta}(\x(\underline{\tau}),\z,\underline{\tau})\|^{2}  \mathrm{d}\tau, 
\end{align}
where $p_N$ denotes the standard $d_x$-dimensional Gaussian density. 

By Lemma C.4 of \cite{chen2023improved}, 
$\smc{K}(p_{\overline{\tau}},p_{N})\leq (d_x+\mathrm{E}_{\x}\|\X\|^2)\exp{(-\overline{\tau})}$
, which is bounded through the exponential convergence of the forward diffusion process.
The second term is bounded by the $L_{2}$-distance of the score function $\mathrm{E}_{\z}I_1(\z) =O(\rho^2(\theta^0,\hat{\theta}))$.
% and third terms are bounded by the $L_{2}$-distance of the score function and the continuity property in Lemma \ref{l-lowertau}:
% \begin{align*}
%  \mathrm{E}_{\z}I_1(\z) =O(\rho^2(\theta^0,\hat{\theta})), \quad 
%    \mathrm{E}_{\z}I_2(\z) =O(\frac{\underline\tau}{\sigma^2_{\underline\tau}}\varepsilon_n^2)+O({\underline{\tau}}^{1+\alpha}).
% \end{align*}

To bound $E_{\z} I_2(\z)$, by the pointwise $L_2$-distance for $\tau$ in Theorem 3.4 of 
\cite{fu2024unveil}, we obtain that  
\begin{align*}
 \int_{\R^{d_x}} p_{\underline{\tau}}(\x|\z) \| \nabla \log p_{\underline{\tau}}(\x|\z) -  \pi\theta_0(\x,\z,\underline{\tau})\|^{2} \mathrm{d}\x = O(\frac{1}{\sigma^2_{\tau}}K^{-\frac{r}{d_x+d_z}}
\log^{\frac{r}{2}+1}K)=O(\frac{1}{\sigma^2_{\tau}}\varepsilon^2).
\end{align*}

To get the estimation error for the point score matching loss at $\underline{\tau}$, we repeat the process to bound $\beta_n$ in the proof of Theorem \ref{thm_diff_general} while modifying the integral upper bounds in \eqref{eq-cb*} and \eqref{l-bound} by replacing $c_b^*$ with $\frac{c_b^*}{\sigma^2_{\underline{\tau}}}$ and $c_b^0$ with $\frac{c_b^0}{\sigma^2_{\underline{\tau}}}$. 

Moreover, together with  the trade-off in the setting, we can bound the total error in the same order of the approximation error, for some constant $c>0$,
\begin{align*}
 \int_{\R^{d_x}} p_{\underline{\tau}}(\x|\z) \| \nabla \log p_{\underline{\tau}}(\x|\z) -  \hat{\theta}(\x,\z,\underline{\tau})\|^{2} \mathrm{d}\x = O(\frac{1}{\sigma^2_{\underline{\tau}}}\varepsilon^2).
\end{align*}
Then 
\begin{align*}
\mathrm{E}_{\z}I_2(\z)\leq &\frac{\underline{\tau}}{2} \mathrm{E}_{\z}\int_{\R^{d_x}} p_{\underline{\tau}}(\x|\z) \| \nabla \log p_{\underline{\tau}}(\x|\z) -  \hat{\theta}(\x,\z,\underline{\tau})\|^{2} \mathrm{d}\x\\
&+ \mathrm{E}_{\z} \int_0^{\underline{\tau}}
    \frac{1}{2} \mathrm{E}_{\x(\tau),\x(\underline{\tau})|\z}\, \| \nabla \log p_{\tau}(\x(\tau)|\z) -  \log p_{\underline{\tau}}(\x(\underline{\tau})|\z)\|^{2}  \mathrm{d}\tau\\
=& O(\frac{\underline{\tau}}{\sigma^2_{\underline{\tau}}}\varepsilon^2)+O({\underline{\tau}}^{1+\alpha}).
\end{align*}
Finally, in \eqref{loss-bound}, by choosing $\underline{\tau}=K^{-c_{\underline\tau}}$ and $\overline{\tau}=c_{\overline{\tau}}\log K$ with sufficiently large $c_{\underline\tau}$ and $c_{\overline\tau}$ such that $\exp{(-\overline{\tau})}+{\underline{\tau}}^{1+\alpha}\leq \varepsilon_n^2$, we obtain $\smc{K}(p^0_{\x|\z},\hat p_{\x|\z})=O(\varepsilon^2_n)$ with high probability. This completes the proof.     
\end{proof}

\subsubsection{Proofs for Subsection \ref{sec_3-2}}
Theorem \ref{thm_diff-detail} gives the non-asymptotic probability bound for the generation error in Theorem \ref{thm_diff}.
\begin{theorem}[Conditional diffusion via transfer learning]
\label{thm_diff-detail}
Under Assumptions \ref{A-independent}-\ref{A-error} and \ref{A_p0}, there exists a wide ReLU network $\Theta_t$ in \eqref{p-space}, with specified hyperparameters: $\mathbb{L}_t = c_L \log^4 K$, $\mathbb{W}_t = c_W K \log^7 K$, $\mathbb{S}_t = c_S K \log^9 K$, $\log \mathbb{B}_t = c_B \log K$, $\log \mathbb{E}_t = c_E \log^4 K$,
such that the error in conditional diffusion generation via transfer learning,
as described in Section \ref{sec_3-2}, with stopping criteria: $\log \underline{\tau}_t = -c_{\underline{\tau}_t}\log K$ and $\overline{\tau}_t = c_{\overline{\tau}_t}\log K$ in \eqref{forward} and \eqref{reverse}, 
is given by: for any $x\geq 1$,
\begin{eqnarray*}
\mathrm{E}_{\z_t}[\mathrm{TV}(p^0_{\x_t|\z_t},\hat p_{\x_t|\z_t})]\leq x(\varepsilon_t+\sqrt{3c_1}\varepsilon_s),r_t>0;\\
\mathrm{E}_{\z_t}[\smc{K}^{\frac{1}{2}}(p^0_{\x_t|\z_t},\hat p_{\x_t|\z_t})]\leq x(\varepsilon_t + \sqrt{3c_1}\varepsilon_s),r_t>1, 
\end{eqnarray*}
with a probability exceeding $1-\exp(-c_5 n^{1-\xi}_t (x\varepsilon_t)^2) - \exp\left(-c_2 n^{1-\xi}_s (x\varepsilon_s)^2\right)$ for some constant $c_5>0$. Here $c_L, c_W, c_S, c_B, c_E, c_{\overline{\tau}_t}, c_{\underline{\tau}_t}$ are sufficiently large constants, $K \asymp n_t^{\frac{d{x_t} + d_{h_t}}{d{x_t} + d_{h_t} + 2r_t}}$, 
$\varepsilon_t\asymp n_t^{-\frac{r_t}{d_{x_t}+d_{h_t}+2r_t}}\log^{m_t} n_t$, and $m_t = \max(\frac{19}{2}, \frac{r_t}{2}+1)$, with $\asymp$ denoting mutual boundedness.
\end{theorem}

\begin{proof}[Theorems \ref{thm_diff} and \ref{thm_diff-detail}]
If the probability bound in Theorem \ref{thm_diff-detail} holds, we obtain
the rate in expectation in Theorem \ref{thm_diff}. Specifically, let $\mathrm{TV}$ be
$\mathrm{E}_{\z_t}[\mathrm{TV}(p^0_{\x_t|\z_t},\hat p_{\x_t|\z_t})]$. Then, 
\begin{align*}
\mathbb{E}_{\smc{D}}\frac{\mathrm{TV}}{\varepsilon_t}&=\frac{1}{\varepsilon_t}\left(\mathbb{E}_{\smc{D}}\mathrm{TV}\mathbb{I}(\mathrm{TV}\leq\varepsilon_t)+\mathbb{E}_{\smc{D}}\mathrm{TV}\mathbb{I}(\mathrm{TV}>\varepsilon_t)\right)\nonumber\\
&\leq 1+\left(\int_1^{\infty} P(\mathrm{TV}>x\varepsilon_t)\mathrm{d}x+\frac{1}{\varepsilon_t}P(\mathrm{TV}>\varepsilon_t)\right) 
\leq 2.
\end{align*}
The last inequality holds by the fact that $\xi$ can be small enough.
The rate of expected KL divergence can be proved in a similar way.
 
 To prove Theorem \ref{thm_diff-detail}, we first show that, under the conditions of Theorem \ref{thm_diff}, the error bound for the excess risk holds, for any $x>1$,
\begin{eqnarray*}
\rho_{t}(\gamma_t^0, \hat \gamma_t)\leq x(\beta_t + \delta_t +\sqrt{3 c_1}\varepsilon_s),
\end{eqnarray*}
with a probability exceeding $1-\exp(-c_5 n^{1-\xi}_t (x(\beta_t + \delta_t))^2) - \exp\left(-c_2 n^{1-\xi}_s (x\varepsilon_s)^2\right)$. Here, the estimation and approximation errors $\beta_t$ and $\delta_t$ are $\beta_t\asymp 
    \sqrt{\frac{K\log^{19} K}{n_t}}$ and $\delta_t\asymp K^{-\frac{r_t}{d_{x_t}+d_{h_t}}}\log^{\frac{r_t}{2}+1}K$.
    This bound is proved using Theorem \ref{theorem1} with the approximation error and the entropy bounds obtained in Lemmas \ref{thm_approx} and \ref{thm_entropy}.
    The two assumptions on the loss function in Theorem \ref{theorem1} can be verified similarly as in the proof of Theorem \ref{thm_diff_general}. 
Hence, we only need to obtain the approximation error $\delta_t$ and the estimation error $\beta_t$.

    Recall the definition of $\delta_t$, $\delta_t=\inf_{\theta_t\in \Theta_t}\E [l_j(\cdot;\theta_t,h^0)-l_j(\cdot;\theta_t^0,h^0)]$. By Assumptions \ref{A_p0}, we bound $\delta_t$ by Lemma \ref{thm_approx} by replacing $\z$ with the $d$-dimensional $h^0(\z)$:
     there exists a ReLU network $\mathrm{NN}_t(\mathbb{L}_t,\mathbb{W}_t,\mathbb{S}_t,\mathbb{B}_t,\mathbb{E}_t)$ with depth $\mathbb{L}_t\asymp\log^4K$, width $\mathbb{W}_t\asymp K\log^{7} K$, effective parameter number $\mathbb{S}_t\asymp K\log^9K$, parameter bound $\log\mathbb{E}_t\asymp\log^4K$, and $\sup_{\x,\z}\|\theta_t(\x,\z,\tau)\|_{\infty}\asymp\sqrt{\log K}/\sigma_\tau$,
such that
$\rho(\theta_t^0,\pi \theta^0_t) =O(K^{-\frac{r}{d_{x_t}+d_{h_t}}}\log^{r_t/2+1}K).
$
For the estimation error $\beta_t$, we apply Lemma \ref{thm_entropy} by replacing the parameters of $\smc{F}$ by those of $\smc{F}_t$.

Setting $\beta_t = \delta_t$ for $K$, ignoring the logarithmic term, yields
$\beta_t = \delta_t \asymp n_t^{-\frac{r_t}{d_{x_t}+d_{h_t}+2r_t}}\log^{m_t} n_t$ with optimal $K \asymp n_t^{\frac{d_{x_t}+d_{h_t}}{d_{x_t}+d_{h_t}+2r_t}}$ and $m_t = \max(\frac{19}{2}, \frac{r_t}{2}+1)$. Thus, the best bound is $\varepsilon_t\asymp n_t^{-\frac{r_t}{d_{x_t}+d_{h_t}+2r_t}}\log^{m_t} n_t$. 
\end{proof}

For the source task, we use the conditional diffusion model to learn $p_{\x_s|\z_s}$. This can be done similarly as estimating $\theta_t$ for $p_{\x_t|\z_t}$. Given $\Theta_s=\{\theta_s(\x,h(\z),\tau),\theta_s\in \mathrm{NN}_s(\mathbb{L}_s,\mathbb{W}_s,\mathbb{S}_s,\mathbb{B}_s,\mathbb{E}_s,\lambda_s)\}$, we define the loss as 
\begin{align*}
    L_{s}(\theta_s,h)&=\sum_{i=1}^{n_s} l_s(\x_s^i,\z_s^i;\theta_s,h)\\
    &=\sum_{i=1}^{n_s}
\int_{\underline{\tau}_s}^{\overline{\tau}_s} \mathrm{E}_{\x(\tau)|\x(0),\z_s}\|\nabla \log p_{\x(\tau)|\x(0),\z_s}(\X(\tau)|\x_s^i,\z_s^i)-\theta_s(\X(\tau),h_s(\z_s^i),\tau)\|^2 \mathrm{d}\tau,
\end{align*}
which
yields that $
\hat\theta_s(\x,\hat h(\z),\tau)=\argmin_{\theta_s\in \Theta_s, h\in \Theta_h} L_s(\theta_s,h)$.

To give the bound of $\varepsilon_s$ in Section 3, we make some assumptions similar to those for $p_{\x_t|\z_t}$.

\begin{assumption}[Source density]
\label{A2_p0}
Assuming the true conditional density of $\X_s$ given $\Z$ is expressed as $p^0_{\x_s|\z_s}(\x|\z_s) = \exp(-c_6\|\x\|^2 / 2) \cdot f_s(\x,h_s^0(\z_s))$, where $f_s$ is a non-negative function and $c_6>0$ is a constant. Additionally, $f_s$ belongs to
a H\"older ball $\smc{C}^{r_s}(\mathbb{R}^{d_{x_s}} \times [0,1]^{d_{h_s}},\R^{d_{x_s}}, B_s)$ and is lower bounded away from zero. 
\end{assumption}

\begin{lemma}[Source error]
\label{thm_diff_source}
Under Assumption \ref{A2_p0},  setting a network  $\Theta_s$'s hyperparameters with sufficiently large constants $\{c_L,c_W, c_S, c_B,c_E,c_{\lambda}, c_{\underline{\tau}_s},c_{\overline{\tau}_s}\}$: $\mathbb{L}_s=c_L \log^4K$, $\mathbb{W}_s= c_W K\log^7K$, $\mathbb{S}_s= c_S K\log^9K$, 
$\log\mathbb{B}_s= c_B\log K$, $\log\mathbb{E}_s= c_E\log^4K$,
and $\lambda_s=c_{\lambda}$ for $r>1$, $\lambda_s= c_{\lambda }/\sigma_{\underline{\tau}}$ for $r\leq 1$,
with diffusion stopping criteria from \eqref{forward}-\eqref{reverse} as $\log \underline{\tau}_s= -c_{\underline{\tau}_s}\log K$ and $\overline{\tau}_s=c_{\overline{\tau}_s}\log K$, we obtain that, for any $x\geq 1$,
\begin{equation*}
P(\rho_{s}(\gamma_s^0, \hat \gamma_s) \geq x\varepsilon_s)\leq \exp(-c_2 n_s^{1-\xi}(x\varepsilon_s)^2),
\end{equation*}
with $\varepsilon_s=\beta_s+\delta_s+\varepsilon^h_s$,
some constant $c_2>0$  and a small $\xi>0$ same in Assumption \ref{A-error}. Here, $\beta_s$ and $\delta_s$ are given by:
$\beta_s\asymp 
    \sqrt{\frac{K\log^{19} K}{n_s}},\quad \delta_s\asymp K^{-\frac{r_s}{d_{x_s}+d_{h_s}}}\log^{{r_s}/2+1}K$.
$\varepsilon^h_s$ satisfies \eqref{eq-entropy-h} in Lemma \ref{l-composite-entroty}.
Setting $K\asymp n_s^{\frac{d_{x_s}}{d_{x_s}+d_{h_s}+2 r_s}}$ yields
$\varepsilon_s\asymp n_s^{-\frac{r_s}{d_{x_s}+d_{h_s}+2r_s}} \log^{m_s} n_s+\varepsilon^h_s$,
where $m_s = \max(\frac{19}{2}, \frac{r_s}{2}+1)$. 
\end{lemma}
\begin{proof}
    When $h_0\in\Theta_h$, $\delta_s\leq\delta_s(h^0)$. Hence, the upper bound can be given by $\delta_s(h^0)$ which is given by Lemma \ref{thm_approx}. Note that with the H\"older continuity in $\Theta$, the statistical error can be decomposed into $\beta_s+\varepsilon_s^h$
    through Lemma \ref{l-composite-entroty}. The bound of $\varepsilon_s^h$ is given by \eqref{eq-entropy-h} in Lemma \ref{l-composite-entroty} and the bound of $\varepsilon_s$ is derived in the same way as $\varepsilon_t$ in the proof of Theorem \ref{thm_diff_general}.
\end{proof}

Theorem \ref{cor-nt-detail} gives the non-asymptotic probability bound for the generation error in Theorem \ref{cor-nt}.
\begin{theorem}[Non-transfer conditional diffusion]
\label{cor-nt-detail}
Under Assumption \ref{A_p0}, a wide ReLU network $\tilde\Theta_t$, as described in \eqref{p-space} and with the same configuration as $\Theta_t$ from Theorem \ref{thm_diff}, exists, with an additional constraint on $\tilde\Theta_t$: $\lambda_t=c_{\lambda}$ for $r>1$ and $\lambda_t=c_{\lambda}/\sigma_{\underline{\tau}}$ for $r\leq 1$, provided $c_{\lambda}$ is sufficiently large. Then, the generation error of the non-transfer conditional diffusion model, adhering to the same stopping criteria from Theorem \ref{thm_diff}, is given by: for any $x\geq 1$
\begin{eqnarray*}
\mathrm{E}_{\z_t}[\mathrm{TV}(p^0_{\x_t|\z_t},\tilde p_{\x_t|\z_t})]\leq x(\tilde\varepsilon_t + \varepsilon^h_t), \text{ if }r>0,
\\
\mathrm{E}_{\z_t}[\smc{K}^{1/2}(p^0_{\x_t|\z_t},\tilde p_{\x_t|\z_t})]\leq x(\tilde\varepsilon_t +  \varepsilon^h_t ), \text{ if } r>1,
\end{eqnarray*} 
with the target probability exceeding $1-\exp(-c_5 n^{1-\xi}_t (x(\tilde\varepsilon_t + \varepsilon^h_t))^2)$ for some constant $c_5>0$. Here, $\tilde\varepsilon_t\asymp n_t^{-\frac{r_t}{d_{x_t}+d_{h_t}+2r_t}}\log^{m_t} n_t$ while $\varepsilon^h_t$ is defined by
\eqref{eq-entropy-h} in Lemma \ref{l-composite-entroty}. 
\end{theorem}

\begin{proof}[Theorems \ref{cor-nt} and \ref{cor-nt-detail}]
     Theorem \ref{cor-nt-detail} can be proved similarly to Lemma \ref{thm_diff_source} by replacing $r_s$ with $r_t$, $n_s$ with $n_t$ and $d_{x_s}$ with $d_{x_t}$.
\end{proof}
Next, we give the metric entropy inequalities to the class of composite functions $\Theta_j\circ\Theta_h=\{\theta_j(\x_j,h_j(\z_j)),h_j(\z_j)=(h(\z),\z_{j^c}), h\in \theta_h,\theta_j\in \Theta_j\}$, $j \in \{s,t\}$. 
\begin{lemma}
\label{l-composite-entroty}
When $\sup_{\theta_j\in\Theta_j} \sup_{\x,\tau,\y\neq\y'}\frac{\|\theta_j(\x,\y,\tau)-\theta_j(\x,\y',\tau)\|}{\|\y-\y'\|^{\alpha_j}}\leq\lambda_{j}$, for $j \in \{s,t\},$
\begin{equation}
\label{eq-entropy-h}
\int_{k_j(\varepsilon^{h}_j)^{2}/16}^{4 c^{1/2}_{vj}(\varepsilon^{h}_j)}
H^{1/2}\left(\frac{u^{\frac{1}{\alpha_j}}}{2\lambda_{l_j}\lambda_{j}(d_{\z})^{\frac{\alpha_j}{2}}},\Theta_h\right) \mathrm{d}u \leq c_{h_j} n_j^{1/2}(\varepsilon^h_{j})^{2},
\end{equation} 
where $\lambda_{l_j}$ is the Lipschitz norm of $l_j$ with respect to $\theta_j$ and $\lambda_{j}$ and $\alpha_j$ are the parameter settings in the neural network class.
\end{lemma}

\begin{proof}
For $\gamma_j, \gamma'_j \in \Theta_j\circ\Theta_h$, 
\begin{align*}
\sup_{\x_j,\z_j}&|l_j(\x_j,\z_j;\gamma_j)-l_j(\x_j,\z_j;\gamma'_j)|\leq 
\sup_{\x_j,\z_j}
\lambda_{l_j}\|\theta_j(\x_j,h_j(\z_j))-\theta'_j(\x_j,h'_j(\z_j))\|_{\infty}\\
&\leq\sup_{\x_j,\z_j}
\lambda_{l_j}\|\theta_j(\x_j,h_j(\z_j))-\theta'_j(\x_j,h_j(\z_j))\|_{\infty}+\sup_{\x_j,\z_j}
\lambda_{l_j}\|\theta'_j(\x_j,h_j(\z_j))-\theta'_j(\x_j,h'_j(\z_j))\|_{\infty}
\\
&\leq \sup_{\x_j,\y} \lambda_{l_j}\|\theta_j(\x_j,\y)-\theta'_j(\x_j,\y)\|_{\infty}+ \sup_{\z} \lambda_{l_j}\lambda_j d_{\z}^{\frac{\alpha_j}{2}}\|h(\z)- h'(\z)\|^{\alpha_j}_{\infty}.
\end{align*}
The last inequality follows from the H\"older continuity of $\theta$ and the definition of $h_j$, $h_j(\z_j)=(h(\z),\z_{j^c})$ for $j \in \{s,t\}$. Hence,
$H_B^{1/2}(u,\smc{F}_s)\leq H^{1/2}(c'_{\theta}u,\Theta)+H^{1/2}(c'_{h}u^{\frac{1}{\alpha_j}},\Theta_h) $, where $c'_{\theta_s}=\frac{1}{2\lambda_{l_s}}$ and $c'_{h}=\frac{1}{2\lambda_{l_s}\lambda_{s}(d_{\z})^{\frac{\alpha_j}{2}}}$.
So, the integral inequality \eqref{entropy-t3} is solved by $\beta_s$ defined in Lemma \ref{thm_diff_source} and $\varepsilon_s^h$ in \eqref{eq-entropy-h}.
The same holds for $\Tilde{\smc{F}}_t$  in \eqref{entropy-t4} with $\beta_t$ in the proof of Theorem \ref{cor-nt-detail} and $\varepsilon_t^h$.

By the proof of Lemma \ref{thm_entropy}, we have $\lambda_{l_j}=c^*\log^{3/2}(n_j)$ in the conditional diffusion models.
\end{proof}

\subsubsection{Proofs of Subsection \ref{sec_3-3}}
We first give the formal version of Theorems \ref{thm_ug} and \ref{cor-nt2} in 
non-asymptotic probability bounds.
\begin{theorem}[Unconditional diffusion via transfer learning]
\label{thm_ug_detail}
Under Assumptions \ref{A_g} and \ref{U-error}, there exists a 
ReLU network $\Theta_{g_t}$ with hyperparameters:
$\mathbb{L}_g=c_L L\log L$, $\mathbb{W}_g= c_W W\log W$, $\mathbb{S}_g= c_S W^2 L\log^2W \log L$, 
$\mathbb{B}_g= c_B$, and $\log\mathbb{E}_g= c_E\log(WL)$, such that the error for unconditional diffusion generation via transfer learning in the Wasserstein distance is
$
\smc{W}(p^0_{\x}, \hat{p}_{\x})\leq x(\varepsilon_t+\varepsilon_s^u)
$
with a probability exceeding $1-\exp(-c_3 n^{1-\xi}_s(x\varepsilon_s^u)^2)-\exp(-c_7 n_t (x\varepsilon_t)^2)$, for any $x\geq 1$ and some constant $c_7>0$. Here, $\{c_L,c_W, c_S, c_B,c_E \}$ are sufficiently large positive constants, $WL\asymp n_t ^{\frac{d_u}{2(d_u+2r_g)}}$, 
$\varepsilon_t\asymp n_t ^{-\frac{{r_g}}{d_u+2{r_g}}}  \log^{m_g} n_t$, and  $m_g=\max(\frac{5}{2},\frac{r_g}{2})$.
\end{theorem}

\begin{theorem}[Non-transfer via unconditional diffusion]
\label{cor-nt2-detail}
Suppose there exists a sequence $\varepsilon_t^u$ indexed by $n_t$ such that $n_t^{1-\xi}(\varepsilon_t^u)^2\rightarrow\infty$ as $n_s\rightarrow\infty$ and
$P(\rho_u(\theta_u^0,\tilde\theta_u)\geq \varepsilon)\leq \exp(-c_3 n_t^{1-\xi} \varepsilon^2)$ for any $\varepsilon\geq \varepsilon_t^u$ and some constants $c_3, \xi>0$.  Under Assumption \ref{A_g} and the same settings for $\Theta_{g_t}$ of Theorem \ref{thm_ug}, the generation error of the non-transfer unconditional diffusion model satisfies: for any $x\geq 1$
$
\smc{W}(P^0_{\x_t}, \tilde{P}_{\x_t})\leq x(\varepsilon_t+\varepsilon_t^u),
$
with a probability exceeding $1-\exp(-c_3 n_t^{1-\xi} (x\varepsilon_t^u)^2)-\exp(-c_7 n_t(x\varepsilon_t)^2)$. Here $\varepsilon_t\asymp n_t ^{-\frac{{r_g}}{d_u+2{r_g}}}\log^{m_g} n_t$.
\end{theorem}

Referencing the definition of excess risk, let $\gamma_t=(g_{t},\theta_u)$. Define the excess risks as $\rho^2_{g_{t}}(g_t^0, g_t)=\mathrm E_{\bm u} (l_{g_{t}}(\U,\X;g_t)-l_{g_{t}}(\U,\X;g_t^0))$ and $\rho^2_{u}(\theta_u,\theta_u^0)=\mathrm E_{\bm u} (l_u(\U;\theta_u)-l_u(\U;\theta_u^0))$, for $g_{t}$ and $\theta_u$, respectively. The total excess risk is denoted as $\rho_{t}^2(\gamma_t,\gamma_t^0)=\rho^2_{g_{t}}(g_{t}^0, g_{t})+\rho^2_{u}(\theta_u,\theta_u^0)$. We adopt the same notation except for the loss functions used in this subsection.

Theorem \ref{thm_ug} and Theorem \ref{cor-nt2} can be obtained directly using the bounds Lemma \ref{thm_g} and converting the excess risk bound to the Wasserstein distance by Lemma \ref{cor-w}.
\begin{lemma}[Estimation error for the mapping $g_t$]
\label{thm_g}
Under Assumption \ref{A_g}, setting the hyperparameters of the neural network $\Theta_g$' with a set of sufficiently large positive constants $\{c_L,c_W, c_S, c_B,c_E \}$ such that $\mathbb{L}_g=c_L L\log L$, $\mathbb{W}_g= c_W W\log W$, $\mathbb{S}_g= c_S W^2 L\log^2W \log L$, 
$\mathbb{B}_g= c_B$, and $\log\mathbb{E}_g= c_E\log(WL)$, with $K=WL$, yields the $L_2$ approximation error: for any $x\geq 1$
\begin{eqnarray*}
\rho_{g_t}(g^0_t,\hat{g}_t)\leq x(\beta_g+\delta_g), 
\end{eqnarray*}
with a probability exceeding $1-\exp(-c_7 n_t(x(\beta_g+\delta_g))^2)$.
Here, the estimation error $\beta_g$ and the approximation error $\delta_g$ are bounded by
$ \beta_g\asymp 
    \sqrt{\frac{K^2\log^5 K}{n_{t}}},\quad \delta_g\asymp K^{\frac{-2{r_g}}{d_u}}\log^{\frac{r_g}{2}} K$.
To obtain the optimal trade-off, we set $\beta_g=\delta_g$ to determine $K$,
after ignoring the logarithmic term, the optimal bound is obtained by $K^2\asymp n_{t}^{\frac{d_u}{2(d_u+2{r_g})}}$. This yields
$\rho_{g_t}(g^0_t,\hat{g}_t) =O_p(n_{t}^{-\frac{r_g}{d_u+2{r_g}}} \log^{m_g} n_{t})$, where $m_g=\max(\frac{5}{2},\frac{r_g}{2})$.
\end{lemma}
\begin{proof}
Using the sub-Gaussian property of $\bm U$ in Assumption \ref{A_g}, we focus our attention on $\bm U \in \smc{B}=[-c_u \sqrt{\log (WL)},c_u \sqrt{\log (WL))}]^{d_u}$ by truncation for some sufficiently large $c_u>0$. Note that $g_t^0$ is bounded by 
$B_g$ by Assumption \ref{A_g} and  
$\sup_{g_t\in \Theta_g,\bm u} \|g_t(\bm u)\|_{\infty}\leq \mathbb{B}_g$ in the setting of $\Theta_g$. Then, by
choosing sufficiently large $c>0$ so that 
\begin{equation}
\label{eq-sub-g}
\int_{\bm u\in \R^{d_u}/\smc{B}}
\|g_t(\bm u)-g^0_t(\bm u)\|^2p_u(\bm u)\mathrm{d}\bm u = O((WL)^{-\frac{4r_g}{d_u}}). 
\end{equation}
We transform $\bm u$ from $I$ into $[0,1]^{d_u}$ and apply Lemma \ref{l_approx_nn}. Specifically, let $\y=\frac{\bm u+c_u \sqrt{\log (WL)}}{2c_u \sqrt{\log (WL)}}$ and $\bar{g}^0_t(\y)=g_t(\bm u)$, , which changes the H\"older-norm $B_g$ to $(2c_u)^{r_g}\log^{\frac{r_g}{2}}(WL)B_g$.
Then there exists an NN $\phi$ with depth $L\log L $ and width $W\log W$ such that
\begin{equation*}
\sup_{\y\in [0,1]^{d_u}}\|\phi(\y)-\bar{g}^0_t(\y)\|_{\infty}=O(\log^{\frac{r_g}{2}}(WL) (WL)^{-\frac{2r_g}{d_u}}).   
\end{equation*}
Then, we let $\pi g_t^0(\bm u)=\phi(\frac{\bm u+c_u \sqrt{\log (WL)}}{2c_u \sqrt{\log (WL)}})$ and obtain the bound
$
\sup_{\bm u\in \smc{B}}\|\pi g^0_t(\bm u)-g^0_t(\bm u)\|_{\infty}=O(\log^{\frac{r_g}{2}}(WL) (WL)^{-\frac{2r_g}{d_u}}),
$ 
which implies
\begin{equation*}
\int_{\smc{B}}
\|\pi g^0_t(\bm u)-g^0_t(\bm u)\|^2p_u(\bm u)\mathrm{d}\bm u \leq 
\sup_{\bm u\in \smc{B}}\|\pi g^0_t(\bm u)-g^0_t(\bm u)\|^2_{\infty}=O(\log^{r_g}(WL) (WL)^{-\frac{4r_g}{d_u}}).
\end{equation*}
By \eqref{eq-sub-g}, $\delta_g=O(\log^{\frac{r_g}{2}}(WL) (WL)^{-\frac{2r_g}{d_u}})$.  The estimation error $\beta_g$ is derived as in Lemma \ref{thm_entropy} with $H(u,\{l_{g_t}(\cdot;g_t),g_t\in\Theta_{g_t}\})\asymp H(u,\Theta_{g_t}\})=O(\mathbb{S}\mathbb{L}\log(\mathbb{E}\mathbb{W}\mathbb{L}R/u))$ where $R=c_u \sqrt{\log (WL)}$ by the truncation.
By the boundedness of $g_t^0$ and $g_t\in\Theta_{g_t}$, the square loss satisfies Assumptions \ref{Variance} and \ref{sub-Gaussian} as in the proof of Theorem \ref{thm_diff_general} for the score matching loss. The generation error is obtained by Proposition \ref{prop1} as in the case of the score matching loss. This completes the proof. 
\end{proof}

\begin{lemma}[Error of unconditional diffusion generation]
\label{cor-w}
Under the conditions in Theorems \ref{thm_ug} and \ref{cor-nt2}, the errors for the transfer and non-transfer models under the Wasserstein distance $\smc{W}$ are bounded by
$
\smc{W}(P^0_{\x_t},\hat P_{\x_t}) \leq c_8 \rho_t(\gamma_t^0,\hat{\gamma}_t)$, $\smc{W}(P^0_{\x_t},\tilde P_{\x_t})\leq c_8 \rho_t(\gamma_t^0,\tilde{\gamma}_t)$,
for some constant $c_8>0$, leading to the error bounds under the Wasserstein
distance $\smc{W}$ from Theorem \ref{thm_ug} and Theorem \ref{cor-nt2}. 
\end{lemma}

\begin{proof}[Lemma \ref{cor-w}]
By the triangle inequality,
\begin{equation*}
    \smc{W}(\hat{P}_{\x_t},P_{\x_t}^0)\leq \smc{W}(\hat{P}_u(\hat{g}_t^{-1}),P_u^0(\hat{g}_t^{-1}))+\smc{W}(P_u^0(\hat{g}_t^{-1}),P_u^0({(g_t^0)}^{-1})).
\end{equation*}

Note that when $g_t$ is bounded, $\hat{P}_{\x}$ and $P^0_{\x}$ are supported in a bounded domain with diameter $c_R$ which depends on $B_g$ and $\mathbb{B}_g$. Then $\smc{W}(\hat{P}_u(\hat{g}_t^{-1}),P_u^0(\hat{g}_t^{-1}))$ can be bounded by the TV distance with $c_{TV}$ given in Lemma \ref{l-KL-Score},
\begin{equation*}
\smc{W}(\hat{P}_u(\hat{g}_t^{-1}),P_u^0(\hat{g}_t^{-1}))\leq c_R\mathrm{TV}(\hat{P}_u(\hat{g}_t^{-1}),P_u^0(\hat{g}_t^{-1}))\leq c_Rc_{TV}\rho_{U}(\theta_u^0,\hat{\theta}_u).
\end{equation*}
By the distance definition $
\smc{W}(\hat{P}_{\x_t},P^0_{\x_t})=\sup_{\|f\|_{Lip}\leq 1} \left|\int f(\x)\mathrm{d}\hat{P}_X-\int f(\x)\mathrm{d}P^0_X\right|
$,
\begin{equation*}
\smc{W}(P_u^0(\hat{g}_t^{-1},P_u^0(({g}_t^{0})^{-1})=\sup_{\|f\|_{Lip}\leq 1} |\mathrm{E}_u{f\circ \hat g_t}-\mathrm{E}_u{f\circ g_t^0}|\leq \mathrm{E}_u\|\hat{g}_t-g_t^0\|\leq \rho_g(g^0_t,\hat{g}_t).
\end{equation*}
Combining the above inequalities leads to the final result. This completes
the proof.
\end{proof}

Next, we will derive an explicit bound for $\rho(\theta^0_u,\hat{\theta}_u)$, 
which yields the generation error rate discussed in Section \ref{sec_diffusion}.

\begin{assumption}[Target density for $\U$]
\label{A_pu} Suppose the latent density of $\U$, denoted $p^0_{\bm u}$, can be expressed as $p_{\bm u}(\bm u) = \exp(-c_9\|\bm u\|^2 / 2) \cdot f_u(\bm u)$, where $c_9>0$ is a constant. Furthermore, $f_u$ is contained in a H\"older ball $\smc{C}^{r_u}(\R^{d_u},\R, B_u)$ of radius $B_u$ and is bounded below by a positive constant. 
\end{assumption}

\begin{lemma}[Latent generation error]
\label{thm_pu}
Under Assumption \ref{A_pu}, if we choose the structure hyperparameters of any neural network in $\Theta_u$ to be $\mathbb{L}_u=c_L \log^4K$, $\mathbb{W}_u= c_W K\log^7K$, $\mathbb{S}_u= c_S K\log^9K$, 
$\log\mathbb{B}_u= c_B\log K$, $\log\mathbb{E}_u= c_E\log^4K$,  with diffusion stopping criteria from \eqref{forward}-\eqref{reverse} 
as $\log \underline{\tau}_u= -c_{\underline{\tau}}\log K$ and $\overline{\tau}_u=c_{\overline{\tau}}\log K$, where $\{c_L,c_W, c_S, c_B,c_E,c_{\underline{\tau}},c_{\overline{\tau}}\}$ are sufficiently large constants,
then the excess risk is bounded by: for any $x\geq 1$, 
\begin{eqnarray*}
P(\rho(\theta^0_u,\hat{\theta}_u)\geq x\varepsilon^u_s)\leq \exp(-c_3 n_s^{1-\xi}(x\varepsilon^u_s)^2), 
\end{eqnarray*}
with some constant $c_3>0$  and a small $\xi>0$ same in Assumption \ref{U-error}.
Here, $\varepsilon^u_s=\beta_u+\delta_u$ with the estimation error $\beta_u$ and the approximation error $\delta_u$ defined as
$\beta_u\asymp 
    \sqrt{\frac{K\log^{19} K}{n_s}},\quad \delta_u\asymp K^{-\frac{r_u}{d_{u}}}\log^{\frac{r_u}{2}+1}K$.
To obtain the optimal trade-off, we set $\beta_u=\delta_u$ to determine $K$,
after ignoring the logarithmic term, the optimal bound is obtained by $K\asymp n_s^{\frac{d_u}{d_u+2{r_u}}}$. This yields
$\varepsilon^u_s\asymp n_s^{-\frac{r_u}{d_u+r_u}} \log^{m_u} n_s$, where $m_u
=\max(\frac{19}{2}, \frac{r_u}{2}+1)$. Similarly, $\varepsilon^u_t\asymp n_{t}^{-\frac{{r_u}}{d_u+2 r_u}} \log^{m_u} n_t$.
\end{lemma}
\begin{proof}
    This lemma is a direct consequence of Theorem \ref{thm_diff_general}.
\end{proof}

\subsection{Proofs of Section 4}
\subsubsection{Generation accuracy of coupling normalizing flows}
Given a training sample set $(\x^i,\z^i)_{i=1}^{n}$, we define the loss function as $l=-\log p_{\x|\z}(\x,\z;\theta)$, and then estimate the flows by minimizing the negative log-likelihood as follows:
\begin{align}
\label{eq-likelihood-g}
    \hat{\theta}_{t}(\x,\z)&=\arg\min_{\theta\in \Theta}-\sum_{i=1}^{n} \log p_{\x|\z}(\x^i,\z^i;\theta) \nonumber\\
    &=\arg\min_{\theta\in\Theta}-\sum_{i=1}^{n}\left(\log p_{\bm v}(\theta(\x^i,\z^i))+\log \left|\det\left(\nabla_{\x}\theta(\x^i,\z^i) \right)\right|\right),
\end{align}
where $\Theta = \mathrm{CF}(\mathbb{L},\mathbb{W},\mathbb{S},\mathbb{B},\mathbb{E},\lambda)$.

Next, we specify some conditions for the true mapping $T^0$.
\begin{assumption}
\label{A-flow-general}
    There exists a map ${{T}}^0:[0,1]^{d_x}\times [0,1]^{d_z}\rightarrow [0,1]^{d_x} $ such that $\bm v={T}^0(\x,\z)$, where $\bm v$ a random vector with a known lower bounded smooth density, $p_{\bm v}\in \smc{C}^{\infty}([0,1]^{d_{x}},\R,B_v)$. For any $\z \in [0,1]^{d_z}$, ${T}^0(\cdot,\z)$ is invertible and $|\det(\nabla_{\x}T^0)|$ is lower bounded. Moreover, $T^0(\bm v,\z)$ and its inverse given $\z$ belong to H\"older ball $\smc{C}^{r+1}([0,1]^{d_{x}} \times [0,1]^{d_z},[0,1]^{d_{x}},B)$.
\end{assumption}

\begin{theorem}[Generation error of coupling flows]
\label{thm_cp_general}
Under the conditions in Assumption \ref{A-flow-general}, we set the neural network's structure hyperparameters within 
$\Theta$ with a set of sufficiently large positive constants $\{c_L,c_W, c_S, c_B,c_E,c_{\lambda}\}$ as follows: $\mathbb{L}=c_L L\log L$, $\mathbb{W}=c_W W\log W$, $\mathbb{S}=c_SW^2L\log^2W\log L$, $\mathbb{B}=c_B$, $\log\mathbb{E}=c_E\log(WL)$ and $\lambda=c_\lambda$. With these conditions, let $K=WL$, and then the error in coupling flow models via transfer learning
under the KL-divergence $\smc{K}$ is bounded: for any $x\geq 1$ and
some constant $c_e>0$,
\begin{eqnarray}
\label{c-rate-general-cp}
P(\mathrm{E}_{\z}[\smc{K}^{1/2}(p^0_{\x|\z},\hat{p}_{\x|\z})] \geq x(\beta_n + \delta_n)) \leq \exp(-c_e n (x(\beta_n + \delta_n))^2). 
\end{eqnarray} 
Here, $\beta_n$ and $\delta_n$ represent the estimation and approximation errors:
$\beta_n\asymp 
    \sqrt{\frac{K^2\log^5K}{n}}$ and $\delta_n\asymp K^{\frac{-2r}{d_x+d_z}}$.
In \eqref{c-rate-general-cp}, setting $\beta_n = \delta_n$ to solve for $K$, and neglecting the logarithmic term, leads to 
$\beta_n = \delta_n \asymp n^{-\frac{r}{d_x+d_z+2r}}$ with 
the optimal $K \asymp \sqrt{n^{\frac{d_x+d_z}{d_x+d_z+2r}}}$. Consequently, 
this provides the best bound for \eqref{c-rate-general-cp} $n^{-\frac{r}{d_x+d_z+2r}}\log^{5/2} n$.

Moreover, when $\Z= \emptyset$ and $d_z=0$, this error bound can be extended to the degenerate case, unconditional flow models,
$
\smc{K}^{1/2}(p^0_{\x},\hat{p}_{\x})=O_p(n^{-\frac{r}{d_x+2r}}\log^{5/2} n).
$
\end{theorem}

\begin{proof}[Theorem \ref{thm_cp_general}]
Let $\theta^0=T^0$. By Assumption \ref{A-flow-general}, $\theta^0$ has a lower bounded determinant of the 
Jacobian matrix, that is, $|\det(\nabla_{\x}\theta^0(\x,\z))|>\underline{c}_{\theta^0}>0$, and $p_v$ is lower bounded by some constant, $p_v\geq \underline{c}_{v}>0$. Moreover,
$|\det(\nabla_{\x}\theta(\x,\z))|\leq \Pi_{i=1}^{d_x}\|\nabla_{\x_i}\theta(\x,\z)\|\leq (\sqrt{d_x}B)^{d_x}$ by Hadamard’s inequality \cite{rozanski2017more}.

Note that $\Theta$ is an invertible class. Then,  $\inf_{\theta\in \Theta,\x,\z}|\det(\nabla_{\x}\theta(\x,\z))|>\underline{c}_{\theta}$ for some constant $\underline{c}_{\theta}>0$. Hence, the densities are lower bounded,
\begin{equation}
\label{eq-density-lb}
p^0_{\x|\z}(\x,\z)\geq  \underline{c}_v \underline{c}_{\theta^0},
\text{  and  }
 p_{\x|\z}(\x,\z)\geq  \underline{c}_v \underline{c}_{\theta},
\end{equation}
implying that  
\begin{align*}
    \frac{p_{\x|\z}}{p^0_{\x|\z}}=\frac{p_{\bm v}(\theta(\x,\z))|\det(\nabla_{\x}\theta(\x,\z))|}
    {p_{\bm v}(\theta^0(\x,\z))|\det(\nabla_{\x}\theta^0(\x,\z)))|}
\geq \underline{c}_r,
\end{align*}
for some constant $\underline{c}_r=\frac{\underline c_{v} \underline c_{\theta}}{B_v(\sqrt{d_x}B)^{d_x}}$. 
Meanwhile, $\sup_{\theta\in \Theta,\x,\z}|\det(\nabla_{\x}\theta(\x,\z))|\leq (\sqrt{d_x}\lambda)^{d_x}$ and 
$
     \frac{p_{\x|\z}}{p^0_{\x|\z}}\leq \frac{B_{v}(\sqrt{d_x}\lambda)^{d_x}}{\underline c_{v},\underline c_{\theta^0}}:=\overline{c}_r
$.
Then, $l(\x,\z;\theta)-l(\x,\z;\theta^0)$ is bounded, which satisfies Assumption \ref{sub-Gaussian} with $c_b=\max(|\log \overline{c}_r|,|\log \underline{c}_r|)$.

To verify the variance condition in Assumption \ref{Variance}, note that the likelihood ratio is bounded
above and below. By Lemmas 4 and 5 of \cite{wong1995probability}, the first and second moments of
the difference of the log-likelihood functions is bounded: 
\begin{equation}
\label{eq-kl-h}
 \mathrm{E}_{\x|\z}(l(\X,\Z;\theta^0) - l(\X,\Z;\theta))^j\leq c_l\|{p}_{\x|\z}^{1/2}- (p_{\x|\z}^0)^{1/2}\|_{L_2}^2   
\end{equation}
for $j=1,2$ and some constant $c_l>0$ depend on $\underline c_r$. This implies that
 \begin{align*}
 \mathrm{Var}_{\x,\z}(l(\X,\Z;\theta^0) - l(\X,\Z;\theta))
&\leq \mathrm{E}_{\x,\z}[(l(\X,\Z;\theta^0) - l(\X,\Z;\theta))^2] \\
&\leq c_{l}  \mathrm{E}_{\z}\|{p}_{\x|\z}^{1/2}- (p_{\x|\z}^0)^{1/2}\|_{L_2}^2\leq c_l\rho^2(\theta^0,\theta).
\end{align*}
Hence, Assumption \ref{Variance} holds with $c_v=c_l$.

 Thus, we can apply Proposition \ref{prop1} together with Lemmas \ref{l-approx-cp} and \ref{l-entropy-cp} to give the desired result in the same manner as the proof of Theorem \ref{thm_diff}. 
By Lemma \ref{l-entropy-cp}, there exists a positive $c_H$ such that $H_B(u,\smc{F})\leq c_H K^2\log^{4}K\log\frac{K}{u}$ . Then
the integral entropy inequality can be solved by 
\begin{align*}
 \int_{k\varepsilon^{2}/16}^{4 c^{1/2}_v \varepsilon}
H_B^{1/2}(u,\smc{F}) \mathrm{d}u &\leq c_H\int_{k\varepsilon^{2}/16}^{4c^{1/2}_{v} \varepsilon} 
K\log^{4}K\log\frac{K}{u}\mathrm{d}u\leq 4c_H c^{1/2}_v\varepsilon \sqrt{ K^2\log^{4}K\log\frac{K}{\varepsilon^2}}.
\end{align*}
Solving $\beta_n$: $4c_H c^{1/2}_v \beta_n \sqrt{ K^2\log^{4}K\log\frac{K}{\varepsilon^2}}\leq c_h\sqrt{n}\beta_n^2$ yields $\beta_n=2\frac{4c_H\sqrt{c_v}}{c_h}\sqrt{\frac{K^2\log^{5} K}{n}}$.

Let $\varepsilon_n\geq \beta_n+\delta_n$ with $\beta_n\asymp\sqrt{\frac{K^2\log^{3} K}{n}}$ so that $\varepsilon_n$ satisfies the conditions of Proposition \ref{prop1}. Note that $c_e\asymp\frac{1}{c^2_v}\asymp 1$ in Proposition \ref{prop1}.
 With $\log K=O(\log n)$, for some constant $c_e>0$,
$
P(\rho(\theta^0,\theta) \geq \varepsilon_n )\leq 4\exp(-c_e n\varepsilon_n^2 )
$.
 \end{proof}

 The next lemma gives the approximation error for the coupling flows in the KL divergence. The coupling network has been shown to possess the universal approximation property \citep{ishikawa2023universal}. But the approximation rate result is largely missing, except \cite{jin2024approximation} provided an approximation error bound for a bi-Lipschitz $T^0$, which cannot be used for a smooth $T^0$ to present the density-based metric result.
Our approximation employs the zero-padding method and the ReQU network to give the explicit rate. The zero-padding technique is used in \cite{lyu2022cflows} to show that coupling flows are also universal approximators for the derivatives, and the ReQU network is used in \cite{belomestny2021rates} to give the approximation rate for the derivatives of smooth functions.
\begin{lemma}[Approximation error]\label{l-approx-cp}
    Under Assumption \ref{A-flow-general}, there exists a coupling network $\pi\theta^0 \in \mathrm{CF}(\mathbb{L},\mathbb{W},\mathbb{S},\mathbb{B},\mathbb{E})$ with $\mathbb{L}=c_L L\log L$, $\mathbb{W}=c_W W\log W$, $\mathbb{S}=c_SW^2L\log^2W\log L$, $\mathbb{B}=c_B$, $\log\mathbb{E}=c_E\log(WL)$, and $\lambda=c_\lambda$, such that
\begin{equation*}
 \rho(\theta^0,\pi\theta^0)=O( (WL)^{\frac{-2r}{d_x+d_z}}).   
\end{equation*}
\end{lemma}
\begin{proof}[Lemma \ref{l-approx-cp}]
    Consider the zero padding method to derive the approximation result for coupling flows. 
   Deep affine coupling networks are shown to be universal approximators in the Wasserstein distance if we
allow training data to be padded with sufficiently many
zeros \citep{koehler2021representational, huang2020augmented}. 
We first present how the zero padding method works in Wasserstein distance. The proof process is to construct a coupling flow $(\X,\bm 0) \mapsto ({{T}}(\X), \bm 0)$ as follows:
\begin{equation*}
\X\xrightarrow{[\bm I;\bm 0]}\underbrace{(\X,\bm 0)}_{\Y^1}\xrightarrow{\phi_1} \underbrace{(\X, {{T}}(\X))}_{\Y^2} \xrightarrow{\phi_2}  \underbrace{({{T}}(\X), {{T}}(\X))}_{\Y^3} \xrightarrow{\phi_3} ({{T}}(\X), \bm 0)\xrightarrow{[\bm I,\bm 0]}T(\X),
\end{equation*}
where $\bm 0$ is the same dimension of $\X$.
To achieve this, three coupling layers are used,
 \begin{align*}
&[\phi_1(\Y^1)]_j = 
\begin{cases} 
\Y^1_j & j=1,\ldots,d \\
 \Y^1_j+[T(\Y^1_{1:d})]_j & j=d+1,\ldots,2d
\end{cases}
\\
&[\phi_2(\Y^2)]_j = 
\begin{cases} 
\Y^2_j-([T^{-1}(\Y^2_{(d+1):2d})]_j+\Y^2_{j+d}) & j=1,\ldots,d \\
 \Y^2_j & j=d+1,\ldots,2d
\end{cases}
\\
&[\phi_3(\Y^3)]_j = 
\begin{cases} 
\Y^3_j & j=1,\ldots,d \\
 \Y^3_j-\Y^3_{j-d} & j=d+1,\ldots,2d.
\end{cases}
\end{align*}

Then, if we have common networks that approximate ${{T}}$ and ${{T}}^{-1}$  well in $\phi_1$ and $\phi_2$, then we can control the approximation error for the coupling network.
 However, it is not sufficient to control the KL divergence by the approximation error in ${T}$, because this mapping is volume-preserving with the Jacobian determinant equal to one.  The existing literature on affine
coupling-based normalizing flows considers weak convergence \cite{teshima2020coupling} in the Wasserstein distance. The
approximability to derivatives remains hardly untouched, except \cite{lyu2022cflows} which takes into account the approximation of derivatives but fails to give an explicit approximation error rate.  

Next, we will adjust the zero padding method to approximate ${T}$ and $|J_{T}|$ simultaneously. The proof process is to construct a coupling flow $(\X) \mapsto ({T}(\X,\Z))$ as follows:

\begin{equation*}
\X\xrightarrow{[\bm I;\bm 0]}\underbrace{(\X,\bm 0)}_{\Y^1}\xrightarrow{\phi_1} \underbrace{(\X, {{T}}(\X,\Z))}_{\Y^2} \xrightarrow{\phi_2}  \underbrace{({{T}}(\X,\Z), \X)}_{\Y^3} \xrightarrow{\phi_3} ({{T}}(\X,\Z), \bm 0)\xrightarrow{[\bm I,\bm 0]}T(\X,\Z).
\end{equation*}
where $\bm 0$ is in the same dimension as $\X$.
To achieve this, two coupling layers and a permutation layer are used,

\begin{align}
[\phi_1(\Y_1)]_j &=
\begin{cases}
\Y_{1,j} & \text{for } j=1,\ldots,d_x, \\
\Y_{1,j} + [T(\Y_1^{1:d_x})]_j & \text{for } j=d_x+1,\ldots,2d_x;
\end{cases}
\nonumber \\
[\phi_2(\Y_2)]_j &=
\begin{cases}
\Y_{2,j+d_x} & \text{for } j=1,\ldots,d_x, \\
\Y_{2,j-d_x} & \text{for } j=d_x+1,\ldots,2d_x;
\end{cases}
\nonumber \\
[\phi_3(Y_3)]_j &= 
\begin{cases} 
\Y_{3,j} & \text{for } j=1,\ldots,d_x, \\ 
\Y_{3,j} - [T^{-1}(\Y_3^{1:d_x})]_j & \text{for } j=d_x+1,\ldots,2d_x.
\end{cases}
\label{eq_p1}
\end{align}
In this process, the Jacobian of the composite transformations remains $\nabla_{\x}T$,
\renewcommand{\arraystretch}{0.8}
\begin{align*}
[\bm I,\bm 0]
   \nabla_{\x}\phi_3\nabla_{\x}{\phi_2} \nabla_{\x}{\phi_1}\left[\begin{array}{cc}
         \bm I \\
      \bm 0
   \end{array}\right]
=   [\bm I,\bm 0] 
   \left[\begin{array}{cc}
         \bm I &0\\
       -\nabla_{\x}T^{-1}&\bm I
   \end{array}\right]
   \left[\begin{array}{cc}
      \bm 0   &\bm I \\
      \bm I &\bm 0
   \end{array}\right]
   \left[\begin{array}{cc}
         \bm I &\bm 0\\
       \nabla_{\x}T & \bm I
   \end{array}\right]
   \left[\begin{array}{cc}
         \bm I \\
      \bm 0
   \end{array}\right]=\nabla_{\x}T. 
\end{align*}
Here, the subscript $j$ refers to the $j$-th component of a vector, and $\Y_i^{1:d_x}$ refers to the first $d_x$ components of $Y_i$. The coupling layer; $j=1,3$, is defined by $\phi_j(\x_1,\x_2)=(\x_1,\x_2+\omega_j(\x_1))$ and $\omega_j(\x_2)$. To ensure invertibility, $\omega_1$ is required to be invertible and $\omega_3=-\omega_1^{-1}$, where $\omega_1$ approximates $T$ using a mixed ReLU and ReQU neural network.  Here, ReQU stands for rectified quadratic unit, with $\sigma(x)=\max^2(0,x)$ as an activation function, which permits an approximate of a smooth function and its derivative simultaneously \citep{belomestny2023simultaneous}.
Meanwhile, the second layer employs a permutation, with $\phi_2$ as a function to permute the first block with the second.

Using Lemma \ref{l_inn}, we construct an invertible $\Phi$ to approximate $T^0$ while obtaining an analytical solution $\Phi^{-1}$ to inverse $\Phi$ under some constraints on $\Phi$. By Lemma \ref{l_inn}, the network $\Phi$ has depth $O(L\log L)$ and width $O(W\log W)$ such that the approximation error is bounded by
\begin{align*}
    \|\Phi-T^0\|_{\infty,2}=O((WL)^{-\frac{2r}{d_x+d_z}})\text{ and }
    \|\nabla_x\Phi-\nabla_{\x} T^0\|_{\infty,2}=O((WL)^{-\frac{2r}{d_x+d_z}}).
\end{align*}
Here, for a $d$ dimension output function $f$, $\|f\|_{\infty,2}=\|\|f\|_{\infty}\|_{L_2}=(\int_{\x}(\max_j|f_j(\x)|)^2\mathrm{d}\x)^{1/2}$, where $f_j$ is the $j$-th element of $f$ and $\|g\|_{L_2}=\sqrt{\int g^2(\x)\mathrm{d}\x}$ is the $L_2$ norm for the univariate output function.
For the coupling flow $\pi\theta^0$, the approximation error is bounded by
\begin{align*}
    \|\pi\theta^0-\theta^0\|_{\infty,2}=\|\Phi-T^0\|_{\infty,2}=O( (WL)^{-\frac{2r}{d_x+d_z}}).
\end{align*}
Meanwhile, by the perturbation bound of the determinant in Lemma \ref{l_det}, we bound the determinant error with some positive constant $c_{d}$,
\begin{align}
\label{eq-det-inf}
    \||\det(\nabla_{\x} \pi\theta^0)|-|\det(\nabla_{\x} \theta^0)|\|_{L_2}
    \leq c_d\|\nabla_{\x} \Phi-\nabla_{\x} T^0\|_{\infty,2}
    = O((WL)^{-\frac{2r}{d_x+d_z}}).
\end{align}

On the other hand, let $p_{\x|\z}$  be the density given by $\pi\theta^0$. 
Note that the likelihood ratio is lower and upper bounded. Then, the KL divergence is upper bounded by
the squared $L_2$ distance from \eqref{eq-density-lb} and  \eqref{eq-kl-h},
\begin{align*}
    \rho^2(\theta^0,\pi\theta^0)=\smc{K}(p^{0}_{\x|\z},p_{\x|\z}) \leq c_l\int (p^{1/2}_{\x|\z}-(p^{0}_{\x|\z})^{1/2})^2
\leq 2c_l(\underline{c}_v\underline{c}_{\theta}+\underline{c}_v\underline{c}_{\theta^0})\|p_{\x|\z}-p^{0}_{\x|\z}\|^2_{L_2}.
\end{align*}
Moreover, by the transformation, we have the decomposition,
\begin{align}
\label{eq-like2L2}
    &\|p_{\x|\z}-p^0_{\x|\z}\|_{L_2}=\|p_{\bm v}(\theta^0)|\det(\nabla_{\x}\theta^0)|-p_{\bm v}(\pi\theta^0)|\det(\nabla_{\x} \pi\theta^0)|\|_{L_2} \nonumber\\
    \leq& \|p_{\bm v}(\theta^0)|\det(\nabla_{\x}\theta^0)|-p_{\bm v}(\pi\theta^0)|\det(\nabla_{\x} \theta^0)|\|_{L_2}\nonumber\\
    &+\|p_{\bm v}(\pi\theta^0)|\det(\nabla_{\x}\theta^0)|-p_{\bm v}(\pi\theta^0)|\det(\nabla_{\x} \pi\theta^0)|\|_{L_2}\nonumber\\
    \leq& (\sqrt{d_x}B)^{d_x}\sqrt{d_x}B_v\|\theta^0-\pi\theta^0\|_{\infty,2}+B_v\||\det(\nabla_{\x} \theta^0)|-|\det(\nabla_{\x} \pi\theta^0)|\|_{L_2}. 
\end{align}
Hence,
\begin{align*}
    K^{1/2}(p_{\x|\z}^0,p_{\x|\z})\asymp \|T^0-\Phi\|_{\infty,2}+\||\det(\nabla_{\x} T^0)|)-|\det(\nabla_{\x} \Phi)|\|_{L_2}=O((WL)^{-\frac{2r}{d_x+d_z}}).
\end{align*}
This completes the proof.
\end{proof}

 The next lemma will show that there exists an invertible neural network $\Phi$ to approximate $T^0$ and $\nabla_{\x} T^0$ simultaneously with a combination use of the ReLU network and ReQU network.
\begin{lemma}
\label{l_inn}
There exists an invertible neural network $\Phi$ with $\mathbb{W}=c_W W\log W$, $\mathbb{L}=c_L L\log L$,  $\mathbb{S}=c_S \mathbb{W}^2\mathbb{L}$, $\mathbb{B}=c_B$, $\log\mathbb{E}=c_E\log(WL)$ and $\lambda=c_{\lambda}$ such that
   \begin{align*}
    \|\Phi-T^0\|_{\infty,2}=O( (WL)^{-\frac{2r}{d_x+d_z}}), \quad 
    \|\nabla_x\Phi-\nabla_xT^0\|_{\infty,2}=O( (WL)^{-\frac{2r}{d_x+d_z}}).
\end{align*}
\end{lemma}
\begin{proof}
Before proceeding, we describe the idea of building a neural network $\Phi$ in
two steps: (1) approximating a smooth map $T^0$ with a local polynomial $\bar T$, as 
in \cite{lu2021deep}; (2) constructing an invertible neural network $\Phi$ to approximate $\bar T$, where $\Phi$ is subject to some constraints. 

Let $\s=(\x,\z)$ and $K=[(WL)^{2/d}]$. We uniformly partition a box area into non-overlapping hypercubes $\{\smc{B}^i\}_{i=1}^{K^d}$ with edge sizes $\frac{1}{K}$. 
Note that $T^0\in \smc{C}^{1+r}$. 
For any $\s\in \smc{B}^i=\{\s:\s_j^i\leq\s_j\leq \s_j^i+\frac{1}{K}
\}$ with $\s^i_j=\frac{\bm K_j}{K}$ and $\bm k \in [K-1]^d$, there 
exists $ \xi_{\s} \in (0,1) $ such that
\begin{align*}
    T^0(\s)&=\sum_{|\bm{\alpha}|\leq \lfloor r\rfloor}\frac{\partial ^{\bm \alpha}T^0(\s^i)}{\bm \alpha ! \partial \s^{\bm \alpha}}(\s-\s^i)^{\bm \alpha}
   + \sum_{|\bm{\alpha}|=\lfloor r\rfloor+1}\frac{\partial ^{\bm \alpha}T^0(\s^i+\xi_{\s}(\s-\s^i))}{\bm \alpha ! }(\s-\s^i)^{\bm \alpha}\\
   &= \sum_{|\bm{\alpha}|\leq \lfloor r\rfloor+1}\frac{\partial ^{\bm \alpha}T^0(\s^i)}{\bm \alpha ! }(\s-\s^i)^{\bm \alpha}+ O(\|\s-\s^i\|^{r+1})\equiv \bar{T}(\s)+ O(\|\s-\s^i\|^{r+1}),
\end{align*}
where $\alpha$ is a multi-index with $|\cdot|$ indicating its size, and
$\sup_{\s}\|\bar{T}(\s)-T^0(\s)\|_{\infty}=O(K^{-(r+1)})$ and $\sup_{\s}\|\nabla_{\x}\bar{T}(\s)-\nabla_{\x} T^0(\s)\|_{\infty}=O(K^{-r})$.
Next, we construct an invertible network to approximate $\bar{T}$ with an error bounded by $O(K^{-r})$ in the $L_2$ distance.

To achieve this, we first adjust $\smc{B}^i$ to a new cube $\{\smc{B}^i(\epsilon)\}$with radius $\epsilon>0$, where $[\smc{B}^i(\epsilon)]_j=[\s^i_j,\s^i_j+\frac{1}{K}-\epsilon]$ for a small $\epsilon>0$. We approximate $\bar T$ in $\{\smc{B}^i(\epsilon)\}$ with its error
controlled by choosing a small $\epsilon$.

The neural network $\Phi$ is constructed in three steps.
\begin{enumerate}
    \item The first step constructs a ReLU network $\Phi_a$ to yield step functions over
$\smc{B}^i(\epsilon)$ such that $\phi_a(\s)=\s^i$ if $\s^i\in \smc{B}^i(\epsilon)$
for $\phi_a \in \Phi_a$. This reduces the function approximation problem to a point-fitting problem at fixed grid points.
    \item The second step constructs a group of ReLU networks $\Phi_b=\{\phi_b^{\bm \alpha}\}$ such that $\phi^{\bm \alpha}_{b}(\s^i)$ is close to $\partial^{\bm \alpha}T^0(\s^i)$ for $|\bm \alpha|\leq \lfloor r\rfloor+1$.
    \item The last step constructs a ReQU network $\Phi_c$ such that $\phi_c(\s) \in \Phi_c$ to yield a polynomial in that $\phi_c(\s-\s^i,\phi^{\bm \alpha}_{b}(\s^i))=\sum_{|\bm\alpha|\leq \lfloor r\rfloor+1}\frac{\phi^{\bm \alpha}_{b}(\s^i)}{\bm\alpha!}(\s-\s^i)^{\bm{\alpha}}$ .
\end{enumerate}

Combining $\Phi_a$--$\Phi_c$, we define an element $\phi(\s)$ in the 
complete network $\Phi$ as 
\begin{equation}
\label{eq-flow-phi}
\phi(\s)=\phi_c(\s-\phi_a(\s),\phi^{\bm \alpha}_{b}(\phi_a(\s)))=\sum_{|\bm\alpha|\leq \lfloor r\rfloor+1}\frac{\phi^{\bm \alpha}_{b}(\s^i)}{\bm\alpha!}(\s-\s^i)^{\bm{\alpha}}.    
\end{equation}

Let $\bm \alpha(\beta,j)=[\bm \beta_1,\ldots,\bm \beta_j+1,\ldots,\bm \beta_{d_s}]$ for $j\in [d_s]$ and $\bm \beta\in [r]^{d_s}$. By the chain rule of derivatives and the fact 
$\nabla \phi_a=\bm 0$, we have, for $s\in \smc{B}^k(\epsilon)$,
\begin{equation}
[\nabla\phi(\s)]_{ij}=\sum_{|\bm\beta|\leq \lfloor r\rfloor}\frac{[\phi^{\bm\alpha(\bm\beta,j)}_{b}(\s^k)]_i}{\bm\beta!}(\s-\s^k)^{\bm\beta}.
\end{equation}
Then, $\nabla \phi=[\nabla_{\x}\phi;\nabla_{\z}\phi]$ and $\nabla_{\x}\phi$ is the first $d_x$ rows of $\nabla\phi$, a square matrix of dimensions $d_x$. 

Specifically,  

\noindent\textbf{1. Approximating the step function. } 
For $\Phi_a$, we use Proposition 4.3 in \cite{lu2021deep}, presented in Lemma \ref{l_step} to construct a ReLU NN with depth $O(L)$ and width $O(W)$ for each dimension to yield the step function. Then $\phi_a$ is constructed with a ReLU network with depth $O(L)$ and width $O(W)$ such that $\phi_a(\s)=\s^i$, if $\s\in \smc{B}^i(\epsilon)$.

\noindent\textbf{2. Point fitting. } 
As to $\Phi_b$, we use Lemma \ref{l_pointfit}, the point-fitting result from \cite{lu2021deep} to construct $\phi^{\bm \alpha}_b=\psi^2 \circ \psi^1$. The construction involves two steps. First, we construct $\psi^1$ bijectively mapping $\{0,1,\ldots,K-1\}^d$ to $\{1,2,\ldots,K^d\}$, where $\psi^1(\bm k/K)=\sum_{j=1}^d\bm k_j K^{j-1}$. Then we construct $[\psi^2(i)]_j$ to approximate $[\partial^{\bm \alpha} T^0(\s^i)]_j$ with the pointwise error of $O((WL)^{-2\alpha})$, where $\psi^2(i)$ is a ReLU NN with depth $O(\alpha L\log L)$ and width $O(d_sW\log W)$. We choose $\alpha$ large enough so that $\max_i\|\psi^{\bm \alpha}_b(s^i)- \partial^{\bm \alpha}(\s^i)\|_{\infty}=\max_i\|\psi^2(i)- \partial^{\bm \alpha}(\s^i)\|_{\infty}=o(\frac{1}{K^{r+1}})$. 

\noindent\textbf{3. Approximating a polynomial. }
For $\Phi_c$, we use the ReQU network to implement the product exactly according to Lemma \ref{ReQU}. 
Let $\phi_c(\s-\s^i,\phi_b^{\bm \alpha}(s^i))=\sum_{\bm \alpha} \phi_c^{\bm \alpha}(\s-\s^i,\phi_b^{\bm \alpha}(s^i))$. Then, each $\phi_c^{\bm \alpha}(\s-\s^i,\phi_b^{\bm \alpha}(s^i))$ is designed as a depth $[\log_2(|\bm \alpha|+1)]$ and width $2^{[\log_2(|\bm \alpha|+1)]+1}$ ReQU network such that $\phi_c^{\bm \alpha}(\s-\s^i,\phi_b^{\bm \alpha}(s^i))=\phi_b^{\bm \alpha}(s^i)(\s-\s^i)^{\alpha}$. Then, $\Phi_c$ is constructed as $\mathrm{NN}([\log_2(\|\alpha\|_1+1)], r^{d_s}2^{[\log_2(\|\alpha\|_1+1)]+1})$.

Combining the networks in the three steps, $\Phi$ is a network with depth $\mathbb{L}\asymp L\log L$ and width $\mathbb{W}\asymp W\log W$. The effective neuron number is not greater than $\mathbb{W}^2\mathbb{L}$. Moreover, for any $\s\in \cup \smc{B}^i(\epsilon)$,
\begin{equation*}
\|\phi(\s)-\bar T(\s)\|_{\infty}\leq \sum_{|\bm \alpha|\leq \lfloor r\rfloor+1}\|\phi_b^{\bm \alpha}(\s^i)-\partial^{\bm \alpha}T^0(\s^i)\|=o(\frac{1}{K^{r+1}}).
\end{equation*}
Similarly, 
\begin{equation*}
\|\nabla_{\x}\phi(\s)-\nabla_{\x}\bar T(\s)\|_{\infty}\leq \max_{j\in[d_s]}\sum_{|\bm \beta|\leq \lfloor r\rfloor+1}\|\phi_b^{\bm \alpha(\beta,j)}(\s^i)-\partial^{\bm \alpha(\beta,j)}T^0(\s^i)\|_{\infty}=o(\frac{1}{K^{r+1}}).
\end{equation*}
Combining the approximation error of $\bar T$, we can show, for any $\s\in \cup \smc{B}^i(\epsilon)$,
\begin{equation*}
\|\phi(\s)- T^0(\s)\|_{\infty}=O(\frac{1}{K^{r+1}}) \text{  and }
\|\nabla\phi(\s)- \nabla T^0(\s)\|_{\infty}=O(\frac{1}{K^{r}}).
\end{equation*}

\textbf{Invertibility constraints. } We next outline the constraints necessary to guarantee the invertibility of $\phi$. It is important to note that $\phi$ is defined as a piecewise polynomial function. To ensure its invertibility, two specific conditions must be met. We first require that
$\phi$ is invertible in each cube and the image areas $\{\smc{Q}^i=\{\phi(\s), \s\in \smc{B}^i(\epsilon)\}\}_{i=1}^{K^d}$ are disjoint. Specifically,
\begin{align}
\label{eq-general-con}
    \left\{{\begin{array}{cc}     \displaystyle \inf_{\s\in\smc{B}^i(\epsilon),\s'\in\smc{B}^j(\epsilon)} \|\phi(s)-\phi(s')\|>c^{(1)}_{ij}&  i<j\in [K^d],\\
\inf_{\s\in\smc{B}^i(\epsilon)} |\det(\nabla_{\x}\phi(s))|>c^{(2)}_{i}&  i\in [K^d].
    \end{array}} \right.
\end{align}
Here, $c^{(1)}_{ij}=O(\epsilon)$ is set to no more than $\frac{1}{\sqrt{d}B}\epsilon$ due to the fact that $\|T^0(\s)-T^0(\s')\|\geq \frac{1}{\sqrt{d}B}\|\s-\s'\|\geq \frac{1}{\sqrt{d}B}\epsilon$, and $c^{(2)}_{i}=O(1)$ is set to no more than $\underline c_{\theta^0}=\inf_{\s} |\det(\nabla_{\x}T^0(\s))|$. 

When $r\leq 1$, the constraints can be simplified. Note that when $r\leq 1$, for $\s\in \smc{B}^i(\epsilon)$,  $\phi(s)=\phi^{\bm 0}_b(\s^i)+\phi_b(\s^i)(\s-\s^i)$ is a piece-wise linear map, where $\phi_b(\s^i)=[\phi^{\bm \alpha}_b(s^i)]_{|\bm \alpha|=1}$ is the Jacobian matrix with multi-index $\bm \alpha$ satisfying $|\bm \alpha|=1$. Then $|\det(\nabla_{\x}\phi(s))|=|\det([\phi_b(s^i)]_{\x})|$, where $[\phi_b(s^i)]_{\x}$ represents the rows associated with $\x$.  The constraints are simplified as 
\begin{align}
\label{eq-linear-con}
    \left\{{\begin{array}{ll}   
    \displaystyle \inf_{\substack{\s\in\smc{B}^i(\epsilon) \\ \s'\in\smc{B}^j(\epsilon)}}
    \|\phi^{\bm{0}}_b(\s^i) + \phi_b(\s^i)(\s-\s^i) - \phi^{\bm{0}}_b(\s^j) - \phi_b(\s^j)(\s'-\s^j)\| > c^{(1)}_{ij},   i<j\in [K^d], \\
    |\det([\phi_b(s^i)]_{\x})| > c^{(2)}_i,  i\in [K^d].
    \end{array}} \right.
\end{align}

When these two conditions hold, given $\z$ and $\Phi(\x,\z)=\y$, the inverse is constructed analytically when $r\leq 1$. Note that
$\Phi(\x,\z)=[\phi_{b}(\s^i)]_{\x}(\x-\x^i)+[\phi_{b}(\s^i)]_{\z}(\z-\z^i)+\phi^{\bm 0}_b(\s^i)$.
Given $\z$, if $\y\in \smc{Q}^i$, $\x$ can be solved by
$\x=[\phi_{b}(\s^i)]_{\x}^{-1}(\y-\phi^{\bm 0}_b(\s^i)-[\phi_{b}(\s^i)]_{\z}(\z-\z^i))$.  When $r>1$, we solve the $\lfloor r\rfloor+1$-th order polynomial on $\smc{B}^i(\epsilon)$ numerically, as in \cite{lyu2022cflows}.

Then, $\Phi$ defined in \eqref{eq-flow-phi}  satisfies the two conditions. Due to $\|\s-\s'\|\geq \sqrt{d_s}\epsilon$ for $\s\in \smc{B}^i(\epsilon)$ and $\s'\in \smc{B}^j(\epsilon)$ and the smooth property of $T^{-1}$, we have, for $y\in \smc{Q}_i$ and $y'\in \smc{Q}_j$,
\begin{align}
 \|\y-\y'\|_{\infty}
 &\geq \|T^0(\s)-T^0(\s')\|_{\infty}-\|\y-T^0(\s)\|_{\infty}-\|y'-T^0(s')\|_{\infty}\nonumber\\
 &\geq 
\frac{1}{\sqrt{d}B}\|\s-\s'\|-2B\sqrt{d}\frac{1}{K^{r+1}}
\geq \frac{1}{B}\epsilon-2B\sqrt{d}\frac{1}{K^{r+1}}
\geq \frac{1}{K^{r+1}}.   
\end{align}
The last inequality holds when choosing $\epsilon=\frac{c_{\epsilon}}{K^{r+1}}$ with $\frac{c_{\epsilon}}{\sqrt{d}B}-2B\sqrt{d}=1$. Hence, $\{\smc{Q}_i\}$ are disjoint with distances no less than $\frac{1}{K^2}$. 

Let $\smc{B}(\epsilon)=\bigcup\smc{B}^i(\epsilon)$. The final approximation is constructed with an additional layer with the indicator function, 
\begin{equation*}
    \phi^*(\s)=\mathbb{I}_{\smc{B}(\epsilon)}(\s)\phi(\s)+(1-\mathbb{I}_{\smc{B}(\epsilon)}(\s))\phi'(\s).
\end{equation*}
Here, $\phi'(\s)$ can be set to any bounded and invertible function with bounded derivatives, and the image of $\phi'$ should be disjoint with any $\{\smc{Q}^i\}_{i=1}^{K^d}$. So, a simple choice for $\phi'$ is $\phi'(\s)=\s+B$, where $B$ is the upper bound for $\phi$ and $T^0$. Then $\sup_s\|\phi'(\s)\|_{\infty}\leq B+1$ and $\sup_s\|\nabla_{\x}\phi'(\s)\|_{\infty}=\|I\|_{\infty}= 1$.

Furthermore, the indicator function can be implemented by $\mathbb{I}_{\smc{B}(\epsilon)}(\s)=1-\mathbb{I}_{[\frac{1}{K},\infty)}(\max_j[\phi_a(\s+\epsilon)-\phi_a(\s-\epsilon)]_j)$.
This is derived from the fact that $\|\phi_a(\s+\epsilon)-\phi_a(\s-\epsilon)\|_{\infty}\leq \max(\|\phi_a(\s+\epsilon)-(j-1)/K\|_{\infty},\|j/K-\phi_a(\s-\epsilon)\|_{\infty})< 1/K$ when $\s$ is an interior point in $\smc{B}(\epsilon)$, $j=1,2,\ldots,K$.

Setting $\epsilon=O(\frac{1}{K^{r+1}})$, we bound the $L_2$ error $\|\phi^*-T^0\|_{\infty,2}$ and $\|\nabla_{\x} \phi^*-\nabla_{\x} T^0\|_{\infty,2}$,
\begin{align}
   \|(\nabla_{\x} \phi^*)-(\nabla_{\x} T^0)\|_{\infty,2}&\leq \sqrt{\int_{[0,1]^d/(\smc{B}(\epsilon))}\|\nabla_{\x}\phi'-\nabla_{\x} T^0\|_{\infty}^2}+ \sqrt{\int_{\smc{B}(\epsilon)} \|\nabla_{\x} \phi-\nabla_{\x} T^0\|_{\infty}^2}\nonumber\\
   &\leq Kd(1+B)\epsilon+B\sqrt{d}\frac{1}{K^r}\leq (dBc_{\epsilon}+B\sqrt{d})\frac{1}{K^r}.
\end{align}
and 
\begin{align}
   \|\phi^*-T^0\|_{\infty,2}&\leq \sqrt{\int_{[0,1]^d/(\smc{B}(\epsilon))}\|\phi'-T^0\|^2_{\infty}}+ \sqrt{\int_{\smc{B}(\epsilon)} \|\phi-T^0\|_{\infty}}\nonumber\\
&\leq 2KdB\epsilon+B\sqrt{d}\frac{1}{K^r}\leq (2+c_{\epsilon}dB)\frac{1}{K^r},
\end{align}
leading to the desired result. This completes the proof. 

\textbf{Implementation and optimization. }
Achieving an invertible estimation of $T^0$ requires minimizing the negative log-likelihood function subject to the invertibility constraints \eqref{eq-general-con} and \eqref{eq-linear-con}. This optimization involves bi-level optimization, wherein lower-level optimization concerning $\s$ is performed within each constraint, while the upper-level optimization is conducted simultaneously.
To simplify this bi-level optimization, one can reformulate it as single-level unconstrained optimization using regularization or the Karush-Kuhn-Tucker (KKT) condition, as discussed in \cite{sinha2017review}. This reformulated problem can then be solved efficiently using stochastic gradient descent.
\end{proof}

\begin{lemma}[Metric entropy]
\label{l-entropy-cp}
For the neural network class $\Theta=\mathrm{NN}(\mathbb{L},\mathbb{W},\mathbb{S},\mathbb{B},\mathbb{E})$ defined in Lemma \ref{l-approx-cp}, 
the metric entropy of $\smc{F}=\{l(\cdot;\theta)-l(\cdot;\pi \theta^0): \theta \in\Theta\}$ 
is bounded by
$H_B(u,\smc{F})= O(K^2\log^4 K\log K/u)$.
\end{lemma}
\begin{proof}[Lemma \ref{l-entropy-cp}]
By \eqref{eq-density-lb}, for any $\theta_1,\theta_2\in\Theta$, the likelihood ratio is bounded. So there exists a $c_r,c^a_r,c^b_r>0$ such that for any $\x,\z$,
\begin{align*}
  |l(\x,\z,\theta_1)&-l(\x,\z,\theta_2)|=\left|\log(\frac{p_{\x|\z}(\x,\z,\theta_2)}{p_{\x|\z}(\x,\z,\theta_1)}-1+1)\right|\leq c_r |p_{\x|\z}(\x,\z,\theta_2)-p_{\x|\z}(\x,\z,\theta_1)|\\
  &\leq c^a_r\|\theta_1(\x,\z)-\theta_2(\x,\z)\|_{\infty}+c^b_r||\det(\theta_2(\x,\z))|-|\det(\theta_1(\x,\z))||.
\end{align*}

In the detailed setting of $\Theta$ in \eqref{eq-flow-phi} in the proof of Lemma \ref{l-approx-cp},  we denote the neural network $\Phi$ in $\theta_1$ and $\theta_2$ as $\Phi(\x,\z;\theta_1)$ and $\Phi(\x,\z;\theta_2)$, respectively. Then 
$\|\theta_1(\x,\z)-\theta_2(\x,\z)\|_{\infty}\leq \|\Phi(\x,\z;\theta_1)-\Phi(\x,\z;\theta_2)\|_{\infty}
$ and $\big||\det(\theta_1(\x,\z))|-|\det(\theta_2(\x,\z))|\big|\leq \sum_{|\bm \alpha|\leq \lfloor r\rfloor}c_d\|\phi^{\bm \alpha}_b(\x,\z;\theta_1)-\phi^{\bm \alpha}_b(\x,\z;\theta_2)\|_{\infty}$.
Both $\Phi$ and $\phi_b$ belong to $\mathrm{NN}(\mathbb{L},\mathbb{W},\mathbb{S},\mathbb{B},\mathbb{E})$.
Hence, 
\begin{equation*}
    H_B(u,\smc{F})\leq H(\frac{1}{2c^a_r}\frac{u}{2},\mathrm{NN}(\mathbb{L},\mathbb{W},\mathbb{S},\mathbb{B},\mathbb{E}))+r^{d_x+d_z}H(\frac{1}{2c^b_rc_dr^{d_x+d_z}}\frac{u}{2},\mathrm{NN}(\mathbb{L},\mathbb{W},\mathbb{S},\mathbb{B},\mathbb{E})).
\end{equation*}

By Lemma 4.2 in \cite{oko2023diffusion}, $H(u,\mathrm{NN}(\mathbb{L},\mathbb{W},\mathbb{S},\mathbb{B},\mathbb{E}))=O(\mathbb{S}\mathbb{L}\log(\mathbb{L}\mathbb{E}\mathbb{W}/u))=O(W^2L^2\log^2 W\log^2 L\log(WL/u))=O(K^2\log^4 K\log K/u)$ with $K=WL$. 
Although $\Theta$ is constructed with ReLU and ReQU layers, there are only ReQU layers in fixed depth and width, so the entropy bound for the ReLU network still holds here when $K$ is large enough.
Therefore, $H(u,\smc{F})=O(K^2\log^4 K\log K/u)$. This completes the proof. 
\end{proof}

 Lemma \ref{l_det} provides the perturbation bound for matrix determinants for
the proof of Lemma \ref{l-entropy-cp}. 

\begin{lemma}[Theorem 2.12 of \cite{ipsen2008perturbation}]\label{l_det}
Let \( A \) and \( E \) be \( d \times d \)  matrices. Then
\[
|\det(A + E) - \det(A)| \leq 
d\|E\|_2\max(\|A\|_2,\|A+E\|_2)^{d-1},
\]
where $\|A\|_2$ is the spectral norm of the matrix $A$.
\end{lemma}

\subsubsection{Proofs of Subsection \ref{flows-sec2}}
Theorem \ref{thm_flow-detail} gives the non-asymptotic probability bound for the generation error in Theorem \ref{thm_flow}.
\begin{theorem} [Conditional flows via transfer learning]
\label{thm_flow-detail}
Under Assumptions \ref{A-independent}-\ref{A-error} and \ref{A_BL}, there exists a coupling network $\Theta_t$ in \eqref{l-theta-cp} with specific hyperparameters: $\mathbb{L}_t=c_L L\log L$, $\mathbb{W}_t= c_W W\log W$, $\mathbb{S}_t= c_S W^2 L\log^2W \log L$, 
$\mathbb{B}_t= c_B$, $\log\mathbb{E}_t= c_E\log(WL)$, and $\lambda=c_{\lambda}$, such that the error of conditional flow generation through transfer learning under
the KL divergence is  
\begin{equation*}
\mathrm{E}_{\z_t}[\smc{K}^{1/2}(p^0_{\x_t|\z_t},\hat p_{\x_t|\z_t})]\leq x(\varepsilon_t+\sqrt{3 c_1}\varepsilon_s),
\end{equation*}
with a probability exceeding $1-\exp(-c_{10} n_t (x\varepsilon_t)^2) - \exp(-c_2 n^{1-\xi}_s (x\varepsilon_s)^2)$ for any $x\geq 1$ and some constant $c_{10}>0$.
Here, $c_L,c_W, c_S, c_B,c_E,c_\lambda$ are sufficiently large constants, $WL\asymp n_t^{\frac{d_{x_t}+d_{h_t}}{2(d_{x_t}+d_{h_t}+2r_t)}}$ and
$\varepsilon_t\asymp n_t^{-\frac{r_t}{d_{x_t}+d_{h_t}+2r_t}}\log^{5/2}  n_t$.
\end{theorem}
\begin{proof}[Theorems \ref{thm_flow} and \ref{thm_flow-detail}]
As in the proof of Theorem \ref{thm_diff}, we prove Theorem \ref{thm_flow} by applying the approximation error bound from Lemma \ref{l-approx-cp} and the estimation error bound from Lemma \ref{l-entropy-cp} as outlined in Theorem \ref{theorem1}.
\end{proof}

Theorem \ref{cor-cp-nt-detail} gives the non-asymptotic probability bound for the generation error in Theorem \ref{cor-cp-nt}.
\begin{theorem}[Non-transfer conditional flows]
\label{cor-cp-nt-detail}
Under Assumption \ref{A_BL}, there exists a coupling network $\Theta_t$ of the same configurations as in Theorem \ref{thm_flow}, the generation error of the non-transfer conditional flow is
\begin{equation*}
\mathrm{E}_{\z_t}[\smc{K}^{1/2}(p^0_{\x_t|\z_t},\tilde p_{\x_t|\z_t})]\leq x(\varepsilon_t+ \varepsilon^h_t)
\end{equation*}
with a probability exceeding $1-\exp(-c_{10} n_t (x(\varepsilon_t+\varepsilon^h_t))^2)$ for any $x\geq 1$ and some constant $c_{10}>0$. Here, 
 $\varepsilon_t\asymp n_t^{-\frac{r_t}{d_{x_t}+d_{h_t}+2r_t}}\log^{5/2}n_t$.
\end{theorem}
\begin{proof}[Theorems \ref{cor-cp-nt} and \ref{cor-cp-nt-detail}]
Theorem \ref{cor-cp-nt} directly follows from the general result in Theorem \ref{thm_cp_general}, which provides the bound for $\varepsilon_t^h$ as established in Lemma \ref{l-composite-entroty}.
\end{proof}

We set $\Theta_s=\mathrm{CF}(\mathbb{L}_s,\mathbb{W}_s,\mathbb{S}_s,\mathbb{B}_s,\mathbb{E}_s,\lambda_s)$ and estimate the mapping $\hat{T}_s$ by minimizing the negative 
log-likelihood,
\begin{align}
\label{eq-likelihood-source}
    \hat{{T}}_s=\hat{\theta}_{s}(\x_s,\hat{h}(\z))&=\arg\min_{\theta_s\in \Theta_t,h\in\Theta_h}\sum_{i=1}^{n_s} -\log p_{\x_s|\z_s}(\x_s^i,\z_s^i;\theta_t;h) \nonumber\\
    &=\arg\min_{\theta_s\in\Theta_s,h\in\Theta_h}-\log p_{\bm v}(\theta_s(\x_s^i,{h}(\z_s^i)))-\log \left|\det( \nabla_{\x}\theta_s(\x_s^i,h(\z_s^i)))\right|,
\end{align}

 Next, we specify some assumptions on the true $T^0_s$.

\begin{assumption}\label{A_BL_source}
  There exists a map $T_s^0(\x_s,\z_s)=\theta_s^0(\x_s,h_s(\z_s))$ such that $\bm V=T^0_s(\X_s,\Z_s)$, where $\bm V$ a random vector with a known lower bounded smooth density $p_v\in\smc{C}^{\infty}([0,1]^{d_{x_s}},\R,B_v)$.
Assume that $T^0_s$ and its inverse belong to a H\"older ball $\smc{C}^{r_s+1}([0,1]^{d_{x_s}+d_{h_s}},[0,1]^{d_{x_s}},B_s)$ 
, while $|\det \nabla_{\x}T^0_s|$ is lower bounded by some positive constant.
\end{assumption}

Then, we obtain the source error rate $\varepsilon_s$.

\begin{lemma}[Source generation error]
\label{thm_floW_source}
Under Assumption \ref{A_BL_source}, setting the hyperparameters of $\Theta_s$ with sufficiently large positive constant set $\{c_L,c_W, c_S, c_B,c_E,c_{\lambda}\}$ as follows: $\mathbb{L}_s=c_L L\log L$, $\mathbb{W}_s=c_W W\log W$, $\mathbb{S}_s=c_SW^2L\log^2W\log L$, $\mathbb{B}_s=c_B$, $\log\mathbb{E}_s=c_E\log(WL)$ and $\lambda_s=c_\lambda$, we obtain that for any $x\geq 1$,
\begin{equation*}
P(\rho_{s}(\gamma_s^0, \hat \gamma_s) \geq x\varepsilon_s)\leq \exp(-c_2 n_s(x\varepsilon_s)^2),
\end{equation*}
with $\varepsilon_s=\beta_s+\delta_s+\varepsilon^h_s$,
some constant $c_2>0$ same in Assumption \ref{A-error}.
Here, $\beta_s=\sqrt{\frac{K^2\log^5 K}{n_s}},
\delta_s=K^{\frac{-2r_s}{d_{s}+d_{h_s}}}$ with $K=WL$,
and $\varepsilon^h_s$ satisfies \eqref{eq-entropy-h} in Lemma \ref{l-composite-entroty}.
Moreover, setting $K\asymp n_s^{\frac{d_{s}}{2(d_{x_s}+d_{h_s}+2r_s)}}$ yields 
\begin{equation*}
    \varepsilon_s\asymp \log^{5/2} n_s\left(n_s^{-\frac{r_s}{d_{x_s}+d_{h_s}+2r_s}}\right)+\varepsilon^h_s.
\end{equation*}
\end{lemma}
\begin{proof}
    This proof of this lemma is the same as that of Theorem \ref{cor-cp-nt},
replacing the approximation error and metric entropy there by those in
Lemma  \ref{l-approx-cp} and Lemma \ref{l-entropy-cp}.
\end{proof}

\subsubsection{Proofs of Subsection \ref{flows-sec3}}
Theorems \ref{thm_flow2-detail} and \ref{cor-cp-nt2-detail} present the formal version of Theorems \ref{thm_flow2} and \ref{cor-cp-nt2} respectively.

\begin{theorem} [Unconditional flows via transfer learning]
\label{thm_flow2-detail}
Under Assumptions \ref{U-error} and \ref{A_g2}, there exists a wide or deep
ReLU network $\Theta_{g_t}$ with specific hyperparameters:
$\mathbb{L}_g=c_L L\log L$, $\mathbb{W}_g= c_W W\log W$, $\mathbb{S}_g= c_S W^2 L\log^2W \log L$, 
$\mathbb{B}_g= c_B$, and $\log\mathbb{E}_g= c_E\log(WL)$, such that the error for unconditional flow generation via transfer learning in Wasserstein distance is,
\begin{equation*}
    P(\smc{W}(P^0_{\x_t}, \hat{P}_{\x_t})\geq x(\varepsilon_t+\varepsilon_s^u))\leq \exp(-c_3 n_s(x\varepsilon^u_s)^2)+\exp(-c_{11} n_t(x\varepsilon_t)^2)
\end{equation*}
for any $x\geq 1$ and 
some constant $c_{11}>0$. 
Here, $c_L,c_W, c_S, c_B,c_E$ are sufficiently large positive constants, $WL\asymp n_t ^{\frac{d_u}{2(d_u+2r_g)}}$ and $\varepsilon_t\asymp n_t^{-\frac{r_g}{d_u+2r_g}}\log^{\frac{5}{2}} n_t$.
\end{theorem} 

\begin{theorem}[Non-transfer unconditional flows]
\label{cor-cp-nt2-detail}
Suppose there exists a sequence $\varepsilon_t^u$ indexed by $n_t$ such that $n_t^{1-\xi}(\varepsilon_t^u)^2\rightarrow\infty$ as $n_s\rightarrow\infty$ and
$P(\rho_u(\theta_u^0,\tilde\theta_u)\geq \varepsilon)\leq \exp(-c_3 n_t^{1-\xi} \varepsilon^2)$ for any $\varepsilon\geq \varepsilon_t^u$ and some constants $c_3, \xi>0$. Under Assumption \ref{A_g2} the same conditions of Theorem \ref{thm_flow2}, 
the error in non-transfer diffusion generation under the Wasserstein distance is, for any $x\geq 1$,
\begin{equation*}
    P(\smc{W}(P^0_{\x_t},\tilde{P}_{\x_t})\geq x(\varepsilon_t+\varepsilon_t^u))\leq \exp(-c_3n_t(x\varepsilon_t^u)^2)+\exp(-c_{11} n_t(x\varepsilon_t)^2).
\end{equation*}
 Here $\varepsilon_t \asymp n_t^{-\frac{r_g}{d_u+2r_g}}\log^{\frac{5}{2}} n_t$.
\end{theorem}

The results in Section \ref{flows-sec3} can be proved similarly as those in Section \ref{sec_3-2}, except that we derive the generation error in the latent variable $\U$ from the flow model theory in Theorem \ref{thm_cp_general}.

\begin{assumption}\label{A_BL_latent}
    There exists a map ${T}^0_u:[0,1]^{d_u}\times \rightarrow [0,1]^{d_u} $ such that $\bm V={T}^0_u(\U)$, where $\bm V$ a random vector with a known lower bounded smooth density in $\smc{C}^{\infty}([0,1]^{d_{u}},\R,B_v)$. Assume that $T^0_u(\bm v)$ and its inverse belong to a H\"older ball $\smc{C}^{r_u+1}([0,1]^{d_{u}},[0,1]^{d_{u}},B_u)$ 
of radius $B_u>0$, while the $|\det\nabla T^0_u|$ is lower bounded by some positive constant.
\end{assumption}
\begin{lemma}[Latent generation error]
\label{thm_pu-flow}
Under Assumption \ref{A_BL_latent},  setting network $\Theta_u$'s hyperparameters with a set of sufficiently large positive constants $\{c_L,c_W, c_S, c_B,c_E,c_{\lambda}\}$ as follows: $\mathbb{L}_u=c_L L\log L$, $\mathbb{W}_u=c_W W\log W$, $\mathbb{S}_u=c_SW^2L\log^2W\log L$, $\mathbb{B}_u=c_B$, $\log\mathbb{E}_u=c_E\log(WL)$, and $\lambda_u=c_\lambda$, we obtain that for any $x\geq 1$,
\begin{eqnarray*}
P(\rho(\theta^0_u,\hat{\theta}_u)\geq x\varepsilon_s^u)\leq  \exp\Big(-c_3 n_s (x\varepsilon_s^u)^2\Big)
, 
\end{eqnarray*}
with $\varepsilon_s^u=\beta_u+\delta_u$ and $c_3>0$ is the same with Assumption \ref{U-error}.
Here, $\beta_u\asymp \sqrt{\frac{K^2\log^5 K}{n_s}},
\delta_u\asymp K^{\frac{-2r_u}{d_u}}$ with $K=WL$.
Moreover, setting $K\asymp n_s^{\frac{d_{s}}{2(d_{u}+2r_u)}}$ yields 
$\varepsilon_s^u\asymp n_s^{-\frac{r_u}{d_u+2r_u}}\log^{5/2} n_s$.
Similarly, $\varepsilon^u_t\asymp n_{t}^{-\frac{{r_u}}{d_u+2 r_u}} \log^{5/2} n_t$.
\end{lemma}

\subsection{ Auxiliary lemmas on neural network approximation theory}
This section restates several neural network approximation results for various
functions, which are used in our proofs.

\label{App-L}
The following lemma constructs a ReLU network to
approximate a step function. Subsequently, denote by
$\lceil x \rceil$ the ceiling of $x$ and denote by $\mathbb{N}$ and $\mathbb{N}^+$ all non-negative and positive integers.

\begin{lemma}[Step function, Proposition 4.3 of \cite{lu2021deep}]
\label{l_step}
 For any \( W, L, d \in \mathbb{N}^+ \) and \( \epsilon > 0 \) with \( K = \lceil W^{1/d} \rceil^2 \lceil L^{2/d} \rceil \) and \( \epsilon \leq \frac{1}{3K} \), there exists a one-dimensional ReLU network $\phi$ with width $4W + 5$ and depth \( 4L + 4 \) such that
\[ 
\phi(x) = \frac{k}{K}, \quad \text{if } x \in \left[\frac{k}{K}, \frac{k+1}{K} - \epsilon \cdot \mathbb{I}_{\{k<K-1\}}\right]; k=0,1,\ldots,K-1.
\]
Moreover, $\phi(\x)$ is linear in $[\frac{k+1}{K} - \epsilon \cdot \mathbb{I}_{\{k<K-1\}},\frac{k+1}{K}]$.
\end{lemma}

The following result allows us to construct a ReLU network with width \( O(s\sqrt{W \log W}) \) and depth \( O(L \log L) \) to approximate function values at
$O(W^2L^2)$ points with an error \( O(W^{-2s} L^{-2s}) \).

\begin{lemma}[Point fitting, Proposition 4.4 of \cite{lu2021deep}]
\label{l_pointfit}
Given any \( W, L, s \in \mathbb{N}^+ \) and \( \zeta_i \in [0, 1] \) for \( i = 0,1,\ldots,W^2L^2 - 1 \), there exists a ReLU network \( \phi \) with width \( 8s(2W + 3)\log_2(4W) \) and depth \( (5L + 8)\log_2(2L) \) such that
\begin{enumerate}
  \item \( |\phi(i) - \zeta_i| \leq W^{-2s}L^{-2s} \), for \( i = 0,1,\ldots,W^2L^2 - 1 \);
  \item \( 0 \leq \phi(x) \leq 1 \), for any \( x \in \mathbb{R} \).
\end{enumerate}
\end{lemma}

The following is a ReLU approximation result for a H\"older class of smooth functions, which is a simplified version of Theorem 1.1 in \cite{lu2021deep} and Lemma 11 in \cite{huang2022error}.
\begin{lemma}[Lemma 11 in \cite{huang2022error}]
\label{l_approx_nn}
 For any $f\in\smc{C}^{r}([0,1]^{d},\R,B)$, there exists a ReLU network $\Phi$ with $\mathbb{W}=c_W(W\log W)$, $\mathbb{L}=c_L(L\log L)$ and $\mathbb{E}=(WL)^{c_E}$ with some positive constants $c_W$, $c_L$ and $c_E$ dependent on $d$ and $r$, such that $
 \sup_{\x\in[0,1]^{d}}|\Phi(\x)-f(\x)|=O( B (WL)^{-\frac{2r}{d}})
 $. 
\end{lemma}

The next lemma describes
how to construct a ReQU network to approximate the multinomial function.
\begin{lemma}[Lemma 1 of \cite{belomestny2023simultaneous}]
\label{ReQU}
For any \( x = (x_1, \ldots, x_k) \in \mathbb{R}^k \) with \( k \in \mathbb{N}, k \geq 2 \), there exists a ReQU neural network 
$
\text{NN} \left( \lceil \log_2 k \rceil, (k, 2^{\lceil \log_2 k \rceil + 1}, 2^{\lceil \log_2 k\rceil}, \ldots, 4, 1) \right),
$
which implements the map \( x \mapsto x_1x_2, \ldots, x_k \). Moreover, this network contains at most \( 5 \cdot 2^{2\left[ \log_2 k \right]} \) non-zero weights.
\end{lemma}

\section{Experiment details}
\label{appendix-B}
\subsection{Conditional image generation}

\paragraph{Datasets and Preprocessing.} 
We conduct our experiments on two standard handwritten‐digit benchmarks.  For source data, we use the MNIST training set (60,000 samples, $28\times28$ grayscale), and for target data we use the USPS training set (7,291 samples, $16\times16$ grayscale).  All images are resized to $16\times16$, converted to tensors, and normalized to $[-1,1]$.  We randomly split USPS into 70\% train (5,103 samples) and 30\% test (2,188 samples) with a fixed random seed for reproducibility.  DataLoaders use a batch size of 256 and shuffle the training splits.

\paragraph{Model architecture.} 
We propose a conditional diffusion model given digit label based on a UNet.  Our \texttt{ClassConditionedUnet} consists of: (1) A learnable embedding layer mapping each digit label $y\in\{0,\dots,9\}$ to a 4-dimensional vector. (2)
A \texttt{UNet2DModel} (from HuggingFace Diffusers) with input channels $1+4=5$, output channels $1$, three down‐sampling blocks \{\texttt{DownBlock2D}, \texttt{AttnDownBlock2D}$\times2$\}, three up‐sampling blocks \{\texttt{AttnUpBlock2D}$\times2$, \texttt{UpBlock2D}\}, and 2 ResNet layers per block.  
At each forward pass, we concatenate the expanded class embedding to the image tensor and predict the noise residual.

\paragraph{Diffusion and optimization.} 
We use a DDPMScheduler with 1,000 timesteps and the “squaredcos\_cap\_v2” beta schedule.  During training, we optimize all parameters with Adam (learning rate \(1\times10^{-4}\)) for 30 epochs on MNIST, recording the loss at each iteration.

\paragraph{Fine‐tuning on USPS.} 
After MNIST pretraining, we freeze the class‐embedding weights and continue training only the UNet backbone on the USPS training split for an additional 30 epochs (Adam, learning rate \(1\times10^{-4}\)).  This transfers digit‐conditioned features learned on MNIST into the USPS domain.

\paragraph{Evaluation.} 
We generate samples by starting from Gaussian noise and iteratively denoising with the learned model, conditioned on test‐set labels. Evalaution metrics (Wasserstein distance) are reported on the USPS test split.

\subsection{Unconditional image generation}

\paragraph{Datasets and preprocessing.}
We extract the digit “3” images from both MNIST and USPS.  All images are resized to $16\times16$, converted to tensors, and normalized to $[-1,1]$.  From MNIST, we take all “3” examples ($N_{\mathrm{MNIST},3}$), and from USPS we similarly filter to $N_{\mathrm{USPS},3}$ examples, then split the latter into 70\% train and 30\% test using a fixed random seed.  DataLoaders use batch size 256, shuffling the training splits.

\paragraph{VAE architecture and joint training.}
We train two independent VAEs (\texttt{AutoencoderKL})---one for MNIST digit ``3'' and one for USPS digit ``3.'' 
Each VAE comprises four down-sampling encoder blocks and four symmetric decoder blocks, with $10$ convolutional layers per block and a latent dimension of $64$.

Both networks are optimized \emph{jointly} for $1\,000$ epochs using Adam (learning rate $1\times10^{-4}$). 
The overall loss function to minimize is
\[
L
=\alpha_{\text{MNIST}}\,
  L_{\text{VAE}}^{(\text{MNIST})}
+\alpha_{\text{USPS}}\,
  L_{\text{VAE}}^{(\text{USPS})}
+\lambda_{\text{align}}\,
  MMD\!\bigl(z^{(\text{MNIST})},z^{(\text{USPS})}\bigr),
\]
where $L_{\text{VAE}}^{(\cdot)}$ denotes the loss for VAE, $\alpha_{\text{MNIST}}=\alpha_{\text{USPS}}=0.1$, 
$\lambda_{\text{align}}=10$, and 
$z^{(\cdot)}$ denotes latent samples from the corresponding VAE. Here MMD encourages the two latent distributions to 
overlap,  aligning the representations learned from MNIST and USPS.

\medskip\noindent\textbf{Maximum Mean Discrepancy (MMD).} 
Given samples $\z^{(1)}=\{z^{(1)}_i\}_{i=1}^{m}$ and $\z^{(2)}=\{z^{(2)}_j\}_{j=1}^{n}$ from two distributions, 
the MMD with kernel $k$ is
\[
MMD^{2}(\z^{(1)},\z^{(2)})=
\frac{\sum_{i\neq i'}  
      k\!\bigl(\z^{(1)}_{i},\z^{(1)}_{i'}\bigr)}{m(m-1)}
+\frac{\sum_{j\neq j'} k\!\bigl(\z^{(2)}_{j},\z^{(2)}_{j'}\bigr)}{n(n-1)}
-\frac{2 \sum_{i=1}^{m}\sum_{j=1}^{n}
      k\!\bigl(z^{(1)}_{i},z^{(2)}_{j}\bigr)}{mn}, 
\]
where we employ a Gaussian (RBF) kernel $k(x,y)=\exp\!\Bigl(-\tfrac{\lVert x-y\rVert^{2}}{2\sigma^{2}}\Bigr)$
with $\sigma=1.0$.

\paragraph{UNet architecture for latent diffusion.}
To model the latent distribution, we use a \texttt{UNet2DModel} with the following configuration:
\textbf{Input/Output channels:} $C_{\mathrm{in}}=C_{\mathrm{out}}=64$ (latent dimensionality).
\textbf{Sample size:} $16\times16$ spatial resolution.
\textbf{Block structure:}
\emph{Down‐sampling}: two blocks, both \texttt{AttnDownBlock2D}.
\emph{Up‐sampling}: two blocks, \texttt{AttnUpBlock2D} then \texttt{UpBlock2D}.
\textbf{Block channels:} \texttt{block\_out\_channels} = (128, 256).
\textbf{Depth:} \texttt{layers\_per\_block} = 10 ResNet layer per block.
\textbf{Normalization:} \texttt{GroupNorm} with \texttt{norm\_num\_groups=1}.

\paragraph{Latent extraction and subsampling.}
After alignment, we encode all MNIST‐“3” images into their 64‐dimensional latents and aggregate them into a pool.  To study sample‐efficiency, we randomly select subsets of size $k\in\{100,500,1000,2000,3500,6000\}$ from this pool to serve as training data for diffusion.

\paragraph{Unconditional diffusion training.}
For each subset size $k$, we train the UNet on the $k$ MNIST latents for 30 epochs.  We use Adam optimizer with learning rate as $1\times10^{-4}$.  

\paragraph{Evaluation.}
We generate USPS‐“3” test samples (198 images) by denoising from Gaussian noise, then use 
the Sinkhorn algorithm to approximate the Wasserstein‐1 distance (blur = 0.05) 
between generated and real USPS latents using the Python GeomLoss library.

\end{document}